\newif\iffigs\figstrue
\documentclass[10pt,a4paper]{article}
\usepackage{latexsym,amssymb,lscape,graphics,setspace}
\usepackage{amsmath}
\usepackage{graphicx}        
\usepackage{longtable}
\usepackage[vcentermath]{youngtab}
\usepackage{multirow}
\usepackage{booktabs}
\usepackage{multirow}
\usepackage{color}
\usepackage{slashed,epsfig}
\usepackage{amsfonts}
\usepackage{cite}
\usepackage{verbatim}
\usepackage{colortbl}
\usepackage[table]{xcolor}
\usepackage{hyperref}       
\usepackage{array}
\usepackage{fancyhdr}
\usepackage{tikz}
\usepackage{multicol}
\usepackage{lscape}
\graphicspath{{images/}}

\newcommand{\mathsym}[1]{{}}

\newtheorem{definizione}{Definition}[section]

\newtheorem{statement}{Statement}[section]

\newtheorem{remark}{Remark}
\newcommand{\bd}{\begin{definizione}}
\newcommand{\ed}{\end{definizione}}

\def\IC{\relax\,\hbox{$\inbar\kern-.3em{\rm C}$}}
\def\IG{\relax\,\hbox{$\inbar\kern-.3em{\rm G}$}}
\def\IB{\relax{\rm I\kern-.18em B}}
\def\ID{\relax{\rm I\kern-.18em D}}
\def\IL{\relax{\rm I\kern-.18em L}}
\def\IF{\relax{\rm I\kern-.18em F}}
\def\IH{\relax{\rm I\kern-.18em H}}
\def\II{\relax{\rm I\kern-.17em I}}
\def\IN{\relax{\rm I\kern-.18em N}}
\def\IP{\relax{\rm I\kern-.18em P}}
\def\IQ{\relax\,\hbox{$\inbar\kern-.3em{\rm Q}$}}
\def\bfzero{\relax\,\hbox{$\inbar\kern-.3em{\rm 0}$}}
\def\IK{\relax{\rm I\kern-.18em K}}
\def\IG{\relax\,\hbox{$\inbar\kern-.3em{\rm G}$}}
 \font\cmss=cmss10 \font\cmsss=cmss10 at 7pt
\def\IR{\relax{\rm I\kern-.18em R}}
\def\ZZ{\relax\ifmmode\mathchoice
{\hbox{\cmss Z\kern-.4em Z}}{\hbox{\cmss Z\kern-.4em Z}}
{\lower.9pt\hbox{\cmsss Z\kern-.4em Z}} {\lower1.2pt\hbox{\cmsss
Z\kern-.4em Z}}\else{\cmss Z\kern-.4em Z}\fi}
\def\bfone{\relax{\rm 1\kern-.35em 1}}

\def\inbar{\vrule height1.5ex width.4pt depth0pt}
\def\bfzero{\relax{\rm I\kern-.18em 0}}
\def\bfone{\relax{\rm 1\kern-.35em 1}}

\DeclareFontFamily{U}{rsf}{} \DeclareFontShape{U}{rsf}{m}{n}{
  <5> <6> rsfs5 <7> <8> <9> rsfs7 <10-> rsfs10}{}
\DeclareMathAlphabet\Scr{U}{rsf}{m}{n}

\newcommand{\Spin}{\mathop{\rm {}Spin}}

\renewcommand{\baselinestretch}{1.0}
\newcommand{\ft}[2]{{\textstyle\frac{#1}{#2}}}
\def\tilde{\widetilde}

\def\1bar{1\hskip -.275cm -}
\def\2bar{2\hskip -.275cm -}
\def\3bar{3\hskip -.275cm -}

\newsavebox{\uuunit}
\sbox{\uuunit}
                 {\setlength{\unitlength}{0.825em}
                      \begin{picture}(0.6,0.7)
                                      \thinlines
                                      \put(0,0){\line(1,0){0.5}}
                                      \put(0.15,0){\line(0,1){0.7}}
                                      \put(0.35,0){\line(0,1){0.8}}
                                     \multiput(0.3,0.8)(-0.04,-0.02){10}{\rule{0.5pt}{0.5pt}}
                      \end {picture}}

\makeatletter \@addtoreset{equation}{section} \makeatother

\def\bfone{\relax{\rm 1\kern-.35em 1}}

\def\bfone{\relax{\rm 1\kern-.35em 1}}
\font\cmss=cmss10 \font\cmsss=cmss10 at 7pt

\newcommand{\so}{\mathfrak{so}}
\newcommand{\su}{\mathfrak{su}}

\newcommand{\sym}{\mathfrak{sp}}
\newcommand{\slal}{\mathfrak{sl}}

\begin{document}
\begin{titlepage}
\begin{center}
\vskip 0.2cm
{{\large {\sc  Tessellation Groups, Harmonic Analysis \\
\vskip 0.2 cm  on Non-compact  Symmetric Spaces  and the Heat Kernel\\
 \vskip 0.2 cm
 in view of Cartan Convolutional neural networks ${}^\dagger$}} }\\
 \vskip 1cm {\sc Pietro Fr\'e\,$^{a,b}$, Federico
Milanesio\,$^{c, d}$, Marcelo Oyarzo\,$^{e}$, \\ Matteo Santoro\,$^{f}$
 and Mario Trigiante\,$^{d, e}$} \vskip 0.5cm
\smallskip
{\sl \small \frenchspacing
${}^a\,$ {\tt Emeritus Professor of} ${}^c\,$ Dipartimento di Fisica, Universit\`a di Torino, Via P. Giuria 1, I-10125 Torino, Italy \\[2pt]
${}^{b}\,${\tt Senior Consultant of } Additati\&Partners Consulting s.r.l, Via Filippo Pacini 36, I-51100 Pistoia, Italy \\[2pt]
${}^d\,$INFN, Sezione di Torino\\[2pt]
${}^e\,$Dipartimento DISAT, Politecnico di Torino,
C.so Duca degli Abruzzi 24, I-10129 Torino, Italy\\[2pt]
${}^f\,$ SISSA (Scuola Internazionale Superiore di Studi Avanzati), Via Bonomea 265, I-34136 Trieste, Italy \\[2pt]

E-mail:  {\tt pietro.fre@unito.it, federico.milanesio@unito.it, msantoro@sissa.it,\\
 mario.trigiante@polito.it, moyarzoca1@gmail.com} }

\begin{abstract}
In this paper, we continue the development of the  Cartan neural networks programme, launched with three previous publications, by focusing on some mathematical foundational aspects that we deem necessary for our next steps forward.  The mathematical and conceptual results are diverse and span various mathematical fields, but the inspiring motivation is unified. The aim is to introduce layers that are mathematically modeled as non-compact symmetric spaces, each mapped onto the next one by solvable group homomorphisms. In particular, in the spirit of Convolutional neural networks, we have introduced the {notion} of Tits Satake (TS) vector bundles where the TS submanifold is the base space. Within this framework, the tiling of the base manifold,  the representation of bundle sections using harmonics, and the need for a general theory of separator walls motivated a series of mathematical investigations that produced both definite and partial results. Specifically, we present the group theoretical construction of the separators for all non-compact symmetric spaces $\mathrm{U/H}$, as well as of the $\Delta_{8,3,2}$ tiling group and its normal  Fuchsian subgroups, respectively yielding the uniformization of the genus $g=3$ Fermat Quartic and of the genus $g=2$ Bolza surface. The quotient automorphic groups are studied. Furthermore, we found a new representation of the Laplacian Green function and the Heat Kernel on Hyperbolic Spaces $\mathbb{H}^{n}$, and a setup for the construction of the harmonic functions in terms of the spinor representation of pseudo-orthogonal groups. Finally, to obtain an explicit construction of the Laplacian eigenfunctions on the Bolza Riemann surface, we propose and conjecture a new strategy relying on the Abel-Jacobi map of the Riemann surface to its Jacobian variety and the Siegel Theta function.
\end{abstract}
\vfill
\end{center}
\noindent \parbox{175mm}{\hrulefill}
\par
${}^\dagger$ P.G. Fr\'e acknowledges support by the Company \textit{Additati\&Partners Consulting s.r.l} during the development of the present research. Furthermore, the Ph.D. fellowships of F. Milanesio and M. Santoro are cofinanced by \textit{Additati\&Partners Consulting s.r.l} at the University of Turin and SISSA, respectively.
\\[5pt]
\end{titlepage}
{\small \tableofcontents} \noindent {}
\newpage
\section{Introduction}
\label{introibo}
In a recent paper \cite{pgtstheory}, we introduced a new mathematical setup for the engineering of neural network architectures under the name of \textbf{PGTS\footnote{PGTS is an acronym for Paint Group Tits Satake} theory of non-compact symmetric spaces}. The essential points of this paradigm are:
\begin{enumerate}
\item The systematic substitution of the Euclidean $\mathbb{R}^n$ space with a symmetric coset manifold $\mathrm{U/H}$ where $\mathrm{U}$ is a
    simple Lie group corresponding to a non-compact real section of a simple complex Lie algebra and $\mathrm{H}$ is the maximal compact
    subgroup of  $\mathrm{U}$. Such manifolds are all Cartan-Hadamard manifolds and as such they admit a \textbf{unique definition of distance
    in terms of geodesic length}. More specifically, the spaces under consideration are hyperbolic in the sense that the curvature is
    strictly negative (see \cite{TSnaviga}) and admit a simply-transitive, completely solvable isometry Lie group  $\mathcal{S}_{U/H}$ (\emph{normal homogeneous Riemannian space}). The last property makes them metrically equivalent (isometric) to the \textbf{solvable Lie group manifold}
    $\mathcal{S}_{U/H}$.\footnote{For the definition of normal homogeneous Riemannian manifolds and for their isometric description in terms of a solvable Lie group manifold, we refer the reader to \cite{Alekseevsky}.} In particular, the parameters of $\mathcal{S}_{U/H}$ provide a \emph{global} coordinate system on $\mathrm{U/H}$ and $\mathcal{S}_{U/H}$ is uniquely associated with the manifold $\mathrm{U/H}$ modulo conjugation in the full isometry group $\mathrm{U}$ \cite{Alekseevsky}.
\item The grouping of these manifolds into universality classes, which provides an ideal mathematical definition
of neural layers.
\item The systematic \textit{suppression of point-wise activation functions} like the sigmoid and its close relatives, the
necessary non-linearity being universally provided by  \textit{generalized exponential maps} from Lie Algebras to the corresponding Lie Groups and
the \textit{generalized logarithm maps} that are the inverse of the former.
\item The use of the rich mathematical structures associated with the new paradigm to improve the interpretability of the learning process.
\end{enumerate}
In a twin pair of papers \cite{TSnaviga,naviga}, it was shown how a generic multi-layer neural network can be reformulated in a form that implements points 1), 2), and 3) of the above paradigm. We named this class of neural network architectures based on the above principles \textbf{Cartan neural networks} in honour of the monumental achievement of Èlie Cartan, who obtained the complete classification of all symmetric spaces and their one-to-one correspondence with the classification of real forms of simple complex Lie algebras \cite{cartan_1926,
Helgasonobook,fre2023book,magnea_introduction_2002}. We will conform to the same name choice for all the networks based on principles 1-4) which might include the further geometric structures discussed in the present paper.
\par
In \cite{TSnaviga,naviga} we addressed the general scheme of \textit{supervised learning} for a  \textit{classification task}. We described an algorithm where each \textit{datum} is linearly mapped (with a
matrix $W_0$ target of learning) to the solvable coordinate vector $\boldsymbol{\Upsilon}$ labeling a point in a non-compact symmetric space $\mathrm{U_1/H_1}$. The latter is the first layer in a sequence of similar layers $\mathrm{U_i/H_i}$,
each being some non-compact symmetric space with some (a priori different) dimension $d_i$.
As discussed at length in \cite{TSnaviga}, the general scheme also allows the non-compact rank and the type of the various $\mathrm{U_i/H_i}$ to be different yet, thanks to the fundamental property of metric equivalence with a suitable solvable group $\mathcal{S}_i$, we can constrain the map from one layer to the next one to be a group homomorphism derived from a linear homomorphism of the corresponding solvable Lie algebras. More specifically, denoting by $K_i$ the map from the  $i^{th}$-layer, described by the space $\mathrm{U_i/H_i}$, to the subsequent one $\mathrm{U_{i+1}/H_{i+1}}$, this map is then described by a group homomorphism between $\mathcal{S}_i$ and $\mathcal{S}_{i+1}$ while the corresponding \emph{push forward} map $K_{i*}$ is a linear homomorphism between the solvable Lie algebras $Solv_i,\,Solv_{i+1}$ generating $\mathcal{S}_i$ and  $\mathcal{S}_{i+1}$, respectively:
$$\forall\,X,\,Y\,\in\, Solv_i\,:\,\,\,\,K_{i*}[X,Y]=[K_{i*}X,\,K_{i*}Y]\,\in \,Solv_{i+1}\,.$$
If $d_i\le d_{i+1}$, $K_i$ can be characterized as an \emph{isometric inclusion} \cite{KN1} characterized by the property that, if $g_i$ and $g_{i+1}$ denote the Riemannian metrics on  $\mathrm{U_i/H_i}$ and  $\mathrm{U_{i+1}/H_{i+1}}$, respectively,
$$\forall\,X,\,Y\,\in\, Solv_i\,:\,\,\,\,g_i(X,Y)=g_{i+1}(K_{i*}X,\,K_{i*}Y)\,.$$
As shown in \cite{KN1}, the mapping $K_{i*}$ between the tangent spaces at corresponding points on the two manifolds, separately isomorphic to $Solv_i$ and $Solv_{i+1}$, is injective. If, on the other hand, $d_i>d_{i+1}$, as a mapping between a linear space $Solv_i$ and a lower-dimensional one  $Solv_{i+1}$, $K_{i*}$ has a non-trivial kernel. If we define the metric $g_i^{(0)}$ on $\mathcal{S}_i$ by the property:
$$\forall\,X,\,Y\,\in\, Solv_i\,:\,\,\,\,g_i^{(0)}(X,Y)=g_{i+1}(K_{i*}X,\,K_{i*}Y)\,,$$
 $g_i^{(0)}$ is singular and thus does not coincide with $g_i$. Nevertheless, being $K_{i*}$  a homomorphism between Lie algebras, ${\rm Ker}(K_{i*})$, of dimension $d_{i}-d_{i+1}$, is an ideal of $Solv_i$ consisting of the \emph{zero-norm} vectors with respect to $g_i^{(0)}$, orthogonal to all the other vectors with respect to the same singular metric.
 Restricted to the solvable Lie algebra $Solv'_i\,\equiv\, Solv_i\,\ominus\, {\rm Ker}(K_{i*})$, $g_i^{(0)}$ coincides with $g_i$ and, therefore, when $d_i>d_{i+1}$, $K_i$ can be characterized as an \emph{isometry} between $\mathcal{S}'_i\equiv \exp(Solv'_i)$, with metric $g_i$ restricted to $Solv'_i\times Solv'_i$, and $\mathcal{S}_{i+1}$.\par
This general characterization of $K_i \, : \, \mathrm{U_i/H_i} \to \mathrm{U_{i+1}/H_{i+1}}$ as an isometric mapping implies its general covariance with respect to the transformations of both the $\mathrm{U_{i}}$ and the $\mathrm{U_{i+1}}$ group.
The action of the two groups on $K_i$ can be formally described as follows:
$$K_i\,\,\rightarrow\,\,\mathrm{U_{i+1}}\circ K_i\circ \mathrm{U_{i}}\,.$$
\subsection{Separators in non-compact symmetric spaces}
In the last layer $\mathrm{U_{N}/H_{N}}$ of the Cartan network, to classify the data points, we introduce separators that are \textit{codimension one homogeneous submanifold of  $\mathrm{U_{last}/H_{last}}$ }. These partition $\mathrm{U_{N}/H_{N}}$ into $K$ chambers, each of which represents one of the $K$ classes. This allows assigning to each data point a probability distribution of belonging to the different classes based on the distance between the image in the last manifold of the datum and each separator. In the above description, the logistic separators are called \textbf{separator walls} because of their similarity to the chamber walls occurring in algebraic geometry quotient varieties $\mathbb{C}^n/\Gamma$ where $\Gamma$ is a discrete group, in particular the Weyl group associated with Lie algebras. To achieve this classification, the separator must be a \textit{non-compact codimension one submanifold halving the ambient space in two subdomains}.
\par
In the case, explicitly treated in \cite{naviga},  where all layers of the Cartan network are hyperbolic spaces $\mathbb{H}^p$:
\begin{equation}\label{Hplani}
  \mathbb{H}^{p} \, \equiv \,  \frac{\mathrm{SO(1,1+p-1)}}{\mathrm{SO(p)}} \, = \, \mathcal{M}^{[1,p-1]}
\end{equation}
the separator walls are totally geodesic submanifolds, and so are symmetric subspaces. For other non-compact symmetric spaces $\mathrm{U/H}$, the separators are homogeneous subspaces but are not totally geodesic. In any case, to achieve the goals of the new Cartan-PGTS paradigm (namely, the geometrical interpretability of the learned parameters), it must be clear how the separator transforms under the isometries of the hosting manifold. This is guaranteed in all cases. In the case of the non-compact rank $r=1$ Tits Satake universality class, whose members are the hyperbolic $p$-spaces, a codimension one symmetric subspace of $\mathbb{H}^{p}$ is simply $\mathbb{H}^{p-1}$ and the \textbf{conjugacy class} of the latter inside the former is parameterized by a set of isometries that are the target of learning in Cartan neural networks.
\par
One of the results of the present paper is the precise group theoretical construction for the other Tits Satake Universality classes of the series
\begin{equation}\label{Mrq}
  \mathcal{M}^{[r,q]} \, \equiv \, \frac{\mathrm{SO(r,r+q)}}{\mathrm{SO(r)} \times\mathrm{SO(r+q)}}
\end{equation}
of convenient $S\mathrm{O(r,r+q)}$ orbits of \textit{homogeneous codimension one separators} that allow the softmax modelling of the probability measure to be extended to all spaces (\ref{Mrq}) (see sect.\ref{separatini}).
\par
\subsection{Fiber bundles and harmonics}\label{sse:fiberharm}
One of the goals of the new Cartan-PGTS paradigm is the clarification of the geometrical nature that lies behind the separation of data into classes. Although it adheres to the principles 1)-4) spelled out above, the algorithms developed in \cite{naviga}  and theorized in \cite{TSnaviga} discard knowledge on the datum and considers each image as a vector that can be mapped to one of the $\mathcal{M}^{[r,q]}$ manifolds, while it is common knowledge that exploiting the translation symmetry of images helps learning. The scheme outlined in \cite{TSnaviga,naviga} can be summarized as:
\begin{equation}\label{formaggio}
 \mathbb{V}_{input} \, \underbrace{\stackrel{\iota_{[\mathcal{Q},\Lambda]}}{\hookrightarrow}}_{\text{injection}} \,\underbrace{\mathcal{M}_1 \, \stackrel{\hat{\mathcal{K}}^1_{[\mathcal{W}_1,\Psi_2]}}{\longrightarrow} \, \mathcal{M}_2\, \longrightarrow \,\dots \,
 \longrightarrow \,\mathcal{M}_{N}}_{\text{hidden layers}} \, \underbrace{\stackrel{\mathfrak{S}}{\longrightarrow}\, \mathcal{M}_+\cup \mathcal{M}_-}_{\text{partition}} \underbrace{\, \stackrel{\sigma}{\longrightarrow}\,\left[0,1\right]_{out}}_{\text{log. regr.}}
\end{equation}
In our setup the covariance of the maps between layers is respected in all steps, yet the geometry of the ambient space is disconnected from the geometrical structure of the analyzed data, precisely because each datum is mapped to the hosting manifold as a single point, while it is better seen as a discretized description of an unknown section, as is implicitly assumed in convolutional networks and formalized in \cite{6_G_CNN}.
In our present effort to apply the Cartan-PGTS paradigm to different types of data, the present paper addresses the task of image classification, which, from our viewpoint, is  essentially defined by the following abstract features:
\begin{enumerate}
\item Each datum of both the training set and, later, of the unknown set to be classified by the trained network,  is in the form of a collection of pairs $(p,\mathbf{v})$, where $p\in \mathcal{M}$ is a point in a manifold $\mathcal{M}$ and $\mathbf{v} \in \mathrm{F}$ is a vector in a vector space or a point in another manifold.
If one gives up the aim of discovering the functional relation between the first and the second array of numbers composing each datum, then a finite number of data can be mapped to differentiable manifolds in a free way where the dimensionalities of the latter become optional. In such cases either there are no hidden functional laws to be discovered, or the hidden laws are consciously ignored and one constructs algorithms that deliberately renounce at enlightening them. The dimensionality of the manifolds to which data are mapped loses any intrinsic meaning as data themselves either do not have any clear-cut dimensionality in their structure, or are deprived of it by decree of the algorithm builder.
  \item The allowed $p$.s are not randomly chosen, rather they belong to an orderly set, typically a grid or a lattice in the manifold
      $\mathcal{M}$, while the vectors $\mathbf{v}$ can assume quite different values at the same $p$ in different data.
  \item There is the supposition that in each datum some kind of hidden law approximately guides the distribution of $\mathbf{v}$ values at
      various $p$.s.  Every datum has its own law, yet the various laws fall into similarity classes, that is the main goal of the geometric
      neural network to define and study at the end of the training.
\end{enumerate}
This typology of data recalls the concept of a fiber bundle. The manifold $\mathcal{M}$
is the base manifold and the second entry $\mathbf{v}$ fixes a point in the fiber $\mathrm{F}$, whether
it is a vector space or another kind of manifold. Hence, each datum is typically the discretization of a
section $\mathfrak{s}$ of the bundle, and the targets of learning are the distinguishing features of each section class, which might very well be geometric in nature.
A more ambitious algorithm, while performing the class separation of the data, should yield a \textit{class separation between submanifolds} in the vector bundle, not points. One can gain insight into the structure of observed phenomena only by enlightening such geometrical similarities.\footnote{
A significant example of what we just stated is provided by the outstanding case of the \textit{geometrical approach to classical thermodynamics}.
We quote only a small collection of papers in a vast literature due a small number of authors
\cite{Ruppeiner_2010,Ruppeiner_2012,Ruppeiner_2012b,Ruppeiner_2013,ruppoRdiag,Ruppeiner_2020,lychaginlecture} who have pursued a study of the phenomenological equations of state for various fluids modeling them as the definition of Legendrian subvarieties in contact varieties.  Because of the complex and
general relationships between contact varieties and symplectic varieties, thermodynamic
states can also be interpreted as Lagrangian subvarieties of symplectic varieties, and the
canonical symplectic form on them naturally connects to a Riemannian metric, which is
the one hypothesized and studied by Ruppeiner, the founder of the geometric approach to
classical thermodynamics. Thermodynamic curvature, something at first sight weird,
proved instead a powerful instrument in the classification of different equations of states.
Consider for instance the classical phase diagram, where we also have separators that
divide the space into sectors corresponding to the gas, liquid, and solid phases. Each
phase is a submanifold with different curvature properties and the separators correspond
to singularities of the curvature. }
Any function or section of a bundle is liable to be uniquely and intrinsically represented in a coordinate-free manner by its harmonic expansion, and its coefficients constitute an efficient and intrinsic representation of the datum. Both the harmonics and their laplacian eigenvalues on manifolds of the $\mathrm{U/H}$ type are completely algebraically determined via PGTS group theory. 
\par


\subsection{The interplay between harmonic expansions and tessellation groups}
\label{intergioco}
\par
In the above-described problem typology, the regularity of the point distributions in the base manifold is a crucial problem. This touches upon the issue of manifold tessellations and tessellation groups that can be generically conceived as discrete subgroups of the isometry group of the manifold.
\par
In the present paper, we confine ourselves to the well-known problem of tessellations of the Hyperbolic Space $\mathbb{H}^2$. As we are interested in discretized sections of vector bundles having the unique Tits Satake element $\mathcal{M}_{TS}^{[r,r+1]}$ of a TS  universality class as base manifold  $\mathcal{M}_B$ and a fixed dimensional vector space
$\mathrm{F}$ as fiber, we arrive at the interplay between the two  main items in the construction:
\begin{description}
  \item[a)] The tessellation group $\mathrm{G_T}$ of the base manifold $\mathcal{M}_B$, which provides a tessellation of the base manifold. The points $p$ of the section classification problem outlined in Section \ref{sse:fiberharm} typically constitute a finite portion of some grid and such grid has to be compared and
      harmonized with a regular tessellation of the base manifold.
  \item[b)] The group theoretical construction of the basis of harmonics $\boldsymbol{\mathfrak{harm}}^{[J]}_m(p)$ for the harmonic expansion of
      any $f\in C^{\infty}(\mathcal{M}_B) \, = \,C^{\infty}\left( \frac{\mathrm{U_{TS}}}{\mathrm{H_{TS}}}\right)$. These, as recalled in
      \cite{pgtstheory}, constitute the eigenfunctions of the ordinary laplacian $\Delta^s$, for scalar functions, or of the appropriate Laplace
      Beltrami operator $\Delta_{LB}^{\mathrm{F}}$ for sections of a vector bundle in a linear representation $\mathrm{F}$ of the compact
      subgroup $\mathrm{H_{TS}}$. The spectrum of eigenvalues of  $\Delta_{LB}^{\mathrm{F}}$ and its eigenfunctions are both completely
      determined in terms of group theory as we have  described in \cite{pgtstheory} specifying to the non-compact symmetric
      spaces $\mathrm{U/H}$ the general principles of harmonic analysis on coset manifolds \cite{castdauriafre}.
\end{description}
The interplay between the two items \textbf{a)-b)} comes from the fact that the eigenvalues $\lambda_J$ of $\Delta_{LB}^{\mathrm{F}}$, while
acting on the space of regular functions defined over the whole coset manifold, are given by the Casimir values $C_J$  for the infinitely many
irreducible linear representations $D^{[J]}$ of $\mathrm{U}$ contributing to the spectrum and the eigenfunctions belonging to a given $\lambda_J$
transform in a corresponding finite dimensional irreducible representation  $D^{[J]}$, namely we have:
\begin{equation}\label{linealeasdehnung}
 \forall g\in \mathrm{U}, \quad\forall p \in \mathrm{U/H} \quad\quad :
 \quad \quad \boldsymbol{\mathfrak{harm}}^{[J]}_m(g\centerdot p) \, = \, D^{[J]}_{mn} (g)
 \, \boldsymbol{\mathfrak{harm}}^{[J]}_n(p)
\end{equation}
In the above equation, by $\boldsymbol{\mathfrak{harm}}^{[J]}_m(p)$ we have denoted the  $\mathrm{dim}D^{[J]}< \infty$ multiplet of
$\Delta_{LB}^{\mathrm{F}}$ eigenfunctions corresponding to the eigenvalue $\lambda_J$, the down indices $m,n$ enumerating them, while by
$g\centerdot p$ we have denoted the action of the group element $g$ on the point $p$ of the base manifold, restituting a point $p^\prime \in
\mathrm{U/H}$. Since the tessellation group $\mathrm{G_T}\subset \mathrm{U}$ is a properly discontinuous infinite subgroup of the continuous
non-compact group $\mathrm{U}$, it follows that each eigenspace of the laplacian provides a $\mathrm{dim}D^{[J]}$ dimensional finite
representation of the tessellation group which typically remains irreducible also when reduced to $\mathrm{G_T}$.
\paragraph{A crucial distinction.} It is at this level that a crucial distinction arises between compact and non-compact coset manifolds.
For compact coset manifolds $\mathrm{U_c/H_c}$ the regular functions $\mathbb{C}^{\infty}\left(\mathrm{U_c/H_c}\right)$ are also square
integrable, namely belong to $L^2\left(\mathrm{U_c/H_c}\right)$, while for the non-compact coset manifolds $\mathrm{U/H}$ with which we are
concerned this is not the case. Furthermore  the finite dimensional representations $D^{[J]}$ that are in one-to-one correspondence  with the
regular harmonics $\boldsymbol{\mathfrak{harm}}^{[J]}_n(p)$ are also unitary in the compact case, while they are not unitary in the non-compact
one. Indeed, as it is well known, all unitary irreducible representations of non-compact groups are infinite dimensional and their carrier space
is a Hilbert space, specifically some appropriate $L^2$ space. Reversely the compact coset manifolds $\mathrm{U_c/H_c}$ do not admit unique
geodesics joining any pair of points and, consequently, are not equipped with a distance function that is an essential token in machine learning
algorithms.
\par
The matter of fact that the regular harmonics are not square integrable does not mean that we cannot consider square integrable functions over
$\mathrm{U/H}$. For instance in the celebrated construction and classification by Bergmann\cite{barmanno} of unitary irreducible representations
(UIR) of $\mathrm{SO(1,2)}$ the Hilbert carrier space $\mathfrak{H}^{J}$ of each representation is $L^2_{J} \left (\mathbb{H}^{2} \right)$ the
label $J$ referring to a specific $\mathrm{U}$ invariant weight function that classifies the UIR. The basis functions of such spaces are no longer
the regular harmonics $\boldsymbol{\mathfrak{harm}}^{[J]}_n(p)$ that do not belong to the space, but others. \\
\\
The important issue in applications to machine learning algorithms is that, whatever data (images for instance) one considers of the type $(p_i,\mathbf{v}_i)$, the number of involved points $p_{i} \in \mathcal{M}_{TS}$ of the base manifold $\mathcal{M}_{TS} \, = \, \mathbb{H}^2$ is finite and it is located in a \textit{compact simply connected region} $\mathfrak{R}\subset \mathbb{H}^2$. Hence the required space of functions is  the space of integrable functions over such region $L^2\left(\mathfrak{R}\right)$. Therefore it is necessary to give an intrinsic group-theoretical characterization of the region $\mathfrak{R}$ in order to have a group-theoretical classification of the basis of functions $L^2\left(\mathfrak{R}\right)$. The tessellations enter the game at this level. The only reasonable way to characterize $\mathfrak{R}$ is to immerse it as a subspace  $\mathfrak{R} \subset \mathcal{F}_{\Gamma}$ in the \textbf{fundamental domain} $\mathcal{F}_{\Gamma}$ of a  finite index normal Fuchsian subgroup $\Gamma \subset \mathrm{G_T}$ of some tessellation group.  The choice of $\mathrm{G_T}$ and of $\Gamma$ is a matter of convenience.  In any case the space of relevant functions becomes $L^2(\mathcal{F}_{\Gamma})$ for which a convenient group theoretical basis is provided by the eigenfunctions of the laplacian:
\begin{equation}\label{conegliano}
  \vartriangle \, \Phi^{\lambda}_{i} \, = \, \lambda \, \Phi^{\lambda}_i \quad ; \quad i\, = \, 1,\dots,n[\lambda]  \quad ; \quad
  \Phi^{\lambda}_{i} \, \in \, L^2(\mathcal{F}_{\Gamma})
\end{equation}
where the degeneracy $n[\lambda]$ is no longer the dimension of an irreducible representation of $\mathrm{PSL(2,\mathbb{R})} \sim
\mathrm{SO(1,2)}$ rather of the finite quotient group:
\begin{equation}\label{caffirone}
  \mathrm{GD} \, = \, \mathrm{G_T}/\Gamma
\end{equation}
Actually if $\mathcal{F}_{\Gamma}$ is smooth and $\Gamma$ provides a tessellation of $\mathbb{H}^2$ by means of all the images  of the former
through each of its elements:
\begin{equation}\label{crisantullo}
  \mathbb{H}^2 \, = \, \bigcup_{\gamma \in \Gamma} \gamma\left(\mathcal{F}_{\Gamma}\right)
\end{equation}
the fundamental domain $\mathcal{F}_{\Gamma}$ is the uniformization of a genus $g\geq 2$ compact, oriented Riemann surface $\Sigma_\Gamma$, so that the grid of data is mapped to such Riemann surface.
\par
 In such classical field, we found some corners so far only partially developed. Explicitly, we worked out in full detail the
(8,3,2) tessellation leading to two different Fuchsian groups of finite index, with 16 generators and 8 generators, respectively. As we were able to infer a posteriori from the literature, the two interlaced constructions respectively correspond to the uniformization of the \textbf{Fermat Quartic} of genus 3 and the uniformization of the \textbf{Bolza surface}  of genus two. The two quotient groups $\mathrm{GD}$ are respectively
of order 96 and order 48, the latter $\mathrm{GD_{48}}$ being a normal subgroup of the former $\mathrm{GD_{96}}$. Relying on the Bolza tessellation \cite{bolzano,cuocogiuseppe,bogobolza,bolzalgo,Attar_2022,saha2023complexitybolzasurface} of $\mathbb{H}^2$, we sketch how this scheme can be used to construct neural networks for image classification (see sect. \ref{pitture}), yet the reader should appreciate that our goal is not the specific exercise with pixels, but rather the illustration of the methods based on one side on tessellations of the base space and the other on the use of suitable harmonic expansions.
\par
\subsection{Tits Satake vector bundles}
\label{Tsvettorifibrati} Let us generically consider the set of manifolds forming a Tits Satake universality class, for instance those of the class
\begin{equation}\label{titstot}
  \mathcal{M}^{[r,q]}\, \equiv \,\frac{\mathrm{SO(r,r+q)}}{\mathrm{SO(r) \times SO(r+q)}}
\end{equation}
Relying on the \textbf{PGTS theory} fully formulated and explained in \cite{pgtstheory} we can consider the Tits Satake projection:
\begin{equation}\label{proietta}
  \pi_{TS} \quad : \quad \mathcal{M}^{[r,q]} \, \longrightarrow \, \mathcal{M}^{[r,1]}
\end{equation}
The fibers $\pi_{TS}^{-1}(p)$ over each point $p\in \mathcal{M}^{[r,1]}$ of the base manifold have the structure of a $r\times (q-1)$ vector space $\mathbb{V}^{(r\mid q-1)}$ that corresponds to the carrier space for the representation $(r\mid q-1)$ of the subgroup $\mathrm{SO(r) \times SO(q-1)}\subset \mathrm{SO(r,r+q)}$. Correspondigly, the original manifold $\mathcal{M}^{[r,q]}$ can be viewed as the total manifold
\begin{equation}\label{totalspazio}
  \text{tot}\left[\mathcal{E}^{[r(q-1)]}\right] \, = \, \mathcal{M}^{[r,q]}
\end{equation}
of a vector bundle:
\begin{equation}\label{carneadtwo}
  \mathcal{E}^{[r(q-1)]} \, \stackrel{\pi_{TS}}{{ \longrightarrow}}\, \mathcal{M}^{[r,1]}
\end{equation}
that has the Tits Satake projection as projection, the vector $\mathbb{V}^{(r\mid q-1)}$ as standard fiber:
\begin{equation}\label{carneadthree}
  \mathrm{F}  \, = \, \mathbb{V}^{(r\mid q-1)}
\end{equation}
and the subpaint group $\mathrm{G_{subPaint}} \, = \, \mathrm{SO(q-1)}$, multiplied by its normalizator $\mathrm{SO(r)}$ in $\mathrm{H_c}$, as
structural group:
\begin{equation}\label{carneadstruc}
  \mathrm{G_{struc}} \, = \, \mathrm{SO(r)\times SO(q-1)}
\end{equation}
At this point it is convenient to recall that, in the case of \textit{convolutional neural networks}, a series of research papers \cite{4_GoverH_CNN1,5_GoverH_CNN2,6_G_CNN,Bronstein_2017,2_covariance} has proposed the modeling of the network architecture in terms of non-linear maps between sections of different associated vector bundles of a principle bundle with structural group
$\mathrm{H}$:
\begin{equation}\label{buonperlui}
\mathcal{P}\, \stackrel{\pi}{\longrightarrow} \, \mathcal{M} \quad : \quad \forall p\in \mathcal{M}  \quad \pi^{-1}(p) \simeq \mathrm{H}
\end{equation}
Given the principal bundle and a collection of $n$ linear finite dimensional representations of the structural group that we name $D_\alpha [\mathrm{H}]$ ($\alpha=1,\dots ,n$), whose dimension we denote $d_\alpha$ and whose carrier $d_\alpha$-dimensional vector space we denote $\mathbb{V}_\alpha$ we have the associated rank $d_\alpha$  vector bundles
\begin{equation}\label{buonperloro}
\mathcal{E}_\alpha \stackrel{\pi_\alpha}{\longrightarrow} \, \mathcal{M} \quad : \quad \forall p\in \mathcal{M}
\quad \pi^{-1}(p) \simeq \mathbb{V}_\alpha
\end{equation}
and each layer of the convolutional neural network is mathematically modeled as the space of sections of the $\alpha$th associated vector bundle:
\begin{equation}\label{curaro}
  \text{$\alpha$th layer } \, = \, \Gamma\left[\mathcal{E}_\alpha, \mathcal{M}\right]
\end{equation}
the operations connecting each $\alpha$ layer to the next one $\alpha+1$, are conceptualized as a non-linear map (\textit{the convolution}):
\begin{equation}\label{stricnina}
\mathcal{I}_\alpha \quad: \quad \Gamma\left[\mathcal{E}_\alpha, \mathcal{M}\right] \, \longrightarrow \,
\Gamma\left[\mathcal{E}_{\alpha+1}, \mathcal{M}\right]
\end{equation}
and the additional requirement of \textbf{equivariance} of such maps with respect to the action of the structural group $\mathrm{H}$ can be introduced.

\begin{remark}[Equivariance and convolution]
As recalled in \cite{6_G_CNN}, every map between sections that satisfies equivariance requirements can be expressed in terms of a convolution with an integral kernel. In this sense the equivariant networks sketched in this paper are \emph{convolutional}.
\end{remark}
\par In this setup, as proposed by \cite{4_GoverH_CNN1,5_GoverH_CNN2}, one particularly interesting case is that of homogeneous spaces $\mathrm{G/H}$ ($\mathrm{G}$ being a Lie Group and $\mathrm{H}\subset \mathrm{G}$ a Lie subgroup). Here, the total space of the principal bundle is just $\mathrm{G}$, the base manifold is  $\mathrm{G/H}$, and the structural group is $\mathrm{H}$:
\begin{equation}\label{buonpertutti}
\mathrm{G}\, \stackrel{\pi}{\longrightarrow} \, \mathrm{G/H} \quad : \quad \forall p\in \mathrm{G/H} \quad \pi^{-1}(p) \simeq \mathrm{H}
\end{equation}
Comparing eq.(\ref{buonpertutti}) with eq.s(\ref{carneadtwo}) and (\ref{proietta}) we see that the PGTS framework of \textbf{Tits Satake universality classes} provides a more articulated but also more concrete way of realizing the architecture of a convolutional neural network as described in \cite{4_GoverH_CNN1,5_GoverH_CNN2}. The essential difference with the setup envisaged by the authors of the quoted papers is that they considered compact homogeneous spaces $\mathrm{G/H}$ while in our PGTS setup, all coset manifolds are non-compact. The base manifold is non-compact, since it is the Tits Satake submanifold of an entire universality class. Vector bundles on non-compact base manifolds are all globally trivial, yet they might develop non-triviality when we reduce them to the compact fundamental domain of some Fuchsian Group $\Gamma$. In that case, the base space becomes a compact Riemann surface, and vector bundles on Riemann surfaces can have non-trivial characteristic classes.
\par
Let us fix the non-compact rank $r$ and consider a fixed minimal value  $q_{min}$ such that $r\times(q_{min} -1)$ is the dimensionality of the feature vector for our class of data.  Indeed, each data point should be thought of as the following pair:
\begin{equation}\label{paiodato}
 \text{datum} \, = \,  \{p,\mathbf{v}\} \quad ; \quad p\in \mathcal{M}^{[r,1]} \quad ; \quad \mathbf{v} \in \mathbb{V}^{(r\mid q_{min}-1)}
\end{equation}
where $p$ is a point in the base manifold and $\mathbf{v}$ is a point in the vector fiber over it. Therefore the structural group is:
\begin{equation}\label{bardaccio}
  \mathrm{G_{struc}} \, = \, \mathrm{SO(r) \times SO(q_{min} -1)}
\end{equation}
In other words the data constitute a collection of points on a surface  $\mathfrak{X}\subset \mathcal{M}^{[r,q_{min}]}$ of dimensionality equal to
that of the Tits Satake submanifold $\mathcal{M}^{[r,1]}$ that corresponds to some suitable section of the vector bundle:
$\mathcal{E}^{[r(q_{min}-1)]}$ namely
 \begin{equation}\label{sezionatore}
   \boldsymbol{\mathfrak{s}} \in \Gamma\left[\mathcal{E}^{[r(q_{min}-1)]},\mathcal{M}^{[r,1]} \right] \quad ; \quad \forall p
   \in \mathcal{M}^{[r,1]}\, ,
   \quad \boldsymbol{\mathfrak{s}}(p) \in \pi_{TS}^{-1} (p) \subset \mathcal{M}^{[r,q_{min}]}
 \end{equation}
Indeed the pair $\{p,\boldsymbol{\mathfrak{s}}(p)\}$ can be viewed as a parametric description of the aforementioned surface  $\mathfrak{X}$.
\par
The appropriate way of intrinsically representing the general form of sections of vector bundles of the considered type is through harmonic analysis on coset manifolds, which is completely algebraic and just founded on representation theory of the involved groups and subgroups. The coefficients $c_D^\ell$ of the harmonic expansion of the bundle section:
\begin{equation}\label{harmexpgenerale}
  \boldsymbol{\mathfrak{s}}(p) \,=\, \sum_{D(\mathrm{G_TS})\in \mathfrak{S}_{[r,q_{min}}]  }\sum_{\ell=0}^{d_D} \boldsymbol{c}_D^\ell \,
  \pmb{\mathfrak{harm}}^{D}_\ell(p) \,
\end{equation}
constitute the coordinate-free intrinsic representation of the section shapes and must be the target of learning.  In eq.(\ref{harmexpgenerale})
the symbol $\mathfrak{S}_{[r,q_{min}]} $ denotes the set of representations of the Tits Satake group $\mathrm{G_{TS}}\, = \, \mathrm{SO(r,r+1)}$
that, once restricted to the compact subgroup $\mathrm{H_{TS}}\, = \, \mathrm{SO(r)\times SO(r+1)}$ contain the representation of
$\mathrm{H_{TS}}$ in which the Tits Satake fibers are located. As we saw above,  with respect to the group
\begin{equation}\label{lukardon}
  \mathrm{H_{TS}}\times \mathrm{G_{subPaint}}\, = \, \mathrm{SO(r)} \, \times\,  \mathrm{SO(r+1)} \, \times \, \mathrm{SO(q-1)}
\end{equation}
the Tits Satake fibers are in the representation
\begin{equation}\label{mediatore}
  \left(r\mid 1 \mid (q-1) \right)
\end{equation}
namely, they are in the vector defining representation of $\mathrm{SO(r)}$ and of the subPaint group and are in the singlet representation of $\mathrm{SO(r+1)}$. Hence the set of representations of $\mathrm{SO(r,r+1)}$ contributing to the expansion in equation (\ref{harmexpgenerale}) are
those that, once reduced to the compact subgroup $\mathrm{SO(r)} \, \times\,  \mathrm{SO(r+1)}$ contain the representation $\left(r\mid 1\right)$.
\par
\par
The homomorphism maps between the total bundle spaces (\ref{totalspazio}):
\begin{equation}\label{totalspaziomappi}
 \mathcal{I}_\alpha \quad : \quad  \text{tot}\left[\mathcal{E}^{[r(q_\alpha -1)]}\right] \, \longrightarrow \, \text{tot}\left[\mathcal{E}^{[r(q_{\alpha+1} -1)]}\right]
\end{equation}
are equivariant maps, as in eq.(\ref{stricnina}), from the space of sections of the $\alpha$-th vector bundle to $(\alpha+1)$-vector bundle on the same base space.
\par
\subsubsection{Representation spectrum in the cases \texorpdfstring{$r=1,2$}{r=1,2}}
\paragraph{\sc $r=1$.} For $r=1$ the harmonic constraint is very simple. Since  $\mathrm{SO(1)}=\mathrm{Id}$, the allowed representations are those that, once reduced to the subgroup $\mathrm{SO(2)} \subset
\mathrm{SO(1,2)}$,  contain the singlet. These are only the even representations in the
tensor product of the fundamental vector representation and they correspond to the
standard harmonic expansion in terms of the fundamental matrix $\mathcal{M}_v^{AB}$
as described in section 10 of \cite{pgtstheory}. Similar conclusion is obtained if we describe the Poincar\'e plane as the coset manifold $\mathrm{SL(2,\mathbb{R})/SO(2)}$, namely utilizing the spinor instead of the vector representation. The admitted representations of $\mathrm{SL(2,\mathbb{R})}$ are those that are even in the tensor product of the fundamental spinor representation, and are obtained from the spinorial symmetric matrix $\mathcal{M}_v^{AB}$ as also described in section 10 of \cite{pgtstheory} (see the explicit development in section \ref{spinatore} of the present paper).
\paragraph{\sc $r=2$.} In the case $r=2$ we can also use the spinor representation of the Tits Satake group $\mathrm{SO(2,3)}$ which is
the fundamental  $4$-dimensional representation of the symplectic group $\mathrm{Sp(4,\mathbb{R})}$. In the spinor representation, the subgroup
$\mathrm{SO(2)\times SO(3)} \subset \mathrm{SO(2,3)}$ is replaced by its double covering  $\mathrm{U(2)} \, = \, \mathrm{U(1) \times SU(2)} \subset \mathrm{Sp(4,\mathbb{R})}$. Hence, in the harmonic expansion, we have a contribution only from those representations of
$\mathrm{Sp(4,\mathbb{R})}$ that, once reduced to $\mathrm{U(1)} \times \mathrm{SU(2)}$ contain the $\mathrm{SU(2)}$ singlet with a non-trivial
$\mathrm{U(1)}$ charge. A careful study of this constraint is on our agenda for a next paper. Indeed, the case $r=2$ is quite challenging in view
of the parabolic generators of the discrete subgroup described in section 10 of \cite{pgtstheory}. Thanks to such parabolic subgroups, planar
Euclidean lattices admit a canonical embedding into the six-dimensional Siegel plane, without any distortion.
\par
\subsection{Outline of the results and of the contents}
We conclude this introduction with a short summary of the results and of the contents.
\subsubsection{Tessellation of the base manifold}
In order  to take advantage of the general conceptual setup presented in the previous lines and develop concrete algorithms that implement such
ideas we had to consider, for the case $r=1$, the possible tessellations of the Tits Satake base manifold, in our case the Hyperbolic Plane
$\mathbb{H}^2$. As already anticipated, we focused on the $(8,3,2)$ tiling group (see below) that, with respect to the classic $(7,3,2)$ one, although it misses the Hurwitz surfaces, has two very much significant advantages :
\begin{description}
  \item[a)] It introduces tilings that, like that associated with the genus $g=2$ Bolza surface (see below) can reproduce quadrangular shapes of
      the tiles, matching the shapes of  Euclidean square tilings.
  \item[b)] Since the trigonometric functions of $\pi/8$ and of its multiples are solutions of algebraic equations up to the fourth order, the
      coordinates of all vertices of a tiling can be expressed in terms of multiple radicals constructed only with square roots.
\end{description}
\par Within the scope of this study we have provided very explicit formulae for the generators of the Fuchsian groups, that are 16 in the case of the Fermat Quartic and 8 in the case of  Bolza surface, deriving also explicit formulae in terms of radicals for the vertices of the entire tessellation. We have also clarified  the structure of the two quotient groups of which we have constructed an explicit  6-dimensional representation which is irreducible for the order 96 group associated with
Fermat Quartic  and contains instead a reducible representation of the order 48 one,
associated with the Bolza surface. Indeed the Bolza quotient group is a normal subgroup
of index 2 inside the Fermat Quartic quotient group of order 96.
\subsubsection{The harmonics}
The main result about the harmonics contained in this paper is their systematic
reformulation on the basis of the spinor representation rather than on the vector ones for
the groups $\mathrm{SO(1,2+q)}$ which introduces a series of distinctive advantages to
be appreciated by the reader after reading the previous sections of the paper.
\subsubsection{Length spectrum and the heat kernel}
In addition to the above results, by revisiting the whole compound of conceptions
associated with the so named \textit{length spectrum} of Fuchsian conjugacy classes and
with the derivation of the \textit{Fuchsian fundamental domain}, we have shown that
many of these structures extend from the 2-dimensional case of the Hyperbolic Plane to
the entire Tits Satake Universality Class of which the Hyperbolic Plane is the unique
maximally split representative. This generalization occurs also for the Green Function of
the Laplacian operator (related with the heat kernel), which we have shown to obey, in all
cases comprised in the considered TS class, a hypergeometric differential equation. This
rewriting improves both the elegance and the practicity of other formulations existing in
the literature \cite{buserbook,macloffo,cuocogiuseppe,Chutassella}, since we do not need
to resort to complexification of the eigenvalues, rather we have a naturally real function
with the appropriate exponentially decreasing behavior that ensures convergence of all
convolutional integrals.  Furthermore the new formulation of the Green function is
adequate for various possible approximate definition of the heat kernel (by truncating the
summation over eigenvalues to a certain order) that are to be explored in forthcoming
work.
\subsubsection{Group theoretical construction of separators in the \texorpdfstring{$r\geq 2$}{r>=2} series}
In section \ref{separatini} we provide an explicit construction of the separator in the $r=2$ case of a Tits Satake universality class. The result that also shows that the
separator is not a symmetric space and hence not geodesically complete inside the
symmetric ambient space is very important for the further development of PGTS
programme. Indeed it was the so far missing ingredient in order to extend the setup of the
twin papers \cite{TSnaviga,naviga} to the $r=2$ case.
\par
Having thoroughly described the conceptual scope of the present paper and shortly summarized the obtained results we go over to the development of
the outlined programme.
\section{Regular Tessellations and Coxeter Groups}
\label{coxetto}
We come now to discuss  the mathematical foundations of  regular tessellation (or tiling)
schemes for the hyperbolic plane $\mathbb{H}^2$. The latter is of interest to us in the
present paper as base manifold of a  vector--bundle of the type described in section
\ref{Tsvettorifibrati}, whose total space we identify with one of the symmetric manifolds
$\mathcal{M}^{[1,q]}$ in the same universality class of which $\mathbb{H}^2$ is the
unique maximally split representative, namely $\mathcal{M}^{[1,1]}$ .
The proper mathematical framework underlying all regular tiling
schemes, not only in two dimensional, but also in higher dimensional constant curvature
manifolds, is provided by the theory of \textbf{Coxeter groups}, which comprises
spherical, flat Euclidean, and hyperbolic cases.  Among the latter we have the Siegel upper
complex planes, and  our next goal, the study of the rank two non maximally split
symmetric spaces $\mathrm{SO(2,2+2s)/SO(2)\times SO(2+2s)}$ leads, as we explained
in \cite{pgtstheory} and already recalled in the introduction,  to the Tits Satake
submanifold $\mathrm{Sp(4,\mathbb{R})/U(2)}$, \textit{i.e.} the Siegel plane of genus 2,
which is a hyperbolic space of dimensions 6. Therefore, the even Coxeter groups
embedded into $\mathrm{Sp(4,\mathbb{R})}$ are next items in our research efforts.
\par
\subsection{Coxeter Groups} \label{gencoxdefi} 
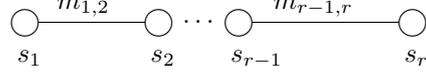
\begin{figure}
\vskip 2cm
\centering
\begin{picture}(50,150)
\put(-30,160){\circle{10}}\put(-33,145){$s_{1}$}
\put(-25,160){\line(1,0){40}} \put(-18,165){$m_{1,2}$}
\put (20,160){\circle {10}} \put (17,145){$s_{2}$} \put (29,160){$\dots$}
\put (50,160){\circle {10}}\put(63,165){$m_{r-1,r}$} \put (47,145){$s_{r-1}$} \put (55,160){\line (1,0){55}} \put
(115,160){\circle {10}} \put (112,145){$s_{r}$}
\end{picture}
\vskip -4cm \caption{\label{coxxetto} The notation for Coxeter graphs: $s_i$
are the idempotent generators $s_i^2 \, = \, \mathbf{e}$ while $m_{i,j}\in
\mathbb{N}$ is the order of the element $s_i\cdot s_j$. This number is
apposed on the edge of the graph if $m_{i,j}>3$. Lines with no marker mean
that the corresponding $m_{i,j}=3$. If there is no line joining vertex $i$
with vertex $j$ it means that $m_{i,j}=2$.   }
\end{figure}

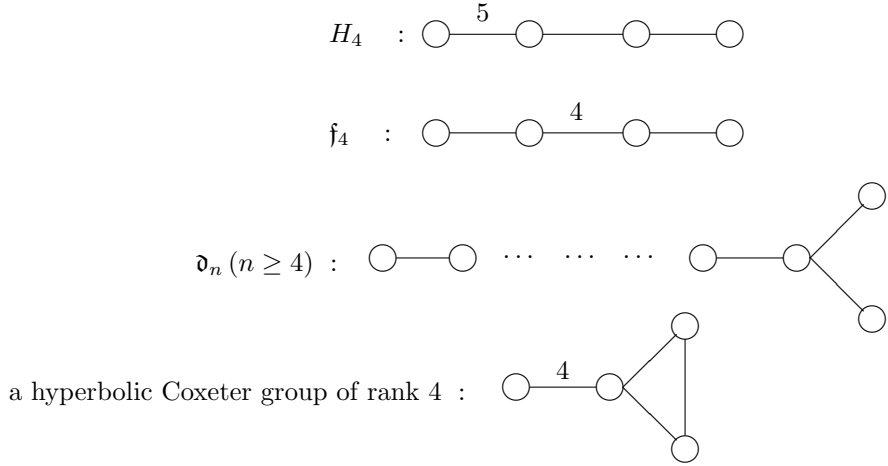
\begin{figure}
\vskip 2cm
\centering
\begin{picture}(50,150)
\put(-70,157){$H_4 \quad:\quad $}\put(-30,160){\circle{10}}
\put(-25,160){\line(1,0){25}} \put(-15,165){$5$}
\put (5,160){\circle {10}} \put(10,160){\line (1,0){30}}
\put (45,160){\circle {10}} \put (50,160){\line (1,0){25}} \put
(80,160){\circle {10}}
\end{picture}
\vskip -4cm
\begin{picture}(50,150)
\put(-70,157){$\mathfrak{f}_4 \quad:\quad $}\put(-30,160){\circle{10}}
\put(-25,160){\line(1,0){25}} \put(20,165){$4$}
\put (5,160){\circle {10}} \put(10,160){\line (1,0){30}}
\put (45,160){\circle {10}} \put (50,160){\line (1,0){25}} \put
(80,160){\circle {10}}
\end{picture}
\vskip -4cm
\begin{picture}(50,170)
\put (-120,165){$\mathfrak{d}_{n}\left(n\geq 4\right)
\,\, :$}\put(-50,170){\circle{10}}\put(-45,170){\line(1,0){20}} \put (-20,170){\circle {10}} \put
(-5,170){$\dots$} \put (18,170){$\dots$} \put (40,170){$\dots$}
\put (70,170){\circle{10}}  \put (75,170){\line (1,0){25}} \put
(105,170){\circle{10}}  \put (110,170){\line (1,1){20}} \put
(110,170){\line (1,-1){20}} \put (133.2,193.2){\circle{10}} \put (133.2,146.8){\circle {10}}
\end{picture}
\vskip -5cm
\begin{picture}(50,190)
\put (-190,165){$\text{a hyperbolic Coxeter group of rank 4}\,\, :$}
\put (0,170){\circle{10}}  \put (5,170){\line (1,0){25}} \put(15,173){$4$}
\put(35,170){\circle{10}}  \put (40,170){\line (1,1){20}} \put
(40,170){\line (1,-1){20}} \put (63.2,193.2){\circle{10}} \put (63.2,146.8){\circle {10}}
\put(63.2,151.8){\line(0,1){36.4}}
\end{picture}
\vskip -3cm
\caption{\label{exampli} In this figure we illustrate the notion of Coxeter diagrams with  four examples.
Two examples, the second and the third, can be recognized by the reader to be Dynkin diagrams of semisimple
Lie algebras in a slightly different notation. Indeed in the new notation all Dynkin diagrams are also Coxeter diagrams,
since all the Weyl groups associated with all root systems are Coxeter groups. On the other hand there are Coxeter diagrams
that are not Dynkin, as the first and the last in the present figure. Indeed there are a lot of Coxeter groups that are not Weyl groups.
The first example $H_4$ is a Coxeter group generated by reflection in an Euclidean space. The last example is an instance of Hyperbolic
Coxeter Group. }
\end{figure}

Coxeter Groups happen to be a very important and fertile generalization of the Weyl groups
associated with semisimple Lie Algebras (for a comprehensive and very clear exposition of the whole theory see the book \cite{humphreybook2}). They
find applications in many issues of geometry and in particular in the issue of regular space tessellations, Euclidean, spherical and hyperbolic.
Although the latter is a classical topic, it continues to be a field of active physical mathematical research, with application in many
directions, ranging from solid state physics, combinatorics in computer science and statistical models of various types. Just for illustrative purposes we quote some quite recent articles that contain also a lot of additional references \cite{howarth2023,Lux_2023,Chutassella}.
\paragraph{\sc The formal definition} \textit{The formal definition of a Coxeter group $\mathfrak{C}$ is provided in terms of a set of generators
$s_i$ ($i=1,\dots,r$) whose number is named \textbf{the rank of the group} and for this reason it is denoted $r$ and a set of relations. The first
primary relations consist of the  compulsory order 2 of all generators  $s_i$:
\begin{equation}\label{idempotente} s_i^2 \, = \, \mathbf{e} \quad ;
\quad i\,=\, 1,\dots, \, r
\end{equation} where $\mathbf{e}\in \mathfrak{C}$
is the neutral element of the group. All other relations are encoded in a positive integer valued symmetric $r \times r$ matrix
\begin{equation}\label{matalone} m_{i,j} \, = \, m_{j,i} \, \in \, \mathbb{N}
\quad ; \quad \forall i\neq j \,\,\, m_{i,j} \, \geq \, 2 \quad ; \quad
m_{i,i}\, = \, 1 \end{equation} such that we have:
\begin{equation}\label{caraglione} \left(s_i \cdot s_j\right)^{m_{ij}} \, =
\,  \mathbf{e} \end{equation} }
\paragraph{\sc The Coxeter graph $\Gamma(\mathfrak{C})$} The previous definition is reminiscent, at the group level,
of the Chevalley--Serre presentation of commutation relations for semisimple Lie Algebra generators;  indeed the matrix $m_{ij}$
is a generalization of the \textbf{Cartan matrix}. Therefore it is not surprising that the whole structure of a Coxeter group
can be encoded in a Coxeter graph $\Gamma(\mathfrak{C})$ very similar to a Dynkin Diagram. The basic observation is that in
Lie Algebra theory the nodes of the Dynkin diagram are supposed to represent the \textbf{simple roots} $\alpha_i$ ($i=1,\dots, r$),
all the roots of the root system being linear combinations of the simple ones with all positive or all negative integer coefficients.
The root system not only identifies the corresponding Lie Algebra but also defines the \textbf{Weyl group},
which is the discrete group of orthogonal transformations in the Euclidean space $\mathbb{R}^r$ generated by all \textbf{reflections}
with respect to all \textbf{simple roots}. The product of two reflections with respect to two different simple roots $\alpha_i$ and $\alpha_j$
is a rotation of
an angle $\theta_{ij}$ which is related to the scalar product $\alpha_i\cdot \alpha_j$ and always
happens to be an integer fraction of $\pi$, so that the rotation $\exp[\theta_{ij}]$ has a definite finite order.
Regarding the Dynkin diagram as a graphical description of the Weyl group provides a natural set of rules to construct the Coxeter
graph $\Gamma(\mathfrak{C})$ of an abstract Coxeter group $\mathfrak{C}$.
\begin{description}
\item[1)] There are as many circular vertices as the elementary generators $s_i$.
\item[2)] If no line joins a pair
$s_i$,$s_j$ of generators it means that $\left( s_i \cdot s_j\right)^2 \, = \, \mathbf{e}$, namely $m_{i,j}=2$.
\item[3)] If there is a simple
line with no marker joining the pair $s_i$,$s_j$ it means that $\left( s_i \cdot s_j\right)^3 \, = \, \mathbf{e}$, namely $m_{i,j}=3$. \item[4)]
If $\left( s_i \cdot s_j\right)^{m_{i,j}} \, = \, \mathbf{e}$ with order ${m_{i,j}}\, = \, n>3$ than the two circular vertices are joined by a
line marked with the number $n$.
\end{description}
This rule is graphically represented in fig.\ref{coxxetto}.
We present an illustration of Coxeter graphs in fig.\ref{exampli} which purposely displays two cases of a
Coxeter graph coinciding with a Dynkin diagram (the corresponding Coxeter group is a Weyl group) and two cases that are not Dynkin,
namely the corresponding Coxeter group is not a Weyl group.
\subsection{Geometric representation of a Coxeter group \texorpdfstring{$\mathfrak{C}$}{C}}
\label{georepcox} As we remarked before, the Coxeter graph $\mathfrak{C}$ is a graphical representation of a symmetric $r\times r$ matrix,
just as the Dynkin diagram is a graphical representation of the Cartan matrix which, however, is not symmetric (it is symmetric only in
the simply laced case). The matrix $m_{i,j}$, can be used to introduce another symmetric $r\times r$ matrix $\kappa_{i,j}$ according with
the following rule: \begin{equation}\label{pernacito}
\kappa_{i,j}\, \equiv \, - \, \cos\left[\frac{\pi}{m_{i,j}} \right]
\end{equation}
If we associate each elementary generator $s_i$ with a basis vector $\boldsymbol{\alpha}_i$ of a real $r$-dimensional vector space:
\begin{equation}\label{rappezzo}
  V \, = \,  \left\{ \mathbf{v} \, | \, \mathbf{v} \, = \, \sum_{i=1}^r \, v^i \, \boldsymbol{\alpha}_i \quad , \quad v^i \in \mathbb{R} \right\}
\end{equation}
and we use the matrix $\kappa_{i,j}$ as the definition of a bilinear form on $V$:
\begin{equation}\label{bilinearform}
 \forall\mathbf{v},\mathbf{u} \,\in \, V \quad : \quad \kappa\left(\mathbf{v},\mathbf{u}\right)\, \equiv \, \kappa_{i,j} v^i \, u^j
\end{equation}
we can realize a homomorphism of the abstract Coxeter group $\mathfrak{C}$ into $\mathrm{Hom}(V,V)$
\begin{equation}\label{homorfo}
  \phi \, : \, \mathfrak{C} \, \longrightarrow \, \mathrm{Hom}(V,V)
\end{equation}
mapping the generators $s_i$ into the reflection with respect to the associated basis vector $\boldsymbol{\alpha}_i$:
\begin{eqnarray}\label{grammofono}
  \phi(s_i) &= & \sigma_{\boldsymbol{\alpha}_i} \nonumber\\
 \forall\mathbf{v}\in  V  \quad :\quad \sigma_{\boldsymbol{\alpha}_i} (\mathbf{v}) &\equiv & \mathbf{v} \,
 - \, 2\, \kappa(\mathbf{v},\boldsymbol{\alpha}_i) \, \boldsymbol{\alpha}_i
\end{eqnarray}
\vskip 0.3 cm
The properties of the above defined elementary reflections are the following ones:
\vskip 0.3 cm
\begin{description}
  \item[1)] By definition, the reflection $\sigma_{\alpha_i}$ maps the corresponding basis vector into minus itself:
\begin{equation}\label{prop1}
\sigma_{\alpha_i}\left(\boldsymbol{\alpha}_i\right) \, = \, - \,\boldsymbol{\alpha}_i
\end{equation}
  \item[2)] Having defined the hyperplane $H_{\boldsymbol{\alpha}_i}$ orthogonal to the vector by the obvious condition
  $\mathbf{w}\in H_{\boldsymbol{\alpha}_i} \Leftrightarrow \kappa(\mathbf{w},\boldsymbol{\alpha}_i)=0$ the reflection $\sigma_{\boldsymbol{\alpha}_i}$
  leaves it invariant pointwise:
\begin{equation}\label{prop2}
\forall \mathbf{w}\in H_{\boldsymbol{\alpha}_i} \quad : \quad
\sigma_{\boldsymbol{\alpha}_i}(\mathbf{w}) \, = \, \mathbf{w}
\end{equation}
\item[3)] The relations to be satisfied by the group  generators are preserved:
  \begin{equation}\label{prop4}
    \left(\sigma_{\boldsymbol{\alpha}_i}\cdot\sigma_{\boldsymbol{\alpha}_j} \right)^{m_{i,j}} \, = \, \text{Id}
\end{equation}
\item[4)] The symmetric form $\kappa$ is invariant under all elementary reflections $\sigma_{\boldsymbol{\alpha}_i}$:
  \begin{equation}\label{prop3}
   \forall\mathbf{v},\mathbf{u} \,\in \, V \quad : \quad \kappa\left(\sigma_{\boldsymbol{\alpha}_i}\left(\mathbf{v}\right),
   \sigma_{\boldsymbol{\alpha}_i}
  \left(\mathbf{u}\right)\right)\, = \,\kappa\left(\mathbf{v},\mathbf{u}\right)\quad\quad i=1, \dots , r
  \end{equation}
\end{description}
The first two properties are the necessary one to name $\sigma_{\alpha_i}$ a generalized reflection.
The third property is the necessary and sufficient one to conclude that $\phi$, as defined through the map to the elementary generalized reflections,
 is a group homomorphism. Finally the 4th property guarantees that the bilinear form $\kappa$ is invariant under the image $\phi(\mathfrak{C})$ in
$\mathrm{Hom}(V,V)$ of the abstract Coxeter group.
\paragraph{\sc A crucial observation}
From the point of view of the consistency of definitions there is no obstacle to the possibility that the matrix element $m_{i,j}$ might
be equal to $\infty$.
In this case it means that the Coxeter group element $T_{ij} \equiv s_i\cdot s_j\in \mathfrak{C} $ has order infinity. This implies that the group
$\mathfrak{C}$ is discrete
yet not finite. The relevant point is that also the image $\phi(T_{ij})$ through the homomorphism induced by eq.(\ref{grammofono}), which
is an $r \times r$ matrix, is of order infinity. To see that it suffices in eq.(\ref{pernacito}) to interpret literally $\kappa_{ij} \, = \, -1$
if $m_{i,j} =\infty$.
\par
The above comment is very much relevant for our purposes, since some of the cases turned up to be relevant in data science applications concern Coxeter groups with one $m_{i,j} \, =\, \infty$.
\subsection{Coxeter Groups of rank \texorpdfstring{$3$}{3} and triangular groups}
\label{rank3triagroup}
\begin{figure}[htb]
\vskip 2cm
\centering
\begin{picture}(50,150)
\put(-85,157){$\Delta_{p,q,r} \quad:\quad $}\put(-30,160){\circle{10}}
\put(-25,160){\line(1,0){25}} \put(-15,165){$p$}
\put (5,160){\circle {10}} \put(5,155){\line (0,-1){25}}
\put(8,140){$q$}\put (5,125){\circle {10}} \put(1,127){\line(-1,1){28}}\put(-22,138){$r$}
\end{picture}
\vskip -3cm
\caption{\label{trenos} The Coxeter graph of the triangular groups $\Delta_{p,q,r}$ }
\end{figure}
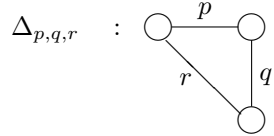

\begin{figure}[htb]
\begin{center}
\vskip 1cm
\includegraphics[width=90mm]{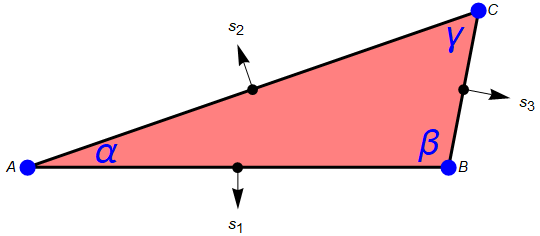}
\vskip 1cm
\caption{\label{triugolgrup} Conceptual visualization of a triangular group. In an ambient space that can be the sphere,
the Euclidean plane or the hyperbolic plane we consider a triangle with vertices $A,B,C$ and sides that are arcs of geodesics between the vertices.
The reflections with respect to the orthogonal vector of each side of the triangle are the generators $s_{1,2,3}$. The composition of two subsequent
reflections is a rotation of an angle $\frac{\pi}{m_{ij}}$.}
\end{center}
\end{figure}

A class of interesting Coxeter Groups, which is related with the tessellation schemes of constant curvature  spaces of dimensions $d=2$, namely
the sphere, the Euclidean plane and the hyperbolic plane, is provided by the rank $r=3$ groups $\Delta_{p,q,r}$ and by their index two subgroups,
composed only by those words that
have an even number of letters (the generators $s_i$). The relevant Coxeter diagram is shown in fig.\ref{trenos}.
Abstractly, the group presentation associated with the diagram of fig.\ref{trenos} is the following one:
\begin{equation}\label{trincagruppo}
  \Delta_{p,q,r}\, =\, \langle s_1,\,s_2,\,s_3 \,\mid \, s_1^2 \,= \, s_2^2 \, = \, s_3^2 \, = \, (s_1s_2)^p \, = \, (s_2s_3)^q \, =\, (s_1 s_3)^r \,
  = \, \mathbf{e}\rangle
\end{equation}
Naming $\mathrm{A}\equiv T_{12}=s_1s_2$, $\mathrm{B}\equiv T_{23}=s_2 s_3$, we obtain the presentation of the even triangular  subgroup
$\Delta_{p,q,r}^+\subset \Delta_{p,q,r}$ as follows:
\begin{equation}\label{barbogruppo}
  \Delta_{p,q,r}^+\, =\, \langle \mathrm{A},\, \mathrm{B} \,\mid \,  \mathrm{A}^p \, = \, \mathrm{B}^q \, =\, (\mathrm{A} \mathrm{B})^r \,
  = \, \mathbf{e}\rangle
\end{equation}
Graphically, the conception is illustrated in fig.\ref{triugolgrup}, where we show a generic triangle with three internal angles
$\alpha,\beta,\gamma$, that, by assumption, must be  integer fractions of $\pi$.
As explained in the figure caption, the rotation angle is just the interior angle between side $i$ and side $j$ in the triangle. The sum of these
angles is therefore:
\begin{equation}\label{angsumma}
  \alpha +\beta +\gamma \, = \, \pi \, \left( \frac{1}{p} \, + \, \frac{1}{q} \, + \, \frac{1}{r} \right) \quad ; \quad p,q,\in \mathbb{N}
\end{equation}
According to geometry, in the positive, null and negative curvature case we respectively have:
\begin{equation}\label{parlapiano}
  \begin{array}{|ccc|c|c|}
  \hline
    \frac{1}{p} \, + \, \frac{1}{q} \, + \, \frac{1}{r} & > & 1 & \mathbb{S}^2 & \text{sphere} \\
    \hline
    \frac{1}{p} \, + \, \frac{1}{q} \, + \, \frac{1}{r} & = & 1 & \mathbb{R}^2 & \text{Euclidean plane} \\
    \hline
    \frac{1}{p} \, + \, \frac{1}{q} \, + \, \frac{1}{r} & < & 1 & \mathbb{H}^2 & \text{Hyperbolic plane} \\
    \hline
  \end{array}
\end{equation}
Recalling the ADE classification of discrete subgroups (actually finite subgroups) of the rotation group $\mathrm{SO(3)}$ (that coincides with the ADE
classification of semisimple Lie Algebras (McKay correspondence))\footnote{For a comprehensive exposition of the ADE classification see
\cite{advancio} and \cite{fre2023book}} we recognize that the first Diophantine inequality is the classifying one for discrete, finite rotations
groups, whose solution always implies at least one of the three integer number to be $2$. Observing the cycle symmetry of the Coxeter diagram in fig.
\ref{trenos}, we see that the naming of the lines is irrelevant: hence we decide that the fixed order $r=2$, required by spherical symmetry
corresponds to the line $3\to 1$. This leads us to consider the Coxeter groups $\Delta_{p,q,2}$, whose Coxeter diagram is displayed
in fig.\ref{ractagruppa}.
\begin{figure}[htb]
\vskip 2cm
\centering
\begin{picture}(50,150)
\put(-85,157){$\Delta_{p,q,2} \quad:\quad $}\put(-30,160){\circle{10}}
\put(-25,160){\line(1,0){25}} \put(-15,165){$p$}
\put (5,160){\circle {10}} \put(10,160){\line (1,0){25}}
\put(20,165){$q$}\put (40,160){\circle {10}}
\end{picture}
\vskip -5cm
\caption{\label{ractagruppa} The Coxeter graph of the triangular groups $\Delta_{p,q,2}$. This class of groups contains all the finite rotations
groups of the ADE classification, but also others that are not spherical, but rather hyperbolic.}
\end{figure}
This class includes as even subgroups $\Delta^+_{p,q,2}$, all the finite rotation groups according with the summary recalled below:
\begin{equation}\label{panettonemotta}
  \begin{array}{|ccc|c|ccc|}
  \hline
     \Delta^+_{p,p,2} & = & \mathbb{Z}_p & \text{cyclic group} & \left|\Delta^+_{p,p,2}\right | & =& p  \\
     \hline
     \Delta^+_{p,2,2} & = & \mathrm{Dih}_p & \text{Dihedral group} & \left| \Delta^+_{p,2,2}\right | & = & 2 p \\
      \hline
     \Delta^+_{3,3,2} & = & \mathrm{T}_{12} & \text{Tetrahedral group} & \left| \Delta^+_{3,3,2}\right | & = & 12\\
      \hline
 \Delta^+_{4,3,2} & = & \mathrm{O}_{24} & \text{Octahedral group} & \left| \Delta^+_{4,3,2}\right | & = & 24\\
 \hline
 \Delta^+_{5,3,2} & = & \mathrm{I}_{60} & \text{Icosahedral group} & \left| \Delta^+_{5,3,2}\right | & = & 60\\
 \hline
   \end{array}
\end{equation}
Besides the above-mentioned \textit{spherical} even Coxeter groups, the triangular class defined by the diagram in fig. \ref{ractagruppa} contains
many more examples that are generically hyperbolic. Note that the condition $r=2$ implies that, referencing fig. \ref{triugolgrup}, the triangle there displayed is a \textit{rectangular triangle}, since the angle $\gamma$ is $\pi/2$.
\subsection{Study of the triangular group \texorpdfstring{$\Delta_{p,3,2}^+$}{Dp32+} and related tessellations of the hyperbolic plane}
\label{Deltap32}
\begin{figure}[htb!]
\vskip 2cm
\centering
\begin{picture}(50,150)
\put(-85,157){$\Delta_{p,3,2} \quad:\quad $}\put(-30,160){\circle{10}}
\put(-25,160){\line(1,0){25}} \put(-15,165){$p$}
\put (5,160){\circle {10}} \put(10,160){\line (1,0){25}}
\put (40,160){\circle {10}}
\end{picture}
\vskip -5cm
\caption{\label{classona} The Coxeter graph of the triangular groups $\Delta_{p,3,2}$. This class of groups contains all the finite rotation
groups that are symmetries of the five platonic solids,  a pair of exceptional spherical groups, plus an infinite series of hyperbolic groups.}
\end{figure}
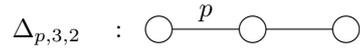

In the spirit of our last observation, we focus next on a narrower subclass of triangular groups $\Delta_{p,q,2}^+$, namely the one where $q=3$,
corresponding to the Coxeter diagram of fig.\ref{classona}.
This class is particularly interesting for the following reasons:
\begin{description}
  \item[A)] It allows a general treatment that illustrates the general concepts introduced in section \ref{georepcox} and permits inside the class
   to distinguish spherical, from hyperbolic and degenerate cases.
  \item[B)] It contains the three finite rotation groups that are symmetry of platonic solids.
  \item[C)] It contains a notable class of tessellation groups of the hyperbolic plane in terms of geodesic triangles such that in each vertex meet
  $p$ lines and each vertex is surrounded by $p$ triangular tiles. Alternatively one can consider the dual tessellation in which the tile are
  $p$-polygons and each vertex is surrounded by $3$ $p$-polygons.
  \item[D)] Choosing the limiting value $p=\infty$, that is perfectly admissible, one obtains the infinite discrete group $\Delta_{\infty,3,2}^+$
  which generates the tessellation of the hyperbolic plane in terms of \textbf{apeirogons} (see further on), which is the
  \textit{seemingly tree-structure} cheered by the authors of \cite{francesi1,francesi2,francesi3}. Note that, were we to consider the
  group $\Delta_{\infty,q,2}^+$
  we would obtain a hyperbolic tessellation in which $q$ apeirogons meet in each vertex.
\end{description}
\par
Hence let us start the analysis of $\Delta_{p,3,2}^+$ by introducing the following notation:
\begin{equation}\label{muggendo}
  \mu \, \equiv \, \cos\left[\frac{\pi}{p} \right] \quad ; \quad p\in \mathbb{N}
\end{equation}
The vector space $V$ advocated in eq.(\ref{rappezzo}) is $3$-dimensional and we can realize the homomorphism (\ref{homorfo}) of the Coxeter
group $\mathfrak{C} = \Delta_{p,3,2}$ into $\mathrm{Hom}(V,V)$ by introducing the three
reflection generators:
\begin{equation}\label{gennis123}
\begin{array}{ccccccccccc}
    s_1 & = & \left(
\begin{array}{c|c|c}
 -1 & 2 \mu  & 0 \\
 \hline
 0 & 1 & 0 \\
 \hline
 0 & 0 & 1 \\
\end{array}
\right) & ; & s_2 & = & \left(
\begin{array}{c|c|c}
 1 & 0 & 0 \\
 \hline
 2 \mu  & -1 & 1 \\
 \hline
 0 & 0 & 1 \\
\end{array}
\right) &;&
    s_3 & = & \left(
\begin{array}{c|c|c}
 1 & 0 & 0 \\
 \hline
 0 & 1 & 0 \\
 \hline
 0 & 1 & -1 \\
\end{array}
\right)
  \end{array}
\end{equation}
The generators of the triangular even subgroup $\Delta_{p,3,2}^+ \subset \Delta_{p,3,2}$ are canonically defined as:
\begin{eqnarray}\label{ABCp32}
   A \,=\, s_1\cdot s_3 & = & \left(
\begin{array}{c|c|c}
 4 \mu ^2-1 & -2 \mu  & 2 \mu  \\
 \hline
 2 \mu  & -1 & 1 \\
 \hline
 0 & 0 & 1 \\
\end{array}
\right) \quad ; \quad B \, = \, s_2\cdot s_3 \, = \, \left(
\begin{array}{c|c|c}
 1 & 0 & 0 \\
 \hline
 2 \mu  & 0 & -1 \\
 \hline
 0 & 1 & -1 \\
\end{array}
\right) \nonumber\\
    C \, = \, s_1\cdot s_3 & = & \left(
\begin{array}{c|c|c}
 -1 & 2 \mu  & 0 \\
 \hline
 0 & 1 & 0 \\
 \hline
 0 & 1 & -1 \\
\end{array} \right)
\end{eqnarray}
One easily verifies that, independently from the value of $\mu$ one has:
\begin{equation}\label{perducco}
  B^3\, = \, C^2 \, = \, \mathbf{1}_{3\times 3} \quad ;\quad A\cdot B \, = \, C
\end{equation}
while
\begin{equation}\label{occudrep}
  A^p \, = \, \mathbf{1}_{3\times 3} \quad ; \quad \text{iff} \quad \mu \, \equiv \, \cos\left[\frac{\pi}{p} \right]\quad ; \quad p\in \mathbb{N}
\end{equation}
Hence with the position (\ref{muggendo}), the presentation (\ref{barbogruppo}) of the triangular group  $\Delta_{p,3,2}^+$ is explicitly
realized in terms of the $3\times 3$ matrices of eq.(\ref{ABCp32}). On the other hand the symmetric bilinear form (\ref{pernacito}) has
the following explicit appearance:
\begin{equation}\label{kappaformetta}
  \kappa \, = \, \left(
\begin{array}{c|c|c}
 1 & -\mu  & 0 \\
 \hline
 -\mu  & 1 & -\frac{1}{2} \\
 \hline
 0 & -\frac{1}{2} & 1 \\
\end{array}
\right)
\end{equation}
and one easily verifies that:
\begin{equation}\label{ortaggi}
  A^T\cdot\kappa \cdot A \, = \,  B^T\cdot\kappa \cdot B \, = \, \,  C^T\cdot\kappa \cdot C \, = \, \kappa
\end{equation}
This guarantees that the group $\phi(\Delta_{p,3,2}^+)$ generated by $A,B$ as defined in eq.(\ref{ABCp32}) is a discrete subgroup of the
group $\mathrm{SO}(\kappa)$ which leaves invariant the bilinear form $\kappa$. Whether this latter group is isomorphic to $\mathrm{SO(3)}$
or to $\mathrm{SO(1,2)}$ depends on the signature of the matrix $\kappa$, provided the latter is not degenerate.
Calculating the determinant of $\kappa$ we find:
\begin{equation}\label{principino}
  \text{Det}(\kappa)\, = \, \frac{3}{4}-\mu ^2
\end{equation}
Hence the only degenerate case occurs for $p=6$.
\paragraph{\sc The degenerate case}
In the degenerate case $p=6$ we find that the following change of basis matrix
\begin{equation}\label{carsima}
S \, = \, \left(
\begin{array}{ccc}
 \frac{\sqrt{3}}{4} & -\frac{1}{2} & \frac{\sqrt{\frac{3}{2}}}{2} \\
 -\frac{1}{2} & 0 & \frac{1}{\sqrt{2}} \\
 \frac{1}{4} & \frac{\sqrt{3}}{2} & \frac{1}{2 \sqrt{2}} \\
\end{array}
\right)
\end{equation}
produces the following transformation on the form $\kappa_{6}=\kappa(\mu=\cos(\pi/6)$:
\begin{equation}\label{pudrus}
 \tilde{ \kappa} \, = \, S^T \, \kappa_{6} \,S \, = \,\left(
 \begin{array}{ccc}
  1 & 0 & 0 \\
  0 & 1 & 0 \\
  0 & 0 & 0 \\
  \end{array}
  \right)
\end{equation}
while we get
\begin{eqnarray}
  \tilde{A}^T &=& S^T \, A_{6}^T \, S \, = \, \left(
\begin{array}{cc|c}
 \frac{1}{2} & -\frac{\sqrt{3}}{2} & \frac{7}{2 \sqrt{2}} \\
 \frac{\sqrt{3}}{2} & \frac{1}{2} & \frac{\sqrt{\frac{3}{2}}}{2} \\
 \hline
 0 & 0 & 1 \\
\end{array}
\right) \quad ; \quad
  \tilde{B}^T \, = \, S^T \, B_{6}^T\,S  \, \, = \, \left(
\begin{array}{cc|c}
 -\frac{1}{2} & -\frac{\sqrt{3}}{2} & \frac{1}{2 \sqrt{2}} \\
 \frac{\sqrt{3}}{2} & -\frac{1}{2} & -\frac{3 \sqrt{\frac{3}{2}}}{2} \\
 \hline
 0 & 0 & 1 \\
\end{array}
\right) \nonumber \\
  \tilde{C}^T &=& S^T \, C_{6}^T \, S\, \, = \, \left(
\begin{array}{cc|c}
 -1 & 0 & -\sqrt{2} \\
 0 & -1 & 0 \\
 \hline
 0 & 0 & 1 \\
\end{array}
\right)
\end{eqnarray}
As one sees the generators are now mapped not into the orthogonal group, rather into the Euclidean group $\mathbb{E}^2\, = \,\mathrm{ISO(2)}$
represented by border matrices. Such a setup for the two dimensional space $\mathbb{R}^2$ is discussed
and illustrated in \cite{fre2023book}. In conclusion  the triangular group $\phi(\Delta_{6,3,2}^+)$ is actually a space group in $\mathbb{R}^2$ and it
produces a tessellation of the ordinary plane by means of triangular tiles that arrange into hexagons surrounding each point of the lattice.
\par
Note to this effect that $1/6+1/3+1/2 \, = \, 1$ so that indeed in this case the triangular group is one of the wallpaper groups
producing $\mathbb{R}^2$ tessellations.
\paragraph{\sc Hyperbolic groups}
For $\mathbb{N} \ni p \geq 7$ the triangular groups $\Delta^+_{p,3,2}$ are all hyperbolic since
\begin{equation}\label{brenno}
  \frac{1}{p} \, + \, \frac{1}{3} \, + \, \frac{1}{2} \, < \, 1 \quad : \quad \text{if}\quad  p \geq 7
\end{equation}
Indeed the eigenvalues of the matrix (\ref{kappaformetta}) are:
\begin{equation}\label{autovalloni}
  \text{eigenvalues of $\kappa$} \, = \, \left\{
\begin{array}{c}
 1 \\
 \null\\
 \frac{1}{2} \left(2-\sqrt{4 \mu ^2+1}\right) \\
 \null\\
 \frac{1}{2} \left(\sqrt{4 \mu ^2+1}+2\right) \\
\end{array}
\right.
\end{equation}
and in order to have positive signature of the invariant metric the condition is:
\begin{equation}\label{pinocchioso}
  4\,\mu^2 +1 <2 \quad \Rightarrow \quad p\geq 7
\end{equation}
In all cases verifying the above condition the triangular group $\Delta^+_{p,3,2}$ is a subgroup of $\mathrm{SO(1,2)}$ and, by means of an appropriate
change of basis $S$ we can reduce the matrices in eq. (\ref{ABCp32}) to respect $\kappa$  in the standard form utilized within the general formalism
of the PGTS paradigm as developed in \cite{pgtstheory}:
\begin{equation}\label{etuc}
 S^T \, \kappa \, S \, = \, \eta \, = \,\left(
                  \begin{array}{ccc}
                   0 & 0 & 1 \\
                    0 & 1 & 0 \\
                    1 & 0 & 0 \\
                  \end{array}
                \right)
\end{equation}
We have derived the explicit form of the matrix $S$ for all values of $\mu$:
\begin{equation}\label{cedrino}
S \, = \,  \left(
\begin{array}{c|c|c}
 -\frac{2 \mu }{\sqrt{\left(8 \mu ^2+2\right) \left(\sqrt{4 \mu
   ^2+1}-2\right)}}-\frac{1}{\sqrt{8 \mu ^2+2}} & \frac{2 \mu }{\sqrt{\left(4 \mu
   ^2+1\right) \left(\sqrt{4 \mu ^2+1}+2\right)}} & \frac{2 \mu }{\sqrt{\left(8 \mu
   ^2+2\right) \left(\sqrt{4 \mu ^2+1}-2\right)}}-\frac{1}{\sqrt{8 \mu ^2+2}} \\
   \hline
 -\frac{1}{\sqrt{2} \sqrt{\sqrt{4 \mu ^2+1}-2}} & -\frac{1}{\sqrt{\sqrt{4 \mu ^2+1}+2}}
   & \frac{1}{\sqrt{2} \sqrt{\sqrt{4 \mu ^2+1}-2}} \\
   \hline
 \frac{\sqrt{2}}{\sqrt{\frac{1}{\mu ^2}+4}}-\frac{1}{\sqrt{\left(8 \mu ^2+2\right)
   \left(\sqrt{4 \mu ^2+1}-2\right)}} & \frac{1}{\sqrt{\left(4 \mu ^2+1\right)
   \left(\sqrt{4 \mu ^2+1}+2\right)}} & \frac{1}{\sqrt{\left(8 \mu ^2+2\right)
   \left(\sqrt{4 \mu ^2+1}-2\right)}}+\frac{\sqrt{2}}{\sqrt{\frac{1}{\mu ^2}+4}} \\
\end{array}
\right)
\end{equation}
Because of the isomorphism $\mathrm{SO(1,2)}\sim \mathrm{SL(2,\mathbb{R})}$, explicitly realized  by the quadratic relation in eq.s(7.11-7.14) of
\cite{pgtstheory} we might try to solve it in order to determine the $2\times 2$ matrices $\mathcal{A},\mathcal{B},\mathcal{C}\in\mathrm{SL(2,\mathbb{R})}$
representing  the generators (\ref{ABCp32}) inside $\mathrm{SL(2,\mathbb{R})}$. This is very much laborious, yet it is not the end of the story,
since, in order to construct explicitly the tessellation one has  to determine also a starting point in the upper complex plane,
that is either invariant with respect to the generator $\mathcal{A}$ or to the generator $\mathcal{B}$.
In the first case the tessellation will be made of triangular tiles, while, in the second it will be its dual, made of $p$-gons.
The determination of such invariant  point is even more laborious
than the determination of the
$\mathrm{SL(2,\mathbb{R})}$ pre-images $\mathcal{A},\mathcal{B},\mathcal{C}$ of the generators $S^{-1}\,A\,S$, $S^{-1}\,B\,S$, $S^{-1}\,C\,S$.
\par
Fortunately, since the local isomorphism $\mathrm{SO(1,2)}\sim \mathrm{SL(2,\mathbb{R})}$ is guaranteed a priori, we can proceed differently, by
choosing a priori the origin $\mathit{i}$ as the invariant point, and choosing for $\mathcal{A}$ the standard form of the generator of a
$\mathbb{Z}_p\subset \mathrm{SL(2,\mathbb{R})}$ which stabilizes $\mathit{i}$, or choosing for $\mathcal{B}$ the standard form of
$\mathbb{Z}_3\subset \mathrm{SL(2,\mathbb{R})}$ subgroup  which once again stabilizes the point $\mathit{i}$.
\par In this way we reduce the problem to that of determining the other generator $\mathcal{B}$, or $\mathcal{A}$
such that all the defining relations are satisfied.
\par
In the sequel we  show the explicit construction of the tessellation group
$\Delta^+_{8,3,2}$ which is mentioned episodically in the literature \cite{howarth2023,cuocogiuseppe}, yet in no place we were able to search and inspect, is presented in full detail and explicitly.
\par
Most of our construction follows analogous steps in the classical construction of $\Delta^+_{7,3,2}$, yet there are considerable
differences at the end point.
\paragraph{Hurwitz Theorem and Hurwitz Riemann surfaces.}
The classical  $\Delta^+_{7,3,2}$ tessellation is  related with Hurwitz theorem on the automorphism
group of genus $g$ compact Riemann surfaces and with their uniformization. Indeed it was shown that every Hurwitz Riemann surface $\Sigma_H$
has an automorphism group that is the quotient of the infinite tiling group $\Delta^+_{7,3,2}$ with respect to a suitable Fuchsian normal subgroup thereof $\Gamma_{\Sigma_H} \subset \Delta^+_{7,3,2}$ of finite index $N\, = \, 84 (g-1)$ where $g$ is the genus of $\Sigma_H$.
\paragraph{The general paradigm for uniformization of Riemann surfaces with $g \geq 2$.}
All $\Delta^+_{p,3,2}$  (with $p > 7$) tiling are related with the uniformization of Riemann surfaces that are not Hurwitz through the very same mechanism.
\par
The paradigm is encoded in the following exact sequence:
\begin{equation}\label{equafun1}
{\text{Id}\overset{\iota }{\longrightarrow }\Gamma \overset{\iota }{\longrightarrow }\Delta ^+_{p,3,2}\overset{\pi }{\longrightarrow
}\text{GD}\equiv \frac{\Delta ^+_{p,3,2}}{\Gamma }\overset{\pi }{\longrightarrow }\text{Id}}
\end{equation}
where $\iota$ denotes the immersion map and $\pi $ the projection map. Following for instance the conceptual illustration of the Ph.D. theses
\cite{vogo,cuocogiuseppe} and of the paper \cite{Chutassella} we know that the Fuchsian normal subgroup $\Gamma$ admits a fundamental domain.
\begin{definizione}
$\Gamma \subset \Delta ^+_{p,3,2} \subset \mathrm{PSL(2,\mathbb{R})}$  is named a Fuchsian group of the first kind, if its elements are all hyperbolic (\textit{i.e.} $\forall \gamma$ we have $\mathrm{Tr} (\gamma) >2$), if $\Gamma$ acts discontinuously on
 $\mathbb{H}^2$ and all the limit points of orbits of $\Gamma$ are on the boundary $\partial \mathbb{H}^2$. Said differently the order of each element $\gamma$ is infinite.
\end{definizione}
\begin{definizione}
A subset $\mathcal{F}_\Gamma \subset \mathbb{H}^2$ is a fundamental domain of a Fuchsian group of the first kind if it is
a finite area, convex, connected region, such that no two points $z_{1,2}\in
\mathcal{F}_\Gamma$ are connected to each other by the action of any $\gamma \in \Gamma$, while every point
$z\in (\mathbb{H}^2 -\mathcal{F}_\Gamma)$ in the complement of the fundamental domain can be mapped into $\mathcal{F}_\Gamma$
by a suitable element $\gamma_z \in \Gamma$. Furthermore the images $\gamma( \mathcal{F}_\Gamma)$ through
elements $\gamma \in \Gamma$ constitute a tessellation of $\mathbb{H}^2$.
\end{definizione}
Furthermore the boundary $\partial\mathcal{F}_\Gamma$ of the fundamental domain $\mathcal{F}_\Gamma$ must be composed  of a certain number
$n_\Gamma$ of geodesic arcs $\ell_{p=1,\dots , n_\Gamma}$ that meet pairwise into $n_\Gamma$ vertices $v_p$ mapped one into the next one by as
many elements  $\mho_{p=1,\dots , n_\Gamma}$ of the group $\Gamma$ which provide  a set of generators for it  since each of them has its own
inverse in the same set.
\paragraph{Tessellations are essentially the same thing as Riemann surfaces.}
In the applications  we are concerned with, what matters for us are the tessellations since we need to assign data to the tiles of some regular
tessellation\footnote{Tipically data form a discrete finite set, so that we will note use all the tiles rather only those pertaining to a finite
compact portion of the hyperbolic plane,  yet those tiles must be arranged regularly as the points of lattice in the Euclidean $2$-space where
they are originally defined.} yet a general theorem  stating  that every finite area hyperbolic surface $S$ can be represented as the
quotient of $\mathbb{H}^2$ by the action of a Fuchsian group of the first kind, \textit{i.e.} $S\equiv \mathbb{H}^2/\Gamma_S$, reveals that the
classification of possible tessellations amounts to the same thing as the classification of uniformized Riemann surfaces with non-trivial
automorphism groups, at least as  long as the tiles are required to be smooth and regular.
\par
In this context a very much relevant computational tool is provided by the following observation. Since the Fuchsian group acts discontinuously,
the fundamental domain $\mathcal{F}_\Gamma$ necessarily shares\cite{cuocogiuseppe} a component of its boundary with a finite
number $n_\Gamma$ of its images through as many elements $\mho_{p} \in \Gamma$ ($p=1,\dots , n_\Gamma$):
\begin{equation}\label{latoni}
  \partial\mathcal{F}_\Gamma\bigcap \mho_{p}\left(\mathcal{F}_\Gamma\right) \, \equiv \, \ell_{p} \neq \varnothing
\end{equation}
Since any two curves $\ell_{p,q}$ are mapped one into the other by the Fuchsian group element $\mho_{p} \cdot \mho_{q}^{-1}$ which is an isometry,
all the $\ell_{p}$  are necessarily geodesic arcs. Furthermore we have:
\begin{equation}\label{nuovafrontiera}
  \partial\mathcal{F}_\Gamma \, = \, \bigcup_{p=1}^{n_\Gamma} \, \ell_p
\end{equation}
Hence we arrive at the conclusion that the fundamental domain $\mathcal{F}_\Gamma$ is surrounded by a necklace of $n_\Gamma$ copies of itself $\mathcal{T}_p \,\equiv\,\mho_{p}\left(\mathcal{F}_\Gamma\right)$, (the first $n_\Gamma$ tiles of the tessellation) with which it shares the side  $\ell_p$. The group elements $\mho_{p}$ can be chosen as generators
of the Fuchsian group. Furthermore since we can always choose the fundamental domain in such a
way that the origin $\mathbf{e} \equiv\{0,0\}$, in the disk model, or its Cayley equivalent $C(\mathbf{e}) \equiv \{0,1\}$ in the upper
half plane model, belongs to it\footnote{Note that, thanks to the metric equivalence of $\mathbb{H}^2$ with a solvable Lie group $\mathcal{S}_{\mathbb{H}^2}$ as provided by the PGTS theory, the origin has an intrinsic meaning, namely it is the neutral element of $\mathcal{S}_{H^2}$} the necklace of tiles $\mathcal{T}_p $ includes a set of vertices:
\begin{equation}
  \hat{v}_{p} \, \equiv \, \mho_{p}(\mathbf{e} )
\end{equation}
which joined together by geodesic arcs belong to a tessellation of the hyperbolic plane, yet are not vertices and sides  of the
tessellation associated with the Riemann surface $\Sigma_\Gamma$,  rather they belong to the dual one. Indeed, being the images
of an interior point of the fundamental domain the points $\hat{v}_{p}$ cannot be on the boundary $\partial\mathcal{F}_\Gamma$.
As later on  we show explicitly  in the case of  Bolza surface, the vertices  $\hat{v}_{p}$ of the dual tessellation
are anyhow useful since they have the same angular coordinate as the true vertices ${v}_{p}$ of the boundary
$\partial\mathcal{F}_\Gamma$, namely the meeting point of  $\ell_p$ with its subsequent $\ell_{p+1}$ for each
$p$ (see eq.(\ref{nuovafrontiera})), although at larger  radial coordinate $\rho$.
\par
The main task therefore is that of determining the correct critical radius $R_c$ at which the vertices of $\partial\mathcal{F}_\Gamma$ are located.
\paragraph{The Dirichelet closed set.} Instrumental for the determination of the $R_c$ value  and, more generally, for the full
construction of the fundamental domain, is its identification with the Dirichlet domain centered in $\mathbf{e}$.
\begin{definizione}
Let $w\in\mathbb{ H}^2$ be a point that has trivial stabilizer in the Fuchsian group $\Gamma$. The Dirichlet domain centered at $w$ is defined as follows:
\begin{equation}\label{dirichello}
  \mathcal{D}(w)= \left\{\, z \in \mathbb{H}^{2} \, | \, d(w,z) < d(w,\gamma (z))\quad\forall \gamma \in \Gamma \right\}
\end{equation}
\end{definizione}
where $d (p_1, p_2)$ denotes the distance of the point separating the two points $p_{1,2}$, which is the length of the unique
geodesic arc joining them.  As it was explained at length in \cite{pgtstheory}, all non-compact symmetric spaces $\mathrm{U/H}$ are metrically equivalent to a \textit{normed solvable Lie group manifold}  $\mathcal{S}_{\mathrm{U/H}}$, the hyperbolic plane $\mathbb{H}^2$ being simply the smallest and simplest example in a vast class. In all these manifolds the unique distance function
is just given by:
\begin{equation}\label{cremagliera}
  d(p_1,p_2) \, \equiv \, \mathrm{N}\left( \mathit{s}_{p_1}^{-1}\cdot \mathit{s}_{p_2} \right)
\end{equation}
$\mathrm{N}(\mathit{s})=\mathrm{N}(\mathit{s}^{-1})$ denoting the norm of any solvable Lie group element $\mathit{s}\in \mathcal{S}_{\mathrm{U/H}}$ and
$\mathit{s}_{p_{1,2}}$ the solvable group elements respectively corresponding to the points $p_{1,2} \in \mathrm{U/H}$.
\par As shown in thee literature \cite{Chutassella,cuocogiuseppe,buserbook,vogo}, the fundamental domain can be identified
with the Dirichlet domain centered at the origin namely at the neutral element of the solvable Lie Group.
\begin{equation}\label{cardospinoso}
  \mathcal{F}_\Gamma \, = \, \mathcal{D}(\mathbf{e})
\end{equation}
In other words the Fuchsian group acts on the normed solvable Lie group as follows:
\begin{equation}\label{putrella}
  \forall \gamma \in \Gamma  \quad / \quad \gamma \, : \, \,\mathcal{S}_{\mathrm{U/H}} \longrightarrow  \mathcal{S}_{\mathrm{U/H}}
\end{equation}
and we are interested in comparing the norm of each element $\mathit{s}\in \mathcal{S}_{\mathrm{U/H}}$ with its image under the elements $\gamma \in \Gamma$. For each element $\gamma $ of the Fuchsian group we define its \textit{norm difference function} as follows:
\begin{equation}\label{ndiffun}
  \mathrm{Nd}_{\gamma }(\mathit{s}) \equiv \mathrm{N}(\gamma (\mathit{s})) - \mathrm{N}(\mathit{s})
\end{equation}
In terms of these notions, the fundamental domain of the Fuchsian group can be seen as the positivity domain of the norm difference function for all elements of the Fuchsian group:
\begin{equation}
  \mathcal{F}_\Gamma \, = \, \left\{ s \in \mathcal{S}_{\mathrm{U/H}} \, \mid \, \mathrm{Nd}_{\gamma }(\mathit{s})>0 \quad\quad \forall \gamma \in \Gamma \right\}
\end{equation}
In view of the partial ordering of the Fuchsian group in terms of trace length spectrum (see section \ref{straccioni} ) it suffices to impose the positivity of the norm difference functions associated with the generators $\mho_p$ namely we can also set the following
\begin{statement}
\label{defioperativa}
Let  $\mho_p$  $(p=1,\dots, n_\Gamma)$ be a standard set of generators of the Fuchsian subgroup $\Gamma$  associated
with the vertices $\hat{v}_p$ of the dual tessellation, then the fundamental domain $\mathcal{F}_\Gamma$ is obtained by the following intersection
\begin{equation}
  \mathcal{F}_\Gamma \, = \, \left\{ s \in \mathcal{S}_{\mathrm{U/H}} \, \mid \, \mathrm{Nd}_{\mho_p}(\mathit{s})>0 \quad\quad p=1,\dots,n_\Gamma\right\}
\end{equation}
\end{statement}
The last reformulation of the definition of $\mathcal{F}_\Gamma$ is the most practical  one and we shall utilize it
in the determination of the fundamental domain
for the Bolza surface.
\paragraph{The Fundamental Domain of the Klein Quartic}.
The very profound and time-honoured Hurvitz theorem \cite{hurvizzo} states that the order of the automorphism group of a genus $g$ Riemann surface is upper bounded by the following equation:
\begin{equation}\label{hurango}
  \mid \mathrm{AutG\left( \Sigma_g \right)} \mid \, \leq \, 84(g-1)
\end{equation}
When a Riemann surface $\Sigma_g^H$ saturates the bound in eq.(\ref{hurango}), it is called a \textit{Hurwitz surface} and the corresponding group $\mathrm{AutG\left( \Sigma_g^H \right)}$ is called a \textit{Hurwitz group}.
The genus $g=3$ Klein quartic \cite{kelinquart} is the first Hurwitz surface, and the corresponding automorphism group of order 168, abstractly corresponding to $\mathrm{PSL(2,\mathbb{Z}_7)}$, is the smallest Hurwitz group.  A refinement of Hurwitz's theorem states that all possible Hurwitz surfaces $\mathrm{\Sigma^H}$ are associated with suitable finite index Fuchsian subgroups $\Gamma(\Sigma^H) \subset  \Delta^+_{7,3,2}$.
The Fuchsian Group $\Gamma_{Klein} \subset \Delta^+_{7,3,2}$ has a fundamental domain with $14$ vertices and sides, corresponding to as many generators.
\paragraph{The  $\Delta^+_{8,3,2}$ tessellations} Given Hurwitz theorem, the regular $\Delta^+_{8,3,2}$ tilings
correspond to non-Hurwitz surfaces and the quotient groups$\Delta^+_{8,3,2}/\Gamma(\Sigma)$ are typically non-Hurwitz groups, yet they might be used to codify graph structures where the coordination numbers of lines entering vertices is not given by the primes $3,7$ rather by $3$ and $8$ or submultiples of the latter.
For the case of picture pixels that, thought abstractly, are among the goals of this paper, we find that within the scope of  $\Delta^+_{8,3,2}$ tilings that associated
with the Bolza surface \cite{bolzano,cuocogiuseppe} can be arranged in such a way as to have the right coordination numbers.
\section{\texorpdfstring{$\Delta^+_{8,3,2}$}{D832+} tilings of the Hyperbolic Plane}
\label{deltotto}
Summarizing the previous discussion, just as it happens in the case of $\Delta^+_{7,3,2}$, there is not just one $\Delta^+_{8,3,2}$ tiling, but rather several, depending on the choice of a normal Fuchsian subgroup $\Gamma \subset \Delta^+_{8,3,2}$ of finite index $n[\Gamma] < \infty $. Correspondingly the associated fundamental domain $\mathcal{F}_\Gamma$, whose copies through elements  $\gamma \in \Gamma$ constitute the tiles of a tessellation of the hyperbolic plane $\mathbb{H}^2$, is the uniformization of a suitable, smooth, orientable Riemann surface $\Sigma_\Gamma$ of appropriate genus $g$. Furthermore the finite group quotient $\mathrm{GD}_{n[\Gamma]} \, = \,\Delta^+_{8,3,2}/\Gamma$ is the automorphic group of the corresponding Riemann surface $\Sigma_\Gamma$.
\par
In this paper we consider the Fuchsian subgroups providing the uniformization of two Riemann surfaces, respectively  named
the \textit{Fermat Quartic} and the \textit{Bolza surface}.
\par
The algebraic equations defining the two surfaces are as follows.
 The  Fermat Quartic $\Sigma^{FQ}$ is the  locus  in \(\mathbb{P}^2\) defined by the following quartic constraint:
 \begin{equation}\label{fermaquattro}
   0 = X_1^4+X_2^4+X_3^4
\end{equation}
and it has genus $g=3$. As we are going to see the corresponding Fuchsian subgroup $\Gamma_{16}\subset\Delta^+_{8,3,2}$
has 16 generators  and it has index $96$. The automorphism group of the Fermat Quartic, that we name the group $\mathrm{GD}_{96}$ is solvable and admits the following chain of normal subgroups:
\begin{equation}\label{normalsubbo2}
\mathrm{GD_{96}} \rhd \mathrm{GD_{48}} \rhd \mathrm{GD_{16}} \sim \mathbb{Z}_4 \times \mathbb{Z}_4
\end{equation}
\par
The second Fuchsian subgroup $\Gamma_{8}\subset\Delta^+_{8,3,2}$ investigated in this paper is the one which uniformizes the Bolza surface of genus $g=2$. A general theorem states that all genus $g=2$ Riemann surfaces are hyperelleptic.
Generically a hyperelliptic  surface of genus $g$ is described by an algebraic equation  of the following form:
\begin{equation}\label{hyperelli}
  w^2=\prod _{a=2}^{2g+2} \left(z-\lambda _a\right)=\mathcal{P}_{2g+2}(z)
\end{equation}
$\lambda _a$ being the $2g+2$ roots of a generic order $2g+2$ polynomial. If one of the roots of $\mathcal{P}_{2g+2}(z)$ goes to infinity, the polynomial degenerates into a polynomial of degree 2g+1. This is the case of the Bolza surface which is described by the following equation (see \cite{bolzano,cuocogiuseppe} and all the more recent references quoted in the latter):
\begin{equation}\label{bolzadue}
  w^2=z\left(z^4-1\right)
\end{equation}
The Fuchsian group that provides its uniformization, namely $\Gamma_{8}$, is generated, as suggested by its name, by $8$ generators that we will describe below. $\Gamma_{8}$ has finite index in $\Delta^+_{8,3,2}$ equal to 48. The order $48$
automorphism group of the Bolza surface named $\mathrm{GD}_{48}$ is isomorphic to the first normal subgroup of the order 96 Fermat Quartic automorphism group as mentioned in equation (\ref{normalsubbo2}). Hence the two Riemann surfaces (\ref{fermaquattro}) and (\ref{bolzadue}) are strictly related at two levels. Both corresponding Fuchsian groups are normal subgroups of the same tiling group $\Delta^+_{8,3,2}$ and the second finite quotient group is a normal subgroup of the first finite quotient group. Therefore it is quite appropriate and illuminating to study the two tiling constructions together.
\par
The Bolza tiling is computationally simpler than the Fermat Quartic tiling and it is the most relevant for our purposes. Indeed it can be done in terms of geodesic quadrangles that have the same graph properties as a quadratic lattice tiling of the usual Euclidean plane. This is just all what one needs in order to map in an unequivocal way the \textit{pixel} data of any image into the base manifold $\mathbb{H}^2$ of a Tits Satake vector bundle, whose total space is another larger coset manifold in the same  Tits Satake universality class as described in previous sections.
\par
In such an outlined perspective we proceed to the construction of the above mentioned $\Delta^+_{8,3,2}$
tilings.
\subsection{The tessellation group}
The first step in the tessellation menu is the construction of the tiling group. We proceed to the construction of
$\Delta^+_{8,3,2}$.
\par
The finite, order 96, quotient group $\mathrm{GD_{96}}$, mentioned before being the automorphism group of the Fermat Quartic and containing as order 2 normal subgroup the automorphism group of Bolza surface  is independently described, starting from  a
faithful, irreducible, crystallographic, $6$-dimensional representation of its, in appendix \ref{margarina96}.
\subsubsection{Presentation of the tiling group}
According with its Coxeter diagram the subgroup of $\mathrm{PSL(2,\mathbb{R})}$ we are looking for must have the following presentation
\begin{equation}\label{equadelta1}
  \text{PSL}(2,\mathbb{R})\supset {\Delta ^+_{8,3,2}}{ }\equiv { }{\left\langle \mathfrak{T},\mathfrak{S} ,
  \mathfrak{R}\left| \mathfrak{T}^8\right.=\mathfrak{S}
^3=\mathfrak{R}^2=\mathfrak{T}\mathfrak{S}\mathfrak{R}= \text{Id}\right\rangle }
\end{equation}
\subsubsection{Explicit form of the generators in D=2}
A solution satisfying all the relations of eq. (\ref{equadelta1}) is provided by the following matrices:
\begin{eqnarray}\label{equadeltagenTS}
  \mathfrak{T}&=&\left(
\begin{array}{cc}
 \frac{\sqrt{\sqrt{2}+2}}{2} & -\frac{1}{2} \sqrt{2-\sqrt{2}} \\
 \frac{\sqrt{2-\sqrt{2}}}{2} & \frac{\sqrt{\sqrt{2}+2}}{2} \\
\end{array}
\right) \quad ; \quad \mathfrak{S}=\left(
\begin{array}{cc}
 \frac{1}{2} \left(-1+\sqrt{-1+\sqrt{2}}\right) & \frac{1}{\sqrt{2}+\sqrt{2 \left(-1+\sqrt{2}\right)}} \\
 \frac{1}{2} \left(-1-\sqrt{2}-\sqrt{1+\sqrt{2}}\right) & \frac{1}{2} \left(-1-\sqrt{-1+\sqrt{2}}\right) \\
\end{array}
\right) \nonumber\\
\mathfrak{R}&=&\left(
\begin{array}{cc}
 \frac{1}{2^{3/4}} & \sqrt{1+\frac{3}{2 \sqrt{2}}-\sqrt{1+\sqrt{2}}} \\
 -\sqrt{1+\frac{3}{2 \sqrt{2}}+\sqrt{1+\sqrt{2}}} & -\frac{1}{2^{3/4}} \\
\end{array}
\right)
\end{eqnarray}
\subsection{The  Fermat Quartic Fuchsian subgroup and the associated tessellation}
The additional key ingredient for the tessellation construction associated with the Fermat Quartic
surface is the following hyperbolic group element
\begin{equation}\label{Udefinition}
  \mathfrak{U} \equiv \mathfrak{T}\mathfrak{R}\mathfrak{S} = \mathfrak{T}^2\mathfrak{S} ^2
\end{equation}
whose explicit matrix representation is the following one:
\begin{equation}\label{Umatrice}
  \mathfrak{U}=\left(
\begin{array}{cc}
 \frac{1}{2} \left(-1-\sqrt{2}-\sqrt{1+\sqrt{2}}\right) & \frac{1}{2} \left(-1+\sqrt{-1+\sqrt{2}}\right) \\
 \frac{1}{2} \left(1+\sqrt{-1+\sqrt{2}}\right) & \frac{1}{2} \left(-1-\sqrt{2}+\sqrt{1+\sqrt{2}}\right) \\
\end{array}
\right)
\end{equation}
Indeed as we are going to see the $16$ generators of the Fuchsian group can be constructed as conjugates with respect to  $\mathfrak{T}^n$ (for
$n=0,\dots,7$) of either $\mathfrak{U}^3$ or of its inverse. If we work in a space where $\mathfrak{U}^3$ acts as the identity all Fuchsian group
generators also collapse to the identity so that  we are left with the quotient group.
\paragraph{\sc Projection onto the quotient group}
\label{gigioproietto}
In view of what we said above, the additional relation that projects onto the quotient group is the following one:
\begin{equation}\label{ombralunga}
  \mathfrak{U}^3 = \text{Id}
\end{equation}
from which  the presentation of the quotient group $\mathrm{GD_{96}}$ follows:
\begin{equation}\label{presentaG96}
  \text{GD}_{96} = \langle \mathfrak{T},\mathfrak{S} ,\mathfrak{R}\left| \mathfrak{T}^8\right.=\mathfrak{S} ^3=\mathfrak{R}^2=
  \mathfrak{T}\mathfrak{S}\mathfrak{R}=(\mathfrak{T}\mathfrak{R}\mathfrak{S})^3\text{
 }=\text{Id}\rangle
\end{equation}
An exhaustive analysis of the finite group defined by (\ref{presentaG96}) is provided in appendix \ref{margarina96} as we said above.
\par
Hence, the specific form of the general exact sequence (\ref{equafun1}) applied to our case is the following:
\begin{equation}\label{sequela2}
 \text{Id}\overset{\iota }{\longrightarrow }\Gamma _{16}\overset{\iota }{\longrightarrow }\Delta ^+_{8,3,2}\overset{\pi }{\longrightarrow
}\text{GD}_{96}\equiv \frac{\Delta ^+_{8,3,2}}{\Gamma _{16}}\overset{\pi }{\longrightarrow }\text{Id}
\end{equation}
As we already anticipated, the infinite normal Fuchsian subgroup was named by us $\Gamma _{16}$ since it can be presented in terms of 16 generators that we discuss next.
\subsection{Description of the normal subgroup \texorpdfstring{$\Gamma_{16}$}{G16}}
\label{Gamma16normale}
The normal subgroup was determined with a procedure completely analogous to that utilized in the case of the $\Delta^{+}_{7,3,2}$ tessellation for the Kleinian Fuchsian subgroup of index 168. Namely we tried to find $2\, p \, =\,16$ points, yielding  the vertices
$\hat{v}_p$ of a 16-gon in the tessellation dual to the searched one of the Fermat Quartic. The true vertices ${v}_p$ of the fundamental domain for $\Gamma_{16}$ might be found with an identical procedure to the one utilized for the smaller case of the Bolza surface in section \ref{bolzaneto}. For the Fermat Quartic, we just construct the 16 Fuchsian generators, we derive the relations that prove that they generate a normal subgroup of $\Delta^{+}_{8,3,2}$, and we confine ourselves to showing the vertices and geodesic arcs of the dual tessellation. As for the quotient we are guaranteed, from the very start, that it is the group $\mathrm{GD_{96}}$ since in terms of our construction of the Fuchsian generators (see below) all of them collapse to the identity when imposing the constraint (\ref{ombralunga}).
\par
Following the above-mentioned logic,  we found that all Fuchsian generators can  be constructed in terms of the following building block
\begin{equation}\label{mattoneQU}
  \mathfrak{Q}\mathfrak{U} \equiv \mathfrak{U}^3 = (\mathfrak{T}\mathfrak{R}\mathfrak{S})^3=\left(\mathfrak{T}^2\mathfrak{S} ^2\right)^3
\end{equation}
Its explicit form as a $2\times 2$ matrix is the following one:
\begin{equation}\label{QUmatrix}
  \mathfrak{Q}\mathfrak{U} = \left(
\begin{array}{cc}
 -2-\sqrt{2}-\sqrt{7+5 \sqrt{2}} & -1-\sqrt{2}+\sqrt{1+\sqrt{2}} \\
 1+\sqrt{2}+\sqrt{1+\sqrt{2}} & -2-\sqrt{2}+\left(1+\sqrt{2}\right)^{3/2} \\
\end{array}
\right)
\end{equation}
We also need the conjugate generator
\begin{eqnarray}
\label{UQmatrix}
  \mathfrak{U}\mathfrak{Q} & \equiv& \mathfrak{R} \mathfrak{U}^3\mathfrak{R} = \mathfrak{Q}\mathfrak{U}^{-1}
\end{eqnarray}
In terms of the hyperbolic generators $\mathfrak{Q}\mathfrak{U},\mathfrak{U}\mathfrak{Q}$ and of the elliptic generator $\mathfrak{T}$ of order  $8$ we construct the following two sequences of $8$ generators each:
\begin{alignat}{3}\label{genni16prima}
  \mathfrak{F}_m^+ & =\mathfrak{T}^{-m}\mathfrak{Q}\mathfrak{U}\mathfrak{T}^m\text{         }&;&\text{   }(m=0,1,2,3,4,5,6,7)\nonumber\\
\mathfrak{F}_m^-&=\mathfrak{T}^{-m-6}\mathfrak{U}\mathfrak{Q} \mathfrak{T}^{m+6}\text{   } &; & \text{   }(m=0,1,2,3,4,5,6,7)
\end{alignat}
and we reorganize the whole set of 16 operators in the following set of 16 objects:
\begin{equation}\label{mholoni}
  \mho _p=
\begin{array}{ll}
  & \left\{
\begin{array}{ll}
 \mathfrak{F}_m^+ & \text{if}\text{    }p=2m+1 \\
 \mathfrak{F}_m^- & \text{if}\text{    }p=2(m+1)  \\
\end{array}\right.
 \\
\end{array}
(m=0,1,2,3,4,5,6,7)
\end{equation}
The set of all words that can be written in terms of the $16$ letters $\mho _p$ that we christen $\Gamma_{16}$ constitutes a group since the set
$\mho_p$ contains the inverse of each of its 16 members, as it can be explicitly verified by matrix multiplication. Indeed one finds
\begin{equation}\label{inversmholoni}
  \begin{array}{ccccc}
 \mho _1\cdot \mho _6 & = & \mho _6\cdot \mho _1 & = & \text{Id} \\
 \mho _2\cdot \mho _{13} & = & \mho _{13}\cdot \mho _2 & = & \text{Id} \\
 \mho _3\cdot \mho _8 & = & \mho _8\cdot \mho _3 & = & \text{Id} \\
 \mho _4\cdot \mho _{15} & = & \mho _{15}\cdot \mho _4 & = & \text{Id} \\
 \mho _5\cdot \mho _{10} & = & \mho _{10}\cdot \mho _5 & = & \text{Id} \\
 \mho _7\cdot \mho _{12} & = & \mho _{12}\cdot \mho _7 & = & \text{Id} \\
 \mho _9\cdot \mho _{14} & = & \mho _{14}\cdot \mho _9 & = & \text{Id} \\
 \mho _{11}\cdot \mho _{16} & = & \mho _{16}\cdot \mho _{11} & = & \text{Id} \\
\end{array}
\end{equation}
The relations (\ref{inversmholoni}) guarantee that $\Gamma_{16}$ is a subgroup of $\Delta^+_{8,3,2}$, yet we do not know  whether $\Gamma_{16}$ is
a normal subgroup. To establish that $\Gamma_{16}\subset \Delta^+_{8,3,2}$ is normal we ought to consider the conjugation of the generators $\mho _p$ with the two generators $\mathfrak{T},\mathfrak{S}$ of the bigger group, and show that the conjugate of each  $\mho _p$ can be
written as a word in the same symbols. This suffices to prove that the conjugate by any $g\in \Delta^+_{8,3,2}$ of each element $\gamma \in \Gamma_{16}$ is still an element of the same subgroup.
\paragraph{\sc Conjugation with respect to $\mathfrak{T}$} With respect to the generator $\mathfrak{T}$ of order $8$ the conjugation of any of the
16 generators of the Fuchsian subgroup $\Gamma_{16}$ yields another generator in the same set. Indeed, we have:
\begin{equation}\label{sposoT}
  \begin{array}{ccc}
 {\mathfrak{T}^{-1}\cdot \mho _1\cdot \mathfrak{T}} & {=} & {\mho _3} \\
 {\mathfrak{T}^{-1}\cdot \mho _2\cdot \mathfrak{T}} & {=} & {\mho _4} \\
 {\mathfrak{T}^{-1}\cdot \mho _3\cdot \mathfrak{T}} & {=} & {\mho _5} \\
 {\mathfrak{T}^{-1}\cdot \mho _4\cdot \mathfrak{T}} & {=} & {\mho _6} \\
 {\mathfrak{T}^{-1}\cdot \mho _5\cdot \mathfrak{T}} & {=} & {\mho _7} \\
 {\mathfrak{T}^{-1}\cdot \mho _6\cdot \mathfrak{T}} & {=} & {\mho _8} \\
 {\mathfrak{T}^{-1}\cdot \mho _7\cdot \mathfrak{T}} & {=} & {\mho _9} \\
 {\mathfrak{T}^{-1}\cdot \mho _8\cdot \mathfrak{T}} & {=} & {\mho _{10}} \\
 \end{array}
 \quad ; \quad
 \begin{array}{ccc}
 {\mathfrak{T}^{-1}\cdot \mho _9\cdot \mathfrak{T}} & {=} & {\mho _{11}} \\
 {\mathfrak{T}^{-1}\cdot \mho _{10}\cdot \mathfrak{T}} & {=} & {\mho _{12}} \\
 {\mathfrak{T}^{-1}\cdot \mho _{11}\cdot \mathfrak{T}} & {=} & {\mho _{13}} \\
 {\mathfrak{T}^{-1}\cdot \mho _{12}\cdot \mathfrak{T}} & {=} & {\mho _{14}} \\
 {\mathfrak{T}^{-1}\cdot \mho _{13}\cdot \mathfrak{T}} & {=} & {\mho _{15}} \\
 {\mathfrak{T}^{-1}\cdot \mho _{14}\cdot \mathfrak{T}} & {=} & {\mho _{16}} \\
 {\mathfrak{T}^{-1}\cdot \mho _{15}\cdot \mathfrak{T}} & {=} & {\mho _1} \\
 {\mathfrak{T}^{-1}\cdot \mho _{16}\cdot \mathfrak{T}} & {=} & {\mho _2} \\
\end{array}
\end{equation}
\paragraph{\sc Conjugation with respect to $\mathfrak{S}$} Under conjugation with respect to the generator $\mathfrak{S}$ of order $3$
the 16 generators of the  Fuchsian subgroup $\Gamma_{16}$ subdivide in two sets. For the $8$ generators of the first set, the conjugate image is the product of three different  $\mho_p$, while for the $8$ generators in the second set, the conjugate image is just a single $\mho_p$. Indeed, we
explicitly find:
\begin{equation}\label{sposoS}
\begin{array}{ccc}
 \mathfrak{S}\cdot \mathfrak{S}\cdot \mho _2\cdot \mathfrak{S} & = & \mho _4\cdot \mho _{13}\cdot \mho _1 \\
 \mathfrak{S}\cdot \mathfrak{S}\cdot \mho _4\cdot \mathfrak{S} & = & \mho _5\cdot \mho _{11}\cdot \mho _1 \\
 \mathfrak{S}\cdot \mathfrak{S}\cdot \mho _9\cdot \mathfrak{S} & = & \mho _4\cdot \mho _{14}\cdot \mho _8 \\
 \mathfrak{S}\cdot \mathfrak{S}\cdot \mho _{11}\cdot \mathfrak{S} & = & \mho _5\cdot \mho _{11}\cdot \mho _{15} \\
 \mathfrak{S}\cdot \mathfrak{S}\cdot \mho _{13}\cdot \mathfrak{S} & = & \mho _5\cdot \mho _9\cdot \mho _{15} \\
 \mathfrak{S}\cdot \mathfrak{S}\cdot \mho _{14}\cdot \mathfrak{S} & = & \mho _3\cdot \mho _9\cdot \mho _{15} \\
 \mathfrak{S}\cdot \mathfrak{S}\cdot \mho _{15}\cdot \mathfrak{S} & = & \mho _6\cdot \mho _{16}\cdot \mho _{10} \\
 \mathfrak{S}\cdot \mathfrak{S}\cdot \mho _{16}\cdot \mathfrak{S} & = & \mho _3\cdot \mho _7\cdot \mho _{10} \\
\end{array} \quad ; \quad \begin{array}{ccc}
 \mathfrak{S}\cdot \mathfrak{S}\cdot \mho _1\cdot \mathfrak{S} & = & \mho _{10} \\
 \mathfrak{S}\cdot \mathfrak{S}\cdot \mho _3\cdot \mathfrak{S} & = & \mho _1 \\
 \mathfrak{S}\cdot \mathfrak{S}\cdot \mho _5\cdot \mathfrak{S} & = & \mho _3 \\
 \mathfrak{S}\cdot \mathfrak{S}\cdot \mho _6\cdot \mathfrak{S} & = & \mho _5 \\
 \mathfrak{S}\cdot \mathfrak{S}\cdot \mho _7\cdot \mathfrak{S} & = & \mho _4 \\
 \mathfrak{S}\cdot \mathfrak{S}\cdot \mho _8\cdot \mathfrak{S} & = & \mho _6 \\
 \mathfrak{S}\cdot \mathfrak{S}\cdot \mho _{10}\cdot \mathfrak{S} & = & \mho _8 \\
 \mathfrak{S}\cdot \mathfrak{S}\cdot \mho _{12}\cdot \mathfrak{S} & = & \mho _{15} \\
\end{array}
\end{equation}
\subsubsection{Grid \texorpdfstring{$\hat{v}_p$}{vp} of first tile centers}
As we explained in section \ref{Deltap32} the necklace of 16 copies of the fundamental domain $\mathcal{F}_{\Gamma_{16}}$
given by $\mho_p\left(\mathcal{F}_{\Gamma_{16}}\right)$ contains the sequence of their central points, given just by the images
$\mho_{p}$ of the identity element $\mathbf{e}$ of the solvable Lie group $\mathcal{S}_{\mathbb{H}^2}$.  These points that fix the angular position of the lower-lying vertices of the fundamental domain boundary $\partial \mathcal{F}_{\Gamma_{16}}$ are just $\mho_p(\mathbf{e})$. In practice, given the $16$ Fuchsian subgroup generators $\mho _{i}$, we can construct the images under the fractional linear transformation of each $\mho _{p}$ of the origin $(0,1)$ in the upper complex plane. These $16$ points are mapped by the Cayley transformation into $16$ points of the unit disk that lie all on the circumference of radius $R=\sqrt{\frac{2\sqrt{2}}{3}}$ while the origin $(0,1)$ is mapped into disk center $(0,0)$.
\par
\begin{figure}[htb]
\begin{center}
\vskip 1cm
\includegraphics[width=70mm]{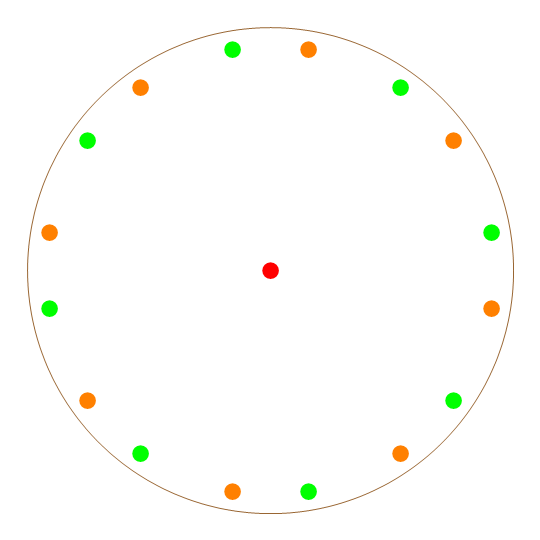}
\vskip 1cm
\caption{\label{fettuccia} In this figure, we display the 16 centers of the necklace of 16 tiles surrounding the fundamental domain for the Fuchsian group $\Gamma_{16}$.
The orange points named $P_a$ in eq.(\ref{trallo}) are the images through the odd numbered generators $\mho_{1+2i}$ of the disk center.  The green points named $Q_a$ are the images of the center under the action of the even-numbered generators $\mho_{2+2i}$. The numerical coordinates of these points are displayed in eq.(\ref{trallo}). }
\end{center}
\end{figure}

Separating among the $16$ points, that are in one-to-one correspondence with the Fuchsian generators, the odd (named $P$) from the even ones (named $Q$), we obtain the following explicit representation  of the $\hat{v}_p$ vertices of the dual tessellation in terms of disk coordinates:
\begin{equation}\label{trallo}
\left\{ \begin{array}{l}
 P_1\text{ = }\left\{\frac{\sqrt{1+\sqrt{2}}}{3},\frac{1}{3} \sqrt{-1+5
   \sqrt{2}}\right\} \\
 P_2\text{ = }\left\{-\frac{1}{3} \sqrt{-1+\sqrt{2}},\frac{1}{3} \sqrt{1+5
   \sqrt{2}}\right\} \\
 P_3\text{ = }\left\{-\frac{1}{3} \sqrt{-1+5
   \sqrt{2}},\frac{\sqrt{1+\sqrt{2}}}{3}\right\} \\
 P_4\text{ = }\left\{-\frac{1}{3} \sqrt{1+5 \sqrt{2}},-\frac{1}{3}
   \sqrt{-1+\sqrt{2}}\right\} \\
 P_5\text{ = }\left\{-\frac{1}{3} \sqrt{1+\sqrt{2}},-\frac{1}{3} \sqrt{-1+5
   \sqrt{2}}\right\} \\
 P_6\text{ = }\left\{\frac{1}{3} \sqrt{-1+\sqrt{2}},-\frac{1}{3} \sqrt{1+5
   \sqrt{2}}\right\} \\
 P_7\text{ = }\left\{\frac{1}{3} \sqrt{-1+5 \sqrt{2}},-\frac{1}{3}
   \sqrt{1+\sqrt{2}}\right\} \\
 P_8\text{ = }\left\{\frac{1}{3} \sqrt{1+5 \sqrt{2}},\frac{1}{3}
   \sqrt{-1+\sqrt{2}}\right\} \\
\end{array}\right. \quad ; \quad  \left\{
\begin{array}{l}
 Q_1\text{ = }\left\{\frac{1}{3} \sqrt{-1+\sqrt{2}},\frac{1}{3} \sqrt{1+5
   \sqrt{2}}\right\} \\
 Q_2\text{ = }\left\{-\frac{1}{3} \sqrt{1+\sqrt{2}},\frac{1}{3} \sqrt{-1+5
   \sqrt{2}}\right\} \\
 Q_3\text{ = }\left\{-\frac{1}{3} \sqrt{1+5 \sqrt{2}},\frac{1}{3}
   \sqrt{-1+\sqrt{2}}\right\} \\
 Q_4\text{ = }\left\{-\frac{1}{3} \sqrt{-1+5 \sqrt{2}},-\frac{1}{3}
   \sqrt{1+\sqrt{2}}\right\} \\
 Q_5\text{ = }\left\{-\frac{1}{3} \sqrt{-1+\sqrt{2}},-\frac{1}{3} \sqrt{1+5
   \sqrt{2}}\right\} \\
 Q_6\text{ = }\left\{\frac{\sqrt{1+\sqrt{2}}}{3},-\frac{1}{3} \sqrt{-1+5
   \sqrt{2}}\right\} \\
 Q_7\text{ = }\left\{\frac{1}{3} \sqrt{1+5 \sqrt{2}},-\frac{1}{3}
   \sqrt{-1+\sqrt{2}}\right\} \\
 Q_8\text{ = }\left\{\frac{1}{3} \sqrt{-1+5
   \sqrt{2}},\frac{\sqrt{1+\sqrt{2}}}{3}\right\} \\
\end{array}
\right.
\end{equation}
The image of the 16 vertices is displayed in fig.\ref{fettuccia}
\subsubsection{Completing the dual tessellation up to the necklace of tile centers}
Having shown the centers of the first necklace of octagon tiles around the fundamental domain of the Fermat quartic, we consider the innermost octagon formed by 8 triangles of type $(8,3,2)$, which constitute the triangular tiles of the dual tessellation. The vertices of such an innermost octagon are the centers of the first octagon tiles, in the direct tessellation,  located around the origin. The innermost dual octagon can be generated by constructing the twins of the Fuchsian subgroup generators with the first power of $\mathfrak{U}$ replacing the third
power. More generally, we can introduce the following operators:
\begin{eqnarray}\label{lagorio}
 I_m^{(p,+)} &\equiv & \mathfrak{T}^{-m}\mathfrak{U}^p\mathfrak{T}^m \,\, \,\quad ;\quad (p=1,2,3) \quad (m,0,1,2,\text{..},7) \nonumber\\
 I_m^{(p,-)}   &\equiv& \mathfrak{T}^{-m}\mathfrak{U}^{-p}\mathfrak{T}^m \quad ; \quad (p=1,2,3) \quad (m,0,1,2,\text{..},7)
\end{eqnarray}
and we find that
\begin{equation}\label{kronino}
  I_m^{(3,\pm )}=\mathfrak{F}_m^{\pm }
\end{equation}
 correspond to the generators of the Fuchsian group $\Gamma _{16}$. The 8 vertices of the above-mentioned innermost dual octagon are the images of the central point of the disk (0,0) with respect to either the generators \(I_m^{(1,+)}\) (m=0,..,7) or to the generators \(I_m^{(1,-)}\) that
 yield identical results.
\begin{figure}[htb]
\begin{center}
\vskip 1cm
\includegraphics[width=160mm]{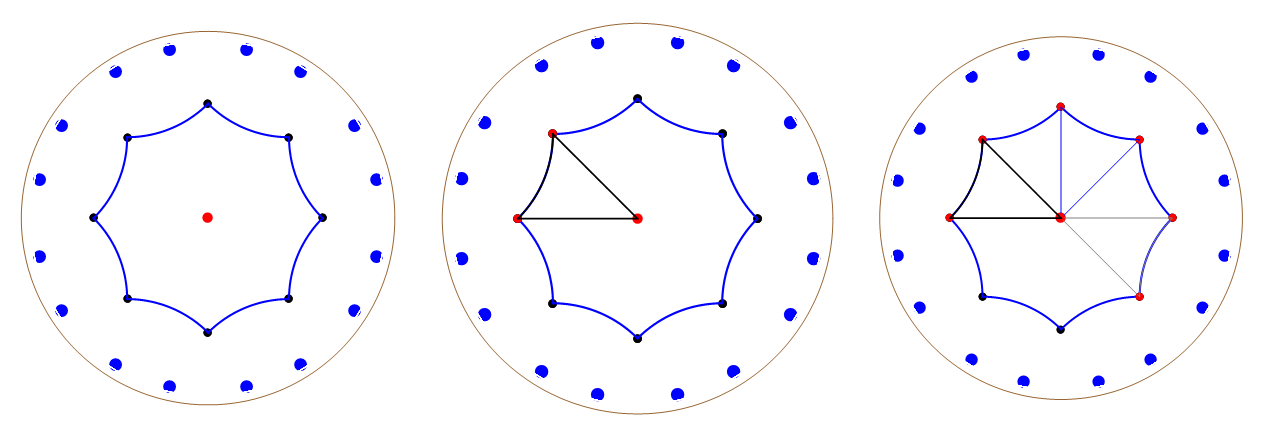}
\vskip 1cm
\caption{\label{balubbo} In this figure, we show the innermost octagon of the dual tessellation for the Fermat Quartic. The vertices are images of
the center through the operators $I_m^{1,+}$. In the central image,  we show the fundamental triangle obtained by adjoining to the center its
two images through the generator $\mathfrak{S}$ of order 3. Then acting repeatedly with the generator $\mathfrak{T}$ of order 8 on the fundamental
triangle, we generate the remaining 7 triangles that provide a tessellation of the innermost octagon, as it is suggested in the right image.  }
\end{center}
\end{figure}
In fig.\ref{balubbo} we show the innermost octagon of the dual tessellation for the Fermat Quartic. Its vertices are the centers of the first ring of octagons of the direct tessellation. The innermost dual octagon is composed of 8 triangles whose vertices are one in the center while the other two of each triangle are located on the octagon boundary.
\begin{figure}
\begin{center}
\vskip 1cm
\includegraphics[width=100mm]{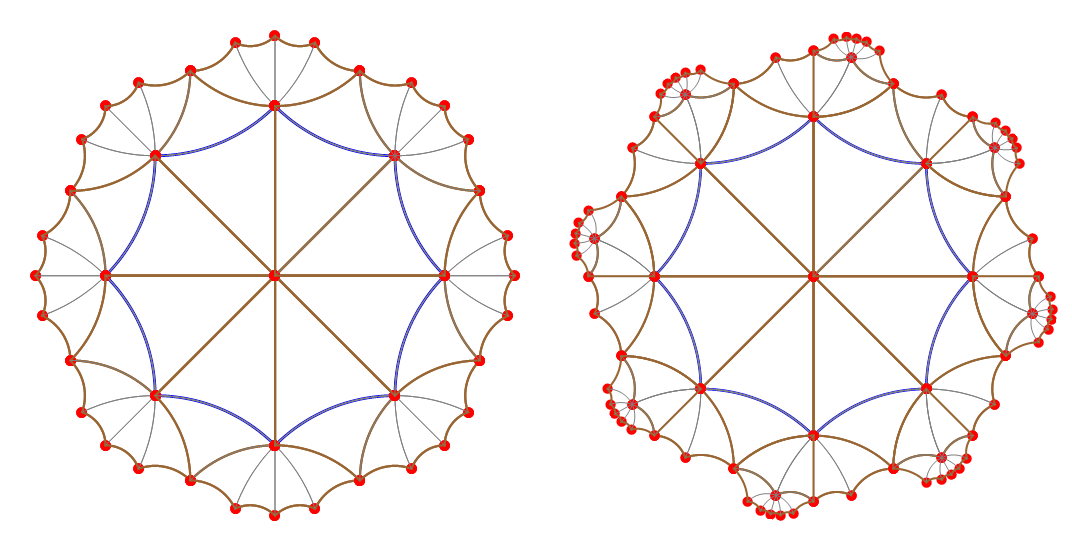}
\includegraphics[width=70mm]{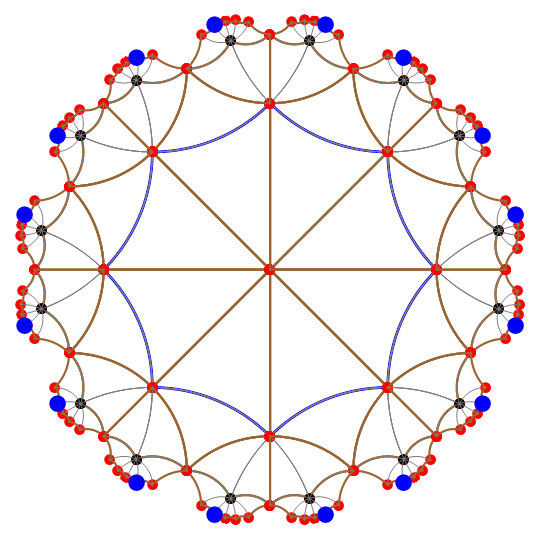}
\vskip 1cm
\caption{\label{obbulab} The first two upper images in this figure show two intermediate steps in the tessellation, via the 48 images (many partially overlapping) of the innermost octagon, equipped with its internal geodesic arcs that slice it into 8 triangles. The lower image displays instead the complete tessellation up to the necklace of direct tile centers. The latter, marked in blue, are the points $\hat{v}_p$  and, as one sees, they are vertices in the dual tessellation of the Fermat Quartic one. }
\end{center}
\end{figure}
Once the innermost dual octagon is constructed, by taking its images through the 48 operators mentioned in eq. (\ref{lagorio}) we obtain the complete dual tessellation up to the necklace of the direct tile centers we mentioned above: see fig.\ref{obbulab}.
The vertices of the innermost dual octagon are explicitly displayed below:
\begin{alignat}{3}\label{innerpuntiFG}
&\mathit{p}_1^o\text{ = }\left\{\sqrt{-\frac{1}{2}+\frac{1}{\sqrt{2}}},\sqrt{-\frac{1}{2}+\frac{1}{\sqrt{2}}}\right\}& \quad\quad
&\mathit{p}_2^o\text{ = }\left\{0,\sqrt{-1+\sqrt{2}}\right\}\nonumber\\
&\mathit{p}_3^o\text{ = }\left\{-\sqrt{-\frac{1}{2}+\frac{1}{\sqrt{2}}},\sqrt{-\frac{1}{2}+\frac{1}{\sqrt{2}}}\right\}& \quad\quad
&\mathit{p}_4^o\text{ = }\left\{-\sqrt{-1+\sqrt{2}},0\right\}\nonumber\\
&\mathit{p}_5^o\text{ = }\left\{-\sqrt{-\frac{1}{2}+\frac{1}{\sqrt{2}}},-\sqrt{-\frac{1}{2}+\frac{1}{\sqrt{2}}}\right\}& \quad\quad
&\mathit{p}_6^o\text{ = }\left\{0,-\sqrt{-1+\sqrt{2}}\right\}\nonumber\\
&\mathit{p}_7^o\text{ = }\left\{\sqrt{-\frac{1}{2}+\frac{1}{\sqrt{2}}},-\sqrt{-\frac{1}{2}+\frac{1}{\sqrt{2}}}\right\}& \quad\quad
&\mathit{p}_8^o\text{ = }\left\{\sqrt{-1+\sqrt{2}},0\right\}
\end{alignat}
\subsection{The Bolza surface Fuchsian subgroup and the associated tessellation}
\label{bolzaneto}
To discuss the Fuchsian subgroup associated to the Bolza surface, it is convenient to introduce a preliminary discussion that focuses on a
tiling subgroup of the original tiling group $\Delta^+_{8,3,2}$. The tiling group  we have in mind is the even subgroup $\Delta^{+}_{4,3,3}$ of
the Coxeter group $\Delta_{4,3,3}^+$ described by the following Coxeter diagram of fig.\ref{quatuor}:
\begin{figure}[htb]
\vskip 1cm
\centering
\begin{picture}(50,150)
\put(-85,157){$\Delta_{4,3,3} \quad:\quad $}\put(-30,160){\circle{10}}
\put(-25,160){\line(1,0){25}} \put(-15,165){$4$}
\put (5,160){\circle {10}} \put(5,155){\line (0,-1){25}}
\put(8,140){$\null$}\put (5,125){\circle {10}} \put(1,127){\line(-1,1){28}}\put(-22,138){$\null$}
\end{picture}
\vskip -4cm
\caption{\label{quatuor} The Coxeter graph of the triangular groups $\Delta_{4,3,3}$ }
\end{figure}
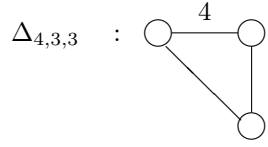
The standard presentation of $\Delta^{+}_{4,3,3}$ is then the following one:
\begin{equation}\label{D433presen}
  \text{PSL}(2,\mathbb{R})\supset\Delta ^+{}_{4,3,3}\, \equiv \,\langle \mathfrak{A},\mathfrak{B},\mathfrak{C}\, |\,
\mathfrak{A}^4=\mathfrak{B} ^3=\mathfrak{C}^3=\mathfrak{A}\mathfrak{B}\mathfrak{C}= \text{Id}\rangle
\end{equation}
The group \(\Delta ^+{}_{4,3,3}\) can be seen to be  a subgroup of \(\Delta _{8,3,2}^+\) since its generators satisfying the relations mentioned in eq.(\ref{D433presen}) can be realized as words in the generators of the latter:
\begin{equation}\label{generatoringen}
  \mathfrak{A}= \mathfrak{T}^2 \quad ; \quad \mathfrak{B}=\mathfrak{S} \quad ; \quad \mathfrak{C}=\mathfrak{T}^2 \mathfrak{S}
\end{equation}
Furthermore, if we define the group element :
\begin{equation}\label{Wgenno}
  \mathfrak{W} \equiv \mathfrak{A}^2\mathfrak{B} \mathfrak{A}^2 \mathfrak{B}^2=\mathfrak{T}^4\mathfrak{S} \mathfrak{T}^4 \mathfrak{S}^2
\end{equation}
and we impose the additional relation
\begin{equation}\label{Wvattene}
  \mathfrak{W} = \text{Id}
\end{equation}
we obtain an abstract finite group of order 48
\begin{equation}\label{gardenia48}
  \text{GD}_{48} =\langle \mathfrak{A},\mathfrak{B},\mathfrak{C}\, |\,
\mathfrak{A}^4=\mathfrak{B} ^3=\mathfrak{C}^3=\mathfrak{A}\mathfrak{B}\mathfrak{C}=\mathfrak{A}^2\mathfrak{B} \mathfrak{A}^2 \mathfrak{B}^2 = \text{Id}\rangle
\end{equation}
We studied the structure of this abstract group in abstract form relying on the faithful 6-dimensional irreducible representation of the larger group \(\text{GD}_{96}\) of which \(\text{GD}_{48}\) is a normal subgroup of index 2 (see appendix \ref{margarina96}).
The group \(\text{GD}_{48}\) can be seen as the quotient of \(\Delta ^+{}_{8,3,2}\) with respect to an infinite normal subgroup \(\Gamma_8\)  whose structure we describe below, according with the exact sequence:
\begin{equation}\label{sequela3}
 \text{Id}\overset{\iota }{\longrightarrow }\Gamma _{8}\overset{\iota }{\longrightarrow }\Delta ^+_{8,3,2}\overset{\pi}{\longrightarrow}\text{GD}_{48}\equiv \frac{\Delta ^+_{8,3,2}}{\Gamma _{8}}\overset{\pi }{\longrightarrow }\text{Id}
\end{equation}
 At first sight one might think that due to the embedding of $\Delta ^+{}_{4,3,3}$ as a subgroup of  $\Delta ^+_{8,3,2}$ and due to the presentation of the quotient as given in eq.(\ref{gardenia48}) we could find $\text{GD}_{48}$ as the quotient of $\Delta ^+{}_{4,3,3}$ by some other Fuchsian subgroup $\Gamma^\prime$ of the latter. This is not excluded, yet the Fuchsian group $\Gamma_8$ that is naturally associated with the boundary of the fundamental domain of Bolza surface has 8 generators $\mho^B_m$ that cannot be written solely in terms of $\mathfrak{A},\mathfrak{B}$ and $\mathfrak{C}$, rather require $\mathfrak{T}$ and $\mathfrak{S}$. Moreover the conjugation relations of the 8 generators that guarantee that they generate a normal subgroup work perfectly within $\Delta ^+_{8,3,2}$, but do not work within $\Delta ^+{}_{4,3,3}$. Henceforth, the conclusion is that the correct interpretation of the quotient is that in equation (\ref{sequela3}).
\subsubsection{Generators of the group \texorpdfstring{$\Delta ^+{}_{4,3,3}$}{D433+}}
According with the  definition in eq.(\ref{generatoringen}), the explicit form of the \(\Delta ^+{}_{4,3,3}\)
generators is  the following one:
\begin{eqnarray}\label{ABCgenni}
 \mathfrak{A} & = &\left(
\begin{array}{cc}
 \frac{1}{\sqrt{2}} & -\frac{1}{\sqrt{2}} \\
 \frac{1}{\sqrt{2}} & \frac{1}{\sqrt{2}} \\
\end{array}
\right) \nonumber\\
 \mathfrak{B} & = & \left(
\begin{array}{cc}
 \frac{1}{2} \left(-1+\sqrt{-1+\sqrt{2}}\right) & \frac{1}{\sqrt{2}+\sqrt{2 \left(-1+\sqrt{2}\right)}} \\
 \frac{1}{2} \left(-1-\sqrt{2}-\sqrt{1+\sqrt{2}}\right) & \frac{1}{2} \left(-1-\sqrt{-1+\sqrt{2}}\right) \\
\end{array}
\right) \nonumber\\
 \mathfrak{C} &= & \left(
\begin{array}{cc}
 \frac{1}{2} \left(1-\sqrt{-1+\sqrt{2}}\right) & \frac{1}{2}+\frac{1}{\sqrt{2}}-\frac{\sqrt{1+\sqrt{2}}}{2} \\
 \frac{1}{2} \left(-1-\sqrt{2}-\sqrt{1+\sqrt{2}}\right) & \frac{1}{2} \left(1+\sqrt{-1+\sqrt{2}}\right) \\
\end{array}
\right)
\end{eqnarray}
\subsubsection{Choice of the building block for the normal subgroup generators}
The generators of the normal Fuchsian subgroup \(\Gamma _8\) are constructed in terms of the already mentioned object $\mathfrak{W}$  derived from the fundamental generators.  It is similar to, but different from the fundamental building block utilized in the case of the of the Fermat Quartic. In the case of the Bolza surface the building block for the generators has been deduced from the information found in the literature \cite{extrasistola,cuocogiuseppe} about
the length of the fundamental systole of the Bolza surface which  is $2 \,\text{ArcCosh}\left [1+\sqrt{2}\right]$ (see section
\ref{straccioni} for the concept of trace length spectra). Therefore the trace length of  the Fuchsian generators  must be $1+\sqrt{2}$ and all the primitive geodesics will form a lattice with lengths $2 \,\text{ArcCosh}[n+m\sqrt{2}]$, (n,m$\in \mathbb{Z}$) as established in the literature \cite{cuocogiuseppe}.
\par
Hence, analogously to what we did in the case of the Fermat Quartic tessellation, we  selected as kernel a hyperbolic element of
$\Delta _{4,3,3}^+$, whose conjugates with respect to elliptic elements were to be the generators of the Fuchsian subgroup defining the Bolza Riemann surface; yet in view of what we just said, we carefully selected such an element (namely $\mathfrak{W}$) in such a way that its trace length were the minimal one for the Bolza surface \textit{i.e}
$2 \,\text{ArcCosh}[1+\sqrt{2}]$.
Obviously there were several such matrices yet we also took care that $\mathfrak{W}$ might  be written as a fairly simple word in the generators
of \(\Delta _{4,3,3}^+\). With our original choice of the generators of \(\Delta _{8,3,2}^+\) which implies the explicit form of the generators
$\mathfrak{A}$,$\mathfrak{B}$,$\mathfrak{C}$ of of \(\Delta _{4,3,3}^+\)  given in eq. (\ref{ABCgenni}) we have found the building block
$\mathfrak{W}$ defined in eq.(\ref{Wgenno}). Its explicit matrix form is the following one:
\begin{equation}\label{esplicW}
  \mathfrak{W}=\left(
\begin{array}{cc}
 -1-\sqrt{2}-\sqrt{2 \left(1+\sqrt{2}\right)} & 0 \\
 0 & -1-\sqrt{2}+\sqrt{2 \left(1+\sqrt{2}\right)} \\
\end{array}
\right)
\end{equation}
\subsubsection{The  generators of the normal Fuchsian subgroup \texorpdfstring{$\Gamma_8$}{G8} for the Bolza surface}
Let us now consider the construction of the 8 generators of the Fuchsian group \(\Gamma _8\) .
For this purpose, we need the above-discussed building block $\mathfrak{W}$.
We proceed as follows. First, imitating what we  previously did in the Fermat Quartic  case we construct the mixed Fuchsian group elements
defined as follows
\begin{equation}\label{pregenniBol}
 \mathfrak{F}\mathfrak{K}_m^{\pm } = \mathfrak{T}^{-m} \mathfrak{W}^{\pm 1}\mathfrak{T}^m \quad ; \quad (m=0,1,\text{..},7)
\end{equation}
and we show that, apart from ordering, the set \(\mathfrak{F}\mathfrak{K}_m^+\) coincides with set \(\mathfrak{F}\mathfrak{K}_m^-\). Hence we have
8 hyperbolic group elements of \(\Delta ^+{}_{8,3,2}\) that we simply name \(\mathfrak{F}\mathfrak{K}_m\) and we identify with
\(\mathfrak{F}\mathfrak{K}_m^+\). They will be used, as we show below, in order to construct  the fundamental domain boundary of the Fuchsian
subgroup \(\Gamma _8\) that defines the Bolza surface as the quotient \(\mathbb{H}^2\)/\(\Gamma _8\).
\paragraph{Explicit form of the $\Gamma_8$ generators}
\begin{alignat}{4} \label{gam8genni}
&\mathfrak{K}\mathfrak{F}_0& = &\left(
\begin{array}{cc}
 -\sqrt{2 \left(1+\sqrt{2}\right)}-\sqrt{2}-1 & 0 \\
 0 & \sqrt{2 \left(1+\sqrt{2}\right)}-\sqrt{2}-1 \\
\end{array}
\right) \nonumber\\
&\mathfrak{K}\mathfrak{F}_1& = &\left(
\begin{array}{cc}
 -\sqrt{1+\sqrt{2}}-\sqrt{2}-1 & \sqrt{1+\sqrt{2}} \\
 \sqrt{1+\sqrt{2}} & \sqrt{1+\sqrt{2}}-\sqrt{2}-1 \\
\end{array}
\right)\nonumber\\
&\mathfrak{K}\mathfrak{F}_2& = &\left(
\begin{array}{cc}
 -1-\sqrt{2} & \sqrt{2 \left(1+\sqrt{2}\right)} \\
 \sqrt{2 \left(1+\sqrt{2}\right)} & -1-\sqrt{2} \\
\end{array}
\right)\nonumber\\
&\mathfrak{K}\mathfrak{F}_3& = &\left(
\begin{array}{cc}
 \sqrt{1+\sqrt{2}}-\sqrt{2}-1 & \sqrt{1+\sqrt{2}} \\
 \sqrt{1+\sqrt{2}} & -\sqrt{1+\sqrt{2}}-\sqrt{2}-1 \\
\end{array}
\right)\nonumber\\
&\mathfrak{K}\mathfrak{F}_4& = &\left(
\begin{array}{cc}
 \sqrt{2 \left(1+\sqrt{2}\right)}-\sqrt{2}-1 & 0 \\
 0 & -\sqrt{2 \left(1+\sqrt{2}\right)}-\sqrt{2}-1 \\
\end{array}
\right)\nonumber\\
&\mathfrak{K}\mathfrak{F}_5& = &\left(
\begin{array}{cc}
 \sqrt{1+\sqrt{2}}-\sqrt{2}-1 & -\sqrt{1+\sqrt{2}} \\
 -\sqrt{1+\sqrt{2}} & -\sqrt{1+\sqrt{2}}-\sqrt{2}-1 \\
\end{array}
\right)\nonumber\\
&\mathfrak{K}\mathfrak{F}_6& = &\left(
\begin{array}{cc}
 -1-\sqrt{2} & -\sqrt{2 \left(1+\sqrt{2}\right)} \\
 -\sqrt{2 \left(1+\sqrt{2}\right)} & -1-\sqrt{2} \\
\end{array}
\right)\nonumber\\
&\mathfrak{K}\mathfrak{F}_7& = &\left(
\begin{array}{cc}
 -\sqrt{1+\sqrt{2}}-\sqrt{2}-1 & -\sqrt{1+\sqrt{2}} \\
 -\sqrt{1+\sqrt{2}} & \sqrt{1+\sqrt{2}}-\sqrt{2}-1 \\
\end{array}
\right)
\end{alignat}
\subsubsection{Proof that \texorpdfstring{$\Gamma_8$}{G8} is a normal subgroup of \texorpdfstring{$\Delta _{8,3,2}^+$}{D832+}}
Just as we did in the case of the Fuchsian subgroup $\Gamma_{16}$ corresponding to the Fermat Quartic surface, in order to prove that the subgroup generated by the 8 generators in eq.(\ref{gam8genni}) is normal in \(\Delta _{8,3,2}^+\) we need to show that the conjugation of each generator \(\pmb{\mathfrak{K}}\mathfrak{F}_i\) with respect to the two generators of the full group \(\pmb{\Delta ^+{}_{8,3,2}}\), namely $\mathfrak{T}$ and $\mathfrak{S}$ is a word in \(\mathfrak{K}\mathfrak{F}_i\).
\paragraph{Conjugation with respect to the order 8 generator $\mathfrak{T}$}
By means of the very definition (\ref{pregenniBol})  we have
\begin{equation}
\begin{array}{ccc}
 \mathfrak{T}^{-1}{\cdot }{\mathfrak{K}}\mathfrak{F}_1{\cdot }{\mathfrak{T}}
 & = & {\mathfrak{K}}\mathfrak{F}_{i+1}  \\
 {\mathfrak{T}}^{-1}{\cdot }{\mathfrak{K}}\mathfrak{F}_7{\cdot }{\mathfrak{T}}
 & {=} & {\mathfrak{K}\mathfrak{F}_1}
\end{array}
 \quad \quad (i,1,\dots ,7)
\end{equation}
\paragraph{Conjugation with respect to the order 3 generator $\mathfrak{S}$}
In a similar way to what happened in the case of the group \(\Gamma _{16}\) for the Fermat Quartic, also in the case \(\Gamma _8\) which leads to
the Bolza surface, the 8 generators \(\mathfrak{K}\mathfrak{F}_i\)  split into two subsets containing 4 elements each. The $\mathfrak{S}$
conjugate of the first subset (the odd-numbered ones) are equal to the product of 3 generators out of the eight, while the $\mathfrak{S}$
conjugate of the second subset (the even-numbered ones) are equal to the product of two generators. Explicitly we have the result  shown below:
\begin{equation}\label{esseconiugio}
  \begin{array}{ccc}
 \mathfrak{S}\cdot \mathfrak{S}\cdot \mathfrak{K}\mathfrak{F}_1\cdot \mathfrak{S} &  =  & \mathfrak{K}\mathfrak{F}_1\cdot \mathfrak{K}\mathfrak{F}_5\cdot
\mathfrak{K}\mathfrak{F}_4 \\
 \mathfrak{S}\cdot \mathfrak{S}\cdot \mathfrak{K}\mathfrak{F}_3\cdot \mathfrak{S} &  =  & \mathfrak{K}\mathfrak{F}_5\cdot \mathfrak{K}\mathfrak{F}_2\cdot
\mathfrak{K}\mathfrak{F}_7 \\
 \mathfrak{S}\cdot \mathfrak{S}\cdot \mathfrak{K}\mathfrak{F}_5\cdot \mathfrak{S} &  =  & \mathfrak{K}\mathfrak{F}_1\cdot \mathfrak{K}\mathfrak{F}_5\cdot
\mathfrak{K}\mathfrak{F}_8 \\
 \mathfrak{S}\cdot \mathfrak{S}\cdot \mathfrak{K}\mathfrak{F}_7\cdot \mathfrak{S} &  =  & \mathfrak{K}\mathfrak{F}_3\cdot \mathfrak{K}\mathfrak{F}_6\cdot
\mathfrak{K}\mathfrak{F}_1 \\
\end{array}
 \quad ; \quad
\begin{array}{ccc}
 \mathfrak{S}\cdot \mathfrak{S}\cdot \mathfrak{K}\mathfrak{F}_2\cdot \mathfrak{S} & = & \mathfrak{K}\mathfrak{F}_4\cdot \mathfrak{K}\mathfrak{F}_7
\\
 \mathfrak{S}\cdot \mathfrak{S}\cdot \mathfrak{K}\mathfrak{F}_4\cdot \mathfrak{S} & = & \mathfrak{K}\mathfrak{F}_5\cdot \mathfrak{K}\mathfrak{F}_8
\\
 \mathfrak{S}\cdot \mathfrak{S}\cdot \mathfrak{K}\mathfrak{F}_6\cdot \mathfrak{S} & = & \mathfrak{K}\mathfrak{F}_3\cdot \mathfrak{K}\mathfrak{F}_8
\\
 \mathfrak{S}\cdot \mathfrak{S}\cdot \mathfrak{K}\mathfrak{F}_8\cdot \mathfrak{S} & = & \mathfrak{K}\mathfrak{F}_4\cdot \mathfrak{K}\mathfrak{F}_1
\\
\end{array}
\end{equation}
\subsection{The boundary of the fundamental domain for Bolza surface}
\begin{figure}[htb]
\begin{center}
\vskip 1cm
\includegraphics[width=120mm]{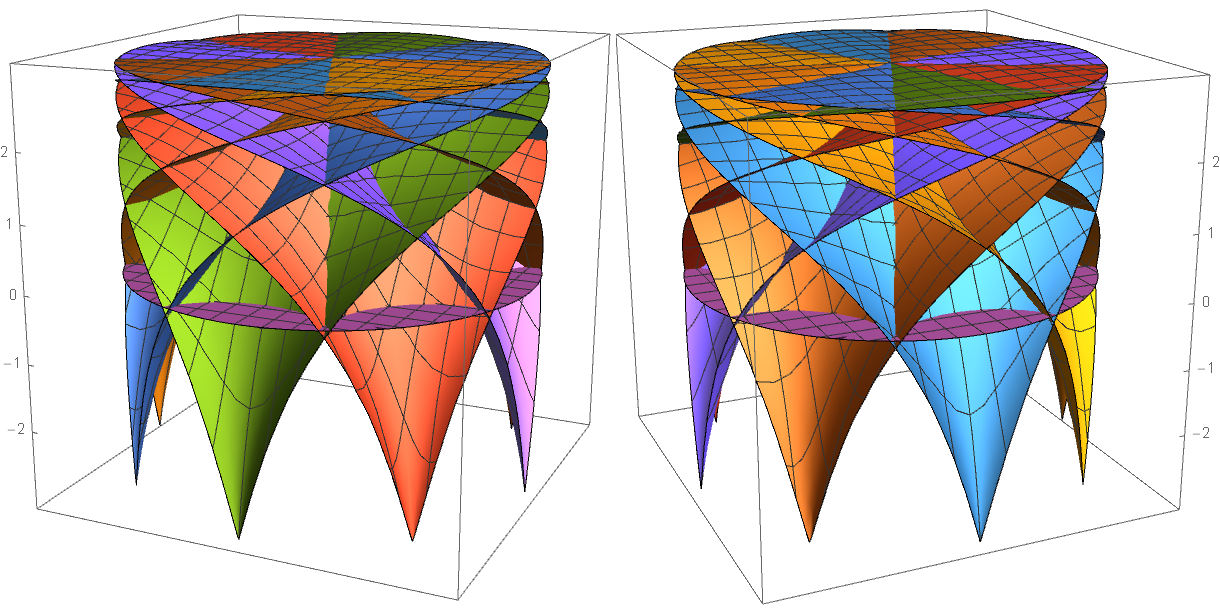}
\vskip 1cm
\caption{\label{cremapendente} The two images in this figure show the plot of the $8$ \textit{norm difference functions}
$\mathrm{Nd}_{\mathfrak{K}\mathfrak{F}_i}$  over the unit disk. Relevant for our arguments is the intersection of these
functions with the disk located at height $0$. The locus of each intersection is a geodesic arc. }
\end{center}
\end{figure}

Utilizing the definition of the \textit{norm difference functions} provided in eq.(\ref{ndiffun}) and relying on statement \ref{defioperativa} we
can easily construct the $8$ functions $\mathrm{Nd}_{\mathfrak{K}\mathfrak{F}_i}$ and plot them over the unit disk as shown in
fig.\ref{cremapendente}.
As we see a boundary of a compact region is cut in the $\mathbb{H}^2$ plane by the crossing of the eight difference functions at height $0$. As
one sees, the boundary of this domain is composed by 8 geodesic sides of a geodesic octagon and the vertices are the meeting points of such sides
that are arranged on a circle of radius $R= \sqrt{2\left(-1+\sqrt{2}\right)}$. These vertices are the images of the centre $(0,0)$ through the
eight Fuchsian group generators, namely they are the vertices $\hat{v}_p$ introduced in eq.(\ref{nuovafrontiera}). The cautious inspector of the
above images notes however a small, yet very significant anomaly. Each of the eight length difference functions crosses the zero plane and becomes
negative at a first pertinent vertex and all along a geodesic arc up to the next vertex. After that vertex it is always positive. However there is
always another one of the 8 \textit{norm difference functions} that crosses the zero plane at the same angular distance of the next vertex but at
a smaller radial distance, so that the octagon defined by the eight vertices that are images of centre (0,0) through the eight Fuchsian generators
\(\mathfrak{K}\mathfrak{F}_i\) cannot be the fundamental domain of the Fuchsian group. We already knew this from the general argument put forward
in section \ref{Deltap32}, yet now we come from the present observation at a concrete strategy for the derivation of the true fundamental domain
boundary. From fig. \ref{cremapendente} we see that the octagon having the points $\hat{v}_p$ as vertices, contains small subregions around each
vertex that are out of the Dirichlet closed set of the Fuchsian group. In particular all the vertices are eminent representatives of such regions.
As already pointed out, since they are the images of the origin through the eight Fuchsian generators, there is always an element (the inverse of
the generator associated with each vertex) that maps each vertex inside the domain and precisely into the origin. In conclusion we have to work
out the precise \textbf{critical radius} on which the vertices of the true octagonal  fundamental domain boundary for the group $\Gamma_8$ are
located. Considering  the images of the origin with respect to the eight generators has been any how useful. Indeed this has fixed the angular
phases of the true vertices although not the radius. We see how to use this information in the next subsection.
\subsubsection{Definition of the distance difference function in polar coordinates}
\begin{figure}[htb]
\begin{center}
\vskip 1cm
\includegraphics[width=70mm]{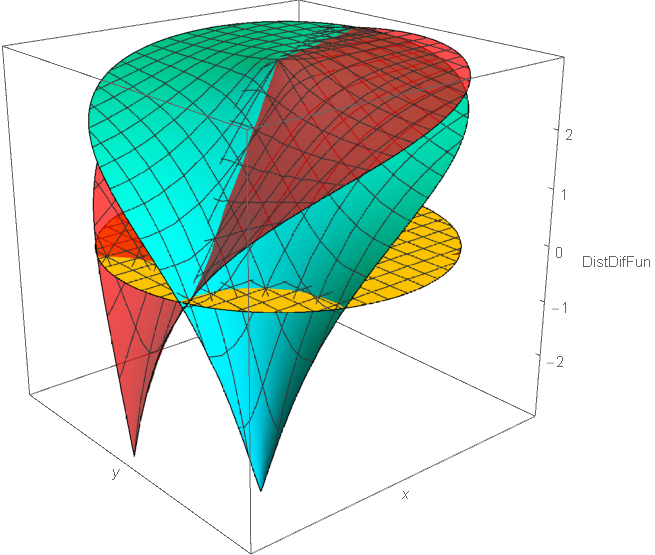}
\includegraphics[width=70mm]{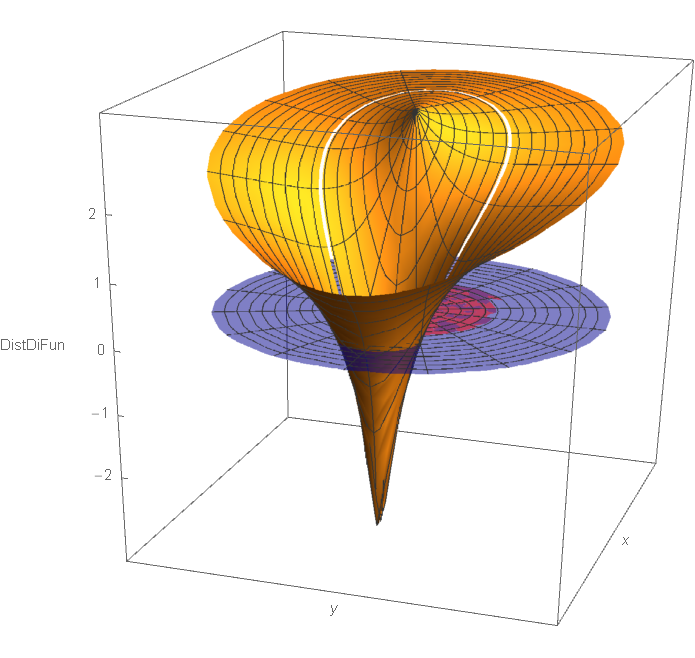}
\vskip 1cm
\caption{\label{gustocono} On the left, the comparison of the norm difference function plots for two adjacent generators of the Fuchsian group. In the image on the right, the full development of the norm difference function of the first Fuchsian generator in the disk of radius $\rho_0 = \sqrt{2 \left(\sqrt{2}-1\right)}$. Note the critical discontinuity at some radius $R_c <\rho_0$.   }
\end{center}
\end{figure}

Next we derive the explicit form of the \textbf{norm difference function} for the generator $\mathfrak{K}\mathfrak{F}_1$
expressed in terms of polar coordinates $(\rho,\phi)$ of a generic point inside the unit disk ($\rho <1$).
We find:
\begin{eqnarray}\label{Radphifun}
\mathrm{Nd}_{\mathfrak{K}\mathfrak{F}_1}(\rho,\phi )&=& \frac{1}{2}\left(\text{ArcCosh}\left[\left(\left(25569+18080\sqrt{2}+4\sqrt{2\left(40860031+28892405\sqrt{2}\right)}\right)\left(1+\rho
^4\right)\right.\right.\right.\nonumber\\
&&\left.\left.\left.+16\left(6392+4520\sqrt{2}+\sqrt{81720062+57784810\sqrt{2}}\right)\rho \left(1+\rho ^2\right)\text{Cos}[\phi ]\right.\right.\right.\nonumber\\
&&\left.\left.\left. +2\rho ^2\left(51251+36240\sqrt{2}+28\sqrt{6700558+4738010\sqrt{2}}\right.\right.\right.\right.\nonumber\\
&&\left.\left.\left.\left.+16\left(1591+1125\sqrt{2}+4\sqrt{2\left(158201+111865\sqrt{2}\right)}\right)\text{Cos}[2\phi
]\right)\right) {\Big{/}} \right.\right.\nonumber\\
&&\left.\left.\left(\left(5+4\sqrt{2}+2\sqrt{2\left(7+5\sqrt{2}\right)}\right)^2\left(-1+\rho ^2\right)^2\right)\right]+\text{Log}\left[\frac{(-1+\rho
)^2}{(1+\rho )^2}\right]\right)
\end{eqnarray}
To understand the behavior of the function, we plot it and we intersect its surface with the circular domain $\rho ^2\leq 2 \left(\sqrt{2}-1\right)$: see fig.\ref{gustocono}. In the image, we compare the norm difference function $\mathrm{Nd}_{\mathfrak{K}\mathfrak{F}_1}(p)$ with that of its adjacent generator, namely $\mathrm{Nd}_{\mathfrak{K}\mathfrak{F}_2(p)}$.
In the second image of fig.\ref{gustocono} we  plot the \textit{norm difference function}
for the Fuchsian generator $\mathrm{Nd}_{\mathfrak{K}\mathfrak{F}_1}(p)$ inside the chosen radial domain $\rho ^2\leq 2 \left(\sqrt{2}-1\right)$.  We immediately see that the function surface displays a topological circular discontinuity line, which is just the image of the boundary of a smaller radial domain $\rho^2\leq R_c^2$  where $R_c$ is a critical radius that we have to determine exactly. We will see how to do that in subsection \ref{belovo}.
\subsubsection{Fundamental domain of the first Fuchsian generator}
\begin{figure}[htb]
\begin{center}
\vskip 1cm
\includegraphics[width=80mm]{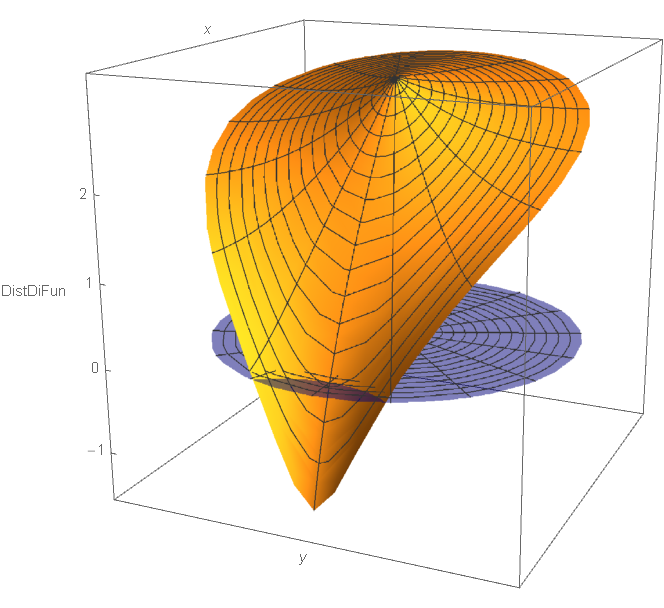}
\vskip 1cm
\caption{\label{linguaccia}} The fundamental domain of the generator $\mathfrak{K}\mathfrak{F}_1$ is the disk of radius $R_c$ deprived of the
small region delimited on one side by circle of radius $R_c$ on the other by the geodesic arc along which the \textit{norm difference function}
vanishes.
\end{center}
\end{figure}

In fig.\ref{linguaccia} we show the image of the \textit{norm difference function} over the radial domain $\rho ^2 \leq R_c^2$.
\par
The positivity subdomain of $\mathrm{Nd}_{\mathfrak{K}\mathfrak{F}_1}(p)$ is what is worth the name of
\textbf{fundamental domain of the Fuchsian generator} $\mathfrak{K}\mathfrak{F}_1$.
Formally  for each generator we can set:
\begin{equation}\label{misterbean}
\mathcal{F}_{\mathfrak{K}\mathfrak{F}_i} \, \equiv \, \langle z\in \mathbb{H}^2 \, \mid \,
\mathrm{Nd}_{\mathfrak{K}\mathfrak{F}_i}(z) \,  \geq \, 0 \rangle \quad \quad ; \quad i=1,\dots, 8
\end{equation}
and we have that
\begin{alignat}{4}\label{labentiasigna}
  &\forall z \in \left(\mathcal{F}_{\mathfrak{K}\mathfrak{F}_i} \, - \, \partial\mathcal{F}_{\mathfrak{K}\mathfrak{F}_i}\right)\quad &, \quad & \mathfrak{K}\mathfrak{F}_i(z) &\notin \mathcal{F}_{\mathfrak{K}\mathfrak{F}_i}\nonumber\\
  &\forall z \in \partial\mathcal{F}_{\mathfrak{K}\mathfrak{F}_i} \quad &, \quad &\mathfrak{K}\mathfrak{F}_i(z) &\in \partial\mathcal{F}_{\mathfrak{K}\mathfrak{F}_i}
\end{alignat}
The frontier of $\mathcal{F}_{\mathfrak{K}\mathfrak{F}_i}$ is obviously defined as:
\begin{equation}\label{marenavigerum}
  \partial\mathcal{F}_{\mathfrak{K}\mathfrak{F}_i} \, \equiv \, \langle z\in \mathbb{H}^2 \, \mid \, \mathrm{Nd}_{\mathfrak{K}\mathfrak{F}_i}(z) \,  = \, 0 \rangle \quad \quad ; \quad i=1,\dots 8
\end{equation}
As we clearly see from the figure this fundamental domain is the compact disk of radius $R_c$ deprived of the sector delimited by the geodesic along which the surface crosses the zero plane. The fundamental domain of the full Fuchsian group is going to be the intersection of all such fundamental domains of each of the eight generators:
\begin{equation}\label{exortumvidit}
 \mathcal{F}_{\Gamma_8} \, = \, \bigcap_{i=1}^{8}\, \mathcal{F}_{\mathfrak{K}\mathfrak{F}_i}
\end{equation}
\subsubsection{Determination of the critical radius}
\label{belovo}
\paragraph{1st step: deriving the angular part of vertex coordinates.} As we said above it appears that the location of the true vertices of the
fundamental domain of the Fuchsian group is at the same angles as that of dual vertices corresponding to the images of the origin under the action
of generators of the Fuchsian group, namely on the circle of radius $\rho_0 =\sqrt{2 \left(-1+\sqrt{2}\right)}$.  Hence we utilize the distance
function in radial coordinates of eq.(\ref{Radphifun}), localizing it on the radial value $\rho =\rho_0$  and we obtain an angular function
depending only on $\phi $. Since the dependence on $\phi $ is only through $\cos[\phi] $ and $\cos[2\phi]$ , we make the replacement
\(\left\{\text{Cos}[\phi ]\to q,\text{Cos}[2\phi]\to -1+2 q^2\right\}\) which holds true for q $<$ 1. We obtain a function:
\begin{eqnarray}\label{radiatore}
 F(q) &=& \frac{1}{2} \left(\text{ArcCosh}\left[\frac{1}{\left(4 \sqrt{2}+2 \sqrt{2 \left(5 \sqrt{2}+7\right)}+5\right)^2}\left(32 q \left(4 \left(15830
\sqrt{2}\right.\right.\right.\right.\right.\nonumber\\
&&\left.\left.\left.\left.\left.+4 \sqrt{2 \left(22148705 \sqrt{2}+31322999\right)}+22387\right) q+127765 \sqrt{2}
\right.\right.\right.\right.\nonumber\\
&&\left.\left.\left.\left.+4 \sqrt{2 \left(1442813945 \sqrt{2}+2040447049\right)}+180687\right)\right.\right.\right.\nonumber\\
&&\left.\left.\left. +3
\left(687440 \sqrt{2}+28 \sqrt{1704865210 \sqrt{2}+2411043502}+972187\right)\right)\right]\right.\nonumber\\
&&\left.+\text{Log}\left[80 \sqrt{2}-4 \sqrt{2 \left(565 \sqrt{2}+799\right)}+113\right]\right)
\end{eqnarray}
whose only zeros $F(q_{1,2}) =0$ are:
\begin{equation}\label{zerotti}
  q_1 \, = \, \frac{1}{4} \left(-3-\sqrt{2}\right)  \quad ; \quad q_2 \, = \,  \frac{1}{2}-\sqrt{2}
\end{equation}
The absolute value of the first zero $q_1$ is larger than $1$ so it has to be discarded. The absolute value of $q_2$ is instead smaller than $1$ which makes it the only viable solution. This means that we have unambiguously determined the angle:
\begin{equation}\label{fizero}
  \phi_0 \, = \, \text{ArcCos}\left[ \frac{1}{2}-\sqrt{2}\right]
\end{equation}
All the others angular variable just differ by a $2\pi/8$ so that we can write:
\begin{equation}\label{phienne}
  \phi_n \, = \, \phi_0 \, + \, n \, \frac{\pi}{4} \quad ; \quad (n=,0,1,\dots,7)
\end{equation}
\paragraph{2nd step: determination of the critical radius}
Inserting the value \(\phi =\phi _0\) in the norm distance function of eq.(\ref{Radphifun}) we obtain a radial function
\begin{equation}\label{calzone}
\mathcal{R}_0(\rho)\, \equiv \,\mathrm{Nd}_{\mathfrak{K}\mathfrak{F}_1}(\rho,\phi_0)
\end{equation}
which by construction vanishes at $\rho=\sqrt{2 \left(-1+\sqrt{2}\right)}$. If we insert instead in  $\mathrm{Nd}_{\mathfrak{K}\mathfrak{F}_1}(\rho,\phi)$ the next vertex angle, namely
value $\phi =\phi _0 +\pi/4$ obtain a function:
\begin{equation}\label{laguerre}
\mathcal{R}_1(\rho)\, = \, \mathrm{Nd}_{\mathfrak{K}\mathfrak{F}_1}\left(\rho,\phi_0+\frac{\pi}{4}\right)
\end{equation}
 that as we  know  vanishes at a smaller value of $\rho $. The zero of $\mathcal{R}_1(\rho)$ in the interval $[0,1]$  is the looked for critical radius $R_c$.
\par
The very nice thing is that
\begin{enumerate}
  \item  the zero $R_c$ of $\mathcal{R}_1(\rho)$ in the interval $[0,1]$ is unique,
  \item notwithstanding the complicated structure of the equation, $R_c$ can be determined exactly in terms of radicals.
\end{enumerate}
 This is the beauty and the usefulness of the Bolza tiling of the hyperbolic plane. All vertices of the tessellation are uniquely defined in terms of square roots or double roots of rational functions of the unique extension of the rational field by means of $\sqrt{2}$. Moreover, as we will shortly show, the entire hyperbolic plane can be tiled by quadrangular squares arranged in the same way as in a square tessellation of the Euclidean plane.
 \par
 The equation $\mathcal{R}_1(\rho)\, = \, 0$ has the following formal structure:
 \begin{equation}\label{formalina}
   \text{ArcCosh}[Z[\rho ]] + \text{Log}[W[\rho ]] = 0
 \end{equation}
 where $Z[\rho]$ is a quartic polynomial in $\rho$ and $W[\rho]$ is a rational function of degree $2$ in $\rho$ that the reader can work out from  eq.( \ref{Radphifun}) and definition (\ref{laguerre}). Eq.(\ref{formalina}) implies that:
 \begin{equation}\label{Z_rho}
   Z[\rho ]=\frac{1+W[\rho ]^2}{2W[\rho ]}=\frac{1+6 \rho ^2+\rho ^4}{\left(-1+\rho ^2\right)^2}
\end{equation}
Reducing eq.(\ref{Z_rho}) to the same denominator one finds that the equation to be solved reduces to the following quartic equation
$\mathfrak{P}(\rho) =0$ where:
\begin{eqnarray}\label{balalayca}
  \mathfrak{P}(\rho ) & = & 8 \left(2 \left(7+5 \sqrt{2}\right)-\sqrt{2 \left(1598+1130 \sqrt{2}+\sqrt{7 \left(400559+283238 \sqrt{2}\right)}\right)}
\rho \right.\nonumber\\
&&\left.+2 \left(28+20 \sqrt{2}+\sqrt{7 \left(31+22 \sqrt{2}\right)}\right) \rho ^2-\sqrt{2 \left(1598+1130 \sqrt{2}+\sqrt{7 \left(400559+283238 \sqrt{2}\right)}\right)}
\rho ^3\right.\nonumber\\
&&\left.+2 \left(7+5 \sqrt{2}\right) \rho ^4\right)
\end{eqnarray}
The polynomial $\mathfrak{P}(\rho)$ has two complex conjugate roots and two real roots, both positive, one larger than 1, the second smaller then 1. Hence the critical radius is completely fixed in terms of radicals and it has the  explicit appearance
displayed below:
\begin{eqnarray}\label{Rcvalue}
  R_c & = & \frac{1}{8} \left(5 \sqrt{2}-7\right) \left(\text{ZU}-2
   \sqrt{\text{TU}+\sqrt{\text{XU} \times \text{YU}}}\right) \nonumber\\
   \text{ZU} & = & \sqrt{2 \left(1598+1130 \sqrt{2}+\sqrt{7 \left(400559+283238
   \sqrt{2}\right)}\right)}\nonumber\\
   &&+\sqrt{2 \left(14+10 \sqrt{2}-56 \sqrt{7 \left(31+22
   \sqrt{2}\right)}-40 \sqrt{434+308 \sqrt{2}}+\sqrt{7 \left(400559+283238
   \sqrt{2}\right)}\right)}\nonumber\\
   \text{TU} &=& -778-550 \sqrt{2}-28 \sqrt{7 \left(31+22 \sqrt{2}\right)}-20 \sqrt{434+308
   \sqrt{2}}+\sqrt{7 \left(400559+283238 \sqrt{2}\right)}\nonumber\\
   \text{XU} & = & 14+10 \sqrt{2}-56 \sqrt{7 \left(31+22 \sqrt{2}\right)}-40 \sqrt{434+308
   \sqrt{2}}+\sqrt{7 \left(400559+283238 \sqrt{2}\right)}\nonumber\\
   \text{YU} & = & 1598+1130 \sqrt{2}+\sqrt{7 \left(400559+283238 \sqrt{2}\right)}
   \end{eqnarray}
   The numerical value of $R_c$ is $R_c \approx 0.799785$
\subsection{Vertices of the Bolza fundamental domain}
What we have obtained in the  previous subsections provides enough information to write down explicitly all the vertices of the fundamental octagonal domain of the Fuchsian group, whose copies will tile the entire  hyperbolic plane $\mathbb{H}^2$.
Indeed the vertices are listed explicitly in table \ref{vertilli}. Furthermore whatever regular  tiling  by means of geodesics we manage to do of the very same fundamental domain will be consistently repeated through the whole hyperbolic plane. As we are going to see there are two principal tilings of the fundamental domain that we can consider, one by means of triangles, the other by means of quadrangles. In any case to this effect we still need an additional information, namely the middle points of the geodesic arcs of the fundamental octagon. This is what we do in the next  subsection.
\par
\begin{table}
\(\mathbf{p}_0\text{ = }\left\{\left(\frac{1}{2}-\sqrt{2}\right) R_c ,\frac{1}{2} \sqrt{-5+4 \sqrt{2}} R_c \right\}\\
\\
\mathbf{p}_1\text{ = }\left\{-\frac{1}{4} \left(4-\sqrt{2}+\sqrt{-10+8 \sqrt{2}}\right) R_c ,\frac{1}{4} \left(-4+\sqrt{2}+\sqrt{-10+8 \sqrt{2}}\right) R_c
\right\}\\
\\
\mathbf{p}_2\text{ = }\left\{-\frac{1}{2} \sqrt{-5+4 \sqrt{2}} R_c ,\left(\frac{1}{2}-\sqrt{2}\right) R_c \right\}\\
\\
\mathbf{p}_3\text{ = }\left\{-\frac{1}{4} \left(-4+\sqrt{2}+\sqrt{-10+8 \sqrt{2}}\right) R_c ,-\frac{1}{4} \left(4-\sqrt{2}+\sqrt{-10+8 \sqrt{2}}\right)
R_c \right\}\\
\\
\mathbf{p}_4\text{ = }\left\{\left(-\frac{1}{2}+\sqrt{2}\right) R_c ,-\frac{1}{2} \sqrt{-5+4 \sqrt{2}} R_c \right\}\\
\\
\mathbf{p}_5\text{ = }\left\{\frac{1}{4} \left(4-\sqrt{2}+\sqrt{-10+8 \sqrt{2}}\right) R_c ,-\frac{1}{4} \left(-4+\sqrt{2}+\sqrt{-10+8 \sqrt{2}}\right) R_c
\right\}\\
\\
\mathbf{p}_6\text{ = }\left\{\frac{1}{2} \sqrt{-5+4 \sqrt{2}} R_c ,\left(-\frac{1}{2}+\sqrt{2}\right) R_c \right\}\\
\\
\mathbf{p}_7\text{ = }\left\{\frac{1}{4} \left(-4+\sqrt{2}+\sqrt{-10+8 \sqrt{2}}\right) R_c ,\frac{1}{4} \left(4-\sqrt{2}+\sqrt{-10+8 \sqrt{2}}\right) R_c
\right\}\)
\caption{\label{vertilli} In this table we display the explicit disk coordinates for the vertices of fundamental domain of the Fuchsian group $\Gamma_8$ corresponding to the uniformization of the Bolza surface. }
\end{table}
\subsubsection{Middle points of the geodesic arcs}
In order to construct the middle points of the geodesic sides of our octagon we consider the following general problem.
Suppose that  we have two points, in the hyperbolic plane disk model, that are located on the same circle of radius $\rho $ at two different angles $\phi $ and $\psi $, namely
\begin{equation}\label{puntinilli}
  p_1=\{\rho \, \text{Cos}[\phi ],\rho \, \text{Sin}[\phi ] \}\quad ; \quad  p_2=\{\rho \, \text{Cos}[\psi ],\text{$\rho$}\, \text{Sin}[\psi ]\}
\end{equation}
the geodesic arc that joins them is on a circle with center $c$ on the boundary of the disk, explicitly determined as
\begin{equation}\label{centrillo}
  c =\left\{\text{Cos}\left[\frac{\phi +\psi }{2}\right],\text{Sin}\left[\frac{\phi +\psi }{2}\right]\right\}
\end{equation}
of radius
\begin{equation}\label{radillo}
  R=\sqrt{1+\rho ^2\, -\, 2 \, \rho \, \text{Cos}\left[\frac{\phi -\psi }{2}\right]}
\end{equation}
The middle point of the geodesic arc joining the two points \(p_{1,2}\) is exactly at
\begin{equation}\label{midillo}
  p_{12}=\left\{\rho ^*\text{Cos}\left[\frac{\phi +\psi }{2}\right],\rho ^*\text{Sin}\left[\frac{\phi +\psi }{2}\right]\right\}
\end{equation}
where
\begin{equation}\label{stardillo}
 \rho ^* = (1-R)
\end{equation}
\begin{table}
\(\mathbf{q}_1\text{ = }\left\{-\frac{1}{2} \sqrt{\sqrt{-\frac{77}{2}+28 \sqrt{2}}+\frac{7}{\sqrt{2}}-2} \rho ^*,\frac{1}{4} \left(\sqrt{-2+3 \sqrt{2}}-\sqrt{26-17
\sqrt{2}}\right) \rho ^*\right\}\\
\\
\mathbf{q}_2\text{ = }\left\{-\frac{1}{2} \sqrt{\sqrt{-\frac{77}{2}+28 \sqrt{2}}-\frac{7}{\sqrt{2}}+6} \rho ^*,-\frac{\rho ^*}{2 \sqrt{\frac{2}{-\sqrt{14
\left(-11+8 \sqrt{2}\right)}+7 \sqrt{2}-4}}}\right\}\\
\\
\mathbf{q}_3\text{ = }\left\{\frac{1}{4} \left(\sqrt{26-17 \sqrt{2}}-\sqrt{-2+3 \sqrt{2}}\right) \rho ^*,-\frac{1}{2} \sqrt{\sqrt{-\frac{77}{2}+28 \sqrt{2}}+\frac{7}{\sqrt{2}}-2}
\rho ^*\right\}\\
\\
\mathbf{q}_4\text{ = }\left\{\frac{1}{4} \sqrt{-2 \sqrt{14 \left(-11+8 \sqrt{2}\right)}+14 \sqrt{2}-8} \rho ^*,-\frac{1}{2} \sqrt{\sqrt{-\frac{77}{2}+28 \sqrt{2}}-\frac{7}{\sqrt{2}}+6}
\rho ^*\right\}\\
\\
\mathbf{q}_5\text{ = }\left\{\frac{1}{4} \left(\sqrt{10+\sqrt{2}}+\sqrt{-18+13 \sqrt{2}}\right) \rho ^*,\frac{1}{4} \left(\sqrt{26-17 \sqrt{2}}-\sqrt{-2+3
\sqrt{2}}\right) \rho ^*\right\}\\
\\
\mathbf{q}_6\text{ = }\left\{\frac{1}{2} \sqrt{\sqrt{-\frac{77}{2}+28 \sqrt{2}}-\frac{7}{\sqrt{2}}+6} \rho ^*,\frac{1}{4} \sqrt{-2 \sqrt{14 \left(-11+8 \sqrt{2}\right)}+14
\sqrt{2}-8} \rho ^*\right\}\\
\\
\mathbf{q}_7\text{ = }\left\{\frac{1}{4} \left(\sqrt{-2+3 \sqrt{2}}-\sqrt{26-17 \sqrt{2}}\right) \rho ^*,\frac{1}{4} \left(\sqrt{10+\sqrt{2}}+\sqrt{-18+13
\sqrt{2}}\right) \rho ^*\right\}\\
\\
\mathbf{q}_8\text{ = }\left\{-\frac{\rho ^*}{2 \sqrt{\frac{2}{-\sqrt{14 \left(-11+8 \sqrt{2}\right)}+7 \sqrt{2}-4}}},\frac{1}{2} \sqrt{\sqrt{-\frac{77}{2}+28
\sqrt{2}}-\frac{7}{\sqrt{2}}+6} \rho ^*\right\}\)
\caption{\label{quilli} The explicit coordinatees of the middle points of all geodesic edges of the fundamental domain for the
Fuchsian group $\Gamma_8$, corresponding to the uniformization of the Bolza surface.}
\end{table}
In our case we have:
\begin{equation}\label{nashpapillo}
  \phi  = \text{ArcCos}\left[\frac{1}{2}-\sqrt{2}\right]+n\frac{\pi }{4}\quad ; \quad \psi  = \text{ArcCos}\left[\frac{1}{2}-\sqrt{2}\right]+(n+1)\frac{\pi}{4}
\end{equation}
Hence we obtain:
\begin{eqnarray}
  \frac{1}{2}(\phi  +\psi ) &=&\text{ArcCos}\left[\frac{1}{2}-\sqrt{2}\right]+n\frac{\pi }{4}+\frac{\pi }{8}\\
  \frac{1}{2}(\phi  -\psi ) &=&-\frac{\pi }{8}
\end{eqnarray}
Using trigonometric identities we get
\begin{equation}\label{rostarro}
  \rho^\star \, = \, 1-\sqrt{R_c^2-\sqrt{2+\sqrt{2}} R_c+1}
\end{equation}
and we determine all the middle points named $\mathbf{q}_i$ as displayed in table \ref{quilli}
\subsection{Tessellation of the Bolza fundamental domain}
\begin{figure}
\begin{center}
\vskip 1cm
\includegraphics[width=80mm]{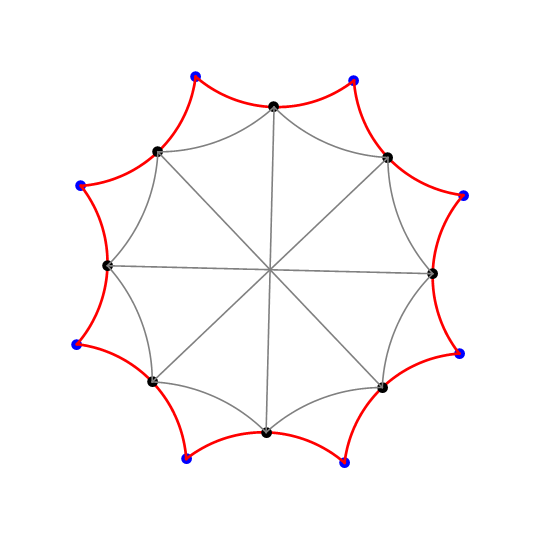}
\includegraphics[width=80mm]{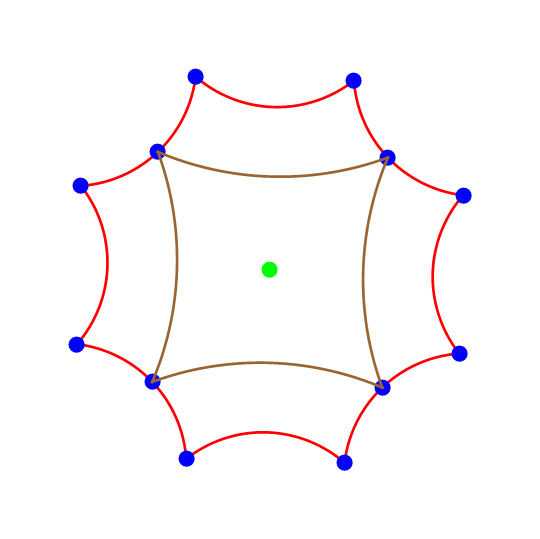}
\vskip 1cm
\caption{\label{crullino} On the left the octagonal tessellation in terms of triangles, on the right the quadrangular tessellation of the fundamental domain}
\end{center}
\end{figure}
We come now to the tessellations of the fundamental domain that will be repeated over the all hyperbolic plane by the action of the generators. On
the left of fig.\ref{crullino} we see the octagonal tessellation in terms of 16 triangles. In this tessellation, one utilizes all the middle points
as vertices and draws all the geodesic internal sides that go from the center to the middle points of the boundary and all the geodesics that join
every middle point to the next one. The result is that the entire fundamental domain is covered by 16 triangles (faces), and there are a total of 16 internal edges plus 16 edges on the boundary (32 edges) and a total of 17 vertices (counting also the center). Since the edges are pair-wise
identified when we consider the tessellation of the compact surface, the number of exterior edges has to be divided by 2, and the number of vertices on the boundary has to be identified accordingly. The 8 vertices of an octagon are identified as just one single vertex, while the 8 middle point vertices
are pair-wise identified and make a total of 4 independent vertices.  The result is that in the tessellation of the compact surface using triangles, we have 16 triangular faces, 24 edges, and 6 vertices. This makes, as it should for genus $g=2$, Euler characteristic $\chi \, = \, 16-24+6 \, = -2$. In the
quadrangular tessellation, instead, one uses only half of the middle points and has a total of 5 quadrangular faces, 10 edges,  and 3 vertices, which makes once again $\chi \, = \, 5-10+3 \, = -2$. This is a powerful check since for a genus $g$ surface one has $\chi=-2(g-1)$.
\begin{figure}
\begin{center}
\vskip 1cm
\includegraphics[width=80mm]{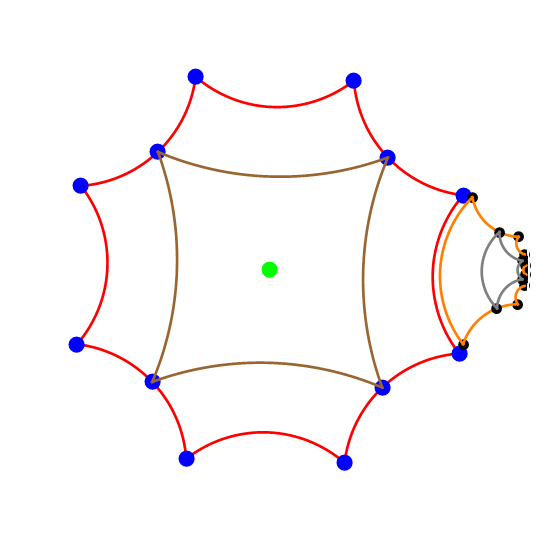}
\includegraphics[width=80mm]{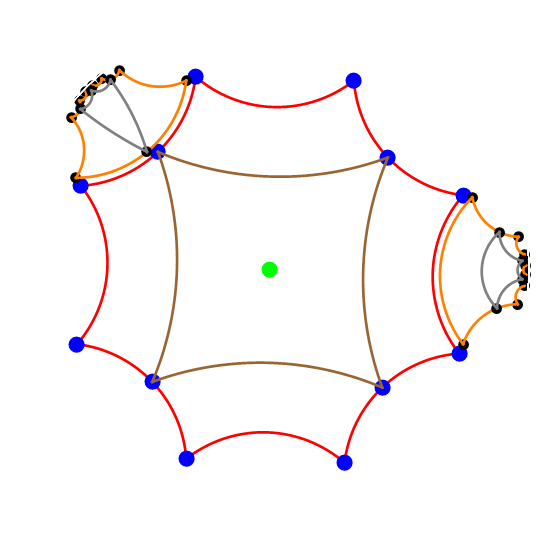}
\vskip 1cm
\caption{\label{prigotto} In this image, we show how the hyperbolic plane starts being tiled by images of the fundamental domain while using the generators of the Fuchsian group. Each of the 8 generators maps the entire fundamental domain into an adjacent copy of it that shares one side with its preimage. }
\end{center}
\end{figure}
\section{Harmonic Expansions both on the base manifold and on the total space}
\label{spinatore}
In the introductory subsection \ref{Tsvettorifibrati} we stated that the intrinsic description of the vector-bundle section (\ref{sezionatore})
is given by the coefficients of the harmonic expansion (\ref{harmexpgenerale}), yet there is an alternative viewpoint  which, while
devising a learning algorithm can prove more efficient. As we emphasized there, the section is a hypersurface in the total bundle space, in our case
$\mathrm{SO(1,3)/SO(3)}$, of dimension equal to the Tits Satake submanifold, in our case $\mathrm{SO(1,2)/SO(2)}$.
The codimension of the hypersurface is the difference between the
dimension of the total space and the dimension of the base space, namely it equals the rank of the vector bundle: \textit{i.d.} $r(q-1)$.
In our case the codimension is just one. Generically a hypersurface of codimension $r(q-1)$ is described as the vanishing locus of $r(q-1)$
independent functions on the total space:
\begin{equation}\label{rettilodattero}
  \mathfrak{X}\, = \, \left\langle p \in \mathcal{M}^{[r,q]}\, \left\| \right.\, \left\{\begin{array}{lcl}
                                                                 \Phi_{1}(p) & = & 0 \\
                                                                 \Phi_{2}(p) & = & 0 \\
                                                                 \vdots & \vdots & \vdots \\
                                                                 \Phi_{r(q-1)}(p) & = & 0
                                                               \end{array}\right. \right \rangle
\end{equation}
Each of the functions $\Phi_{a}(p)$ admits its own harmonic expansion:
\begin{equation}\label{armoniosefunzie}
  \Phi_{a}(p)\, = \, \sum_{\hat{D}(\mathrm{U})\in \mathfrak{S}[r,q_{min}]}\,\sum_{\ell=0}^{d_{\hat{D}}} \, \hat{c}_{\hat{D},a}^\ell \,
  \pmb{\mathfrak{harm}}^{\hat{D}}_\ell(p)  \quad ; \quad \mathrm{U} \equiv \mathrm{SO(r,r+q)}
\end{equation}
The number of coefficients to be determined is equivalent in eq. (\ref{harmexpgenerale}) and in equation (\ref{armoniosefunzie}).
Indeed the coefficients in equation (\ref{harmexpgenerale}) are boldfaced since they are vectors of dimension $r(q-1)$ just as those in
equation \ref{armoniosefunzie}. The relevant difference is that harmonics $\hat{D}$ in the second equation are those associated
with irreducible representation of the full group $\mathrm{SO(r,r+q)}$ while those in eq.(\ref{harmexpgenerale}) are those associated
with irreducible representations of the Tits Satake subgroup $\mathrm{SO(r,r+1)}$. Determining $\hat{c}_{\hat{D},a}$ or $\boldsymbol{c}_{D}^\ell$ is
equivalent. Indeed from one expansion one can derive, with some algebraic work the other and viceversa.
\par
In our case we have found that the relation between the two kind of harmonics can be optimized if we use in both cases
the spinor representation instead of the vector one. Furthermore, the spinor representation reduces the algebraic complexity of the basic harmonic
functions out of which we construct all the other by tensor product. For this reason, we begin by rewriting the geometry of the manifold
$\mathrm{SO(1,3)/SO(3)}$ in terms of the double covering simply connected manifold $\mathrm{Spin(1,3)/SU(2)}$.
\subsection{The manifold \texorpdfstring{$\mathrm{Spin(1,3)/SU(2)}$}{Spin(1,3)/SU(2)}}
Since the charge conjugation matrix for spinors in $D=(1+3)$ is antisymmetric (see \cite{castdauriafre} Vol. 1), the spinor group
$\mathrm{Spin(1,3)}\subset \mathrm{Sp(4,\mathbb{R})}$ is a subgroup of the 4-dimensional real symplectic group. The latter, as thoroughly reviewed
in \cite{pgtstheory}, is the double covering of $\mathrm{SO(2,3)}$ and corresponds to its spinor representation. Hence in the carrier space of
$\mathrm{Sp(4,\mathbb{R})}$ do
coexist both the 10-dimensional Lie algebra $\so(2,3)$ and the 6-dimensional algebra $\so(1,3)$ which is an obvious subalgebra of the first.
The trading channel between the vector and the spinor representation is provided by a well adapted basis of \textit{gamma matrices} satisfying
the Clifford algebra either in the first or in the second line of the next equation:
\begin{equation}
\begin{array}{rclcrclcrcl}
 \left\{\Gamma_A , \Gamma_B\right\} & = & 2 \eta^t_{AB} \, \mathbf{1}_{4\times 4} & ; &  A,B & =& 1,2,\dots,5 \quad &; & \quad \eta^t & =& \left(
\begin{array}{ccccc}
 0 & 0 & 0 & 0 & 1 \\
 0 & 0 & 0 & 1 & 0 \\
 0 & 0 & 1 & 0 & 0 \\
 0 & 1 & 0 & 0 & 0 \\
 1 & 0 & 0 & 0 & 0 \\
\end{array}\right) \\
\null&\null&\null&\null&\null&\null&\null&\null&\null&\null&\null\\
\left\{\Sigma_a , \Sigma_b\right\} & = & 2 \eta^t_{ab} \, \mathbf{1}_{4\times 4} & ; &  a,b & =& 1,2,\dots,4 \quad &; & \quad \eta^t & =& \left(
\begin{array}{cccc}
 0 & 0 & 0 & 1  \\
 0 & 0 & 1 & 0 \\
 0 & 1 & 0 & 0 \\
 1 & 0 & 0 & 0  \\
\end{array}
\right)
\end{array}\label{cliffone}
\end{equation}
Clearly with a suitable subset of the first $5$ gamma matrices $\Gamma_A$ satisfying the Clifford algebra
in the first line we might obtain $4$ gamma matrices $\Sigma_a$ satisfying the Clifford algebra in the second line, yet we will not do that and
we rather utilize two sets of gamma matrices, one well adapted to the $\so(2,3)$ Lie algebra (the first), the other well-adapted to the $\so(1,3)$
Lie algebra, the second.  The five $\Gamma_A$ are displayed in eq. (7.43) of \cite{pgtstheory} and we report them here for reader's convenience.
\par
\begin{equation}
\label{gammoni}
\begin{array}{|ccc|ccc|}
  \hline
     \Gamma_1 & = & \left(
\begin{array}{cccc}
 0 & 0 & 0 & \sqrt{2} \\
 0 & 0 & -\sqrt{2} & 0 \\
 0 & 0 & 0 & 0 \\
 0 & 0 & 0 & 0 \\
\end{array}
\right) & \Gamma_2 & = & \left(
\begin{array}{cccc}
 0 & \sqrt{2} & 0 & 0 \\
 0 & 0 & 0 & 0 \\
 0 & 0 & 0 & 0 \\
 0 & 0 & \sqrt{2} & 0 \\
\end{array}
\right) \\
     \hline
     \Gamma_3 & = & \left(
\begin{array}{cccc}
 1 & 0 & 0 & 0 \\
 0 & -1 & 0 & 0 \\
 0 & 0 & 1 & 0 \\
 0 & 0 & 0 & -1 \\
\end{array}
\right) & \Gamma_4 & = & \left(
\begin{array}{cccc}
 0 & 0 & 0 & 0 \\
 \sqrt{2} & 0 & 0 & 0 \\
 0 & 0 & 0 & \sqrt{2} \\
 0 & 0 & 0 & 0 \\
\end{array}
\right) \\
     \hline
     \Gamma_5 & = & \left(
\begin{array}{cccc}
 0 & 0 & 0 & 0 \\
 0 & 0 & 0 & 0 \\
 0 & -\sqrt{2} & 0 & 0 \\
 \sqrt{2} & 0 & 0 & 0 \\
\end{array}
\right) & \null & \null & \null\\
\hline
   \end{array}
\end{equation}
The second set is displayed in the next equation (\ref{sigmoni})
 \begin{equation}
\label{sigmoni}
\begin{array}{|ccc|ccc|}
  \hline
     \Sigma_1 & = & \left(
\begin{array}{cccc}
 0 & 0 & 0 & 0 \\
 -\sqrt{2} & 0 & 0 & 0 \\
 0 & 0 & 0 & 0 \\
 0 & 0 & \sqrt{2} & 0 \\
\end{array}
\right) & \Sigma_2 & = & \left(
\begin{array}{cccc}
 1 & 0 & 0 & 0 \\
 0 & -1 & 0 & 0 \\
 0 & 0 & -1 & 0 \\
 0 & 0 & 0 & 1 \\
\end{array}
\right) \\
     \hline
     \Sigma_3 & = & \left(
\begin{array}{cccc}
 0 & 0 & -1 & 0 \\
 0 & 0 & 0 & -1 \\
 -1 & 0 & 0 & 0 \\
 0 & -1 & 0 & 0 \\
\end{array}
\right) & \Sigma_4 & = & \left(
\begin{array}{cccc}
 0 & -\sqrt{2} & 0 & 0 \\
 0 & 0 & 0 & 0 \\
 0 & 0 & 0 & \sqrt{2} \\
 0 & 0 & 0 & 0 \\
\end{array}
\right) \\
     \hline
     \end{array}
\end{equation}
In both cases the generators of the corresponding Lie algebra are given by the commutators of the corresponding gamma matrices:
\begin{equation}\label{gennicoli}
  J^{Sp(4)}_{AB} \, = \, \ft{1}{4} \left[\Gamma_A \, , \, \Gamma_B\right] \quad ; \quad
  J^{Spin(1,3)}_{ab} \, = \, \ft{1}{4} \left[\Sigma_a \, , \, \Sigma_b\right]
\end{equation}
The 10 generators $J^{Sp4}_{AB}$ satisfy the condition:
\begin{equation}\label{carignello}
  C\,J^{Sp(4)}_{AB}+ \left(J^{Sp(4)}_{AB}\right)^T \, C \, = \, \mathbf{0}_{4\times 4} \quad ; \quad  C\, = \, \left(
\begin{array}{cccc}
 0 & 0 & 1 & 0 \\
 0 & 0 & 0 & 1 \\
 -1 & 0 & 0 & 0 \\
 0 & -1 & 0 & 0 \\
\end{array}
\right)
\end{equation}
where $C$ is an antisymmetric matrix that squares to $-\mathbf{1}_{4\times 4}$. This shows that $J^{Sp(4)}_{AB}$ belong to the Lie Algebra
$\sym(4,\mathbb{R})$ and actually constitute a complete basis for it.
\par
The  6 generators $J^{Spin(1,3)}_{ab}$ satisfy  a completely analogous condition with a different antisymmetric matrix:
\begin{equation}\label{cavernicoli}
  \tilde{C}\,J^{Spin(1,3)}_{ab}+ \left(J^{Spin(1,3)}_{ab}\right)^T \, \tilde{C} \, = \, \mathbf{0}_{4\times 4} \quad ; \quad  \tilde{C}\, = \,
  \left(
\begin{array}{cccc}
 0 & -\frac{1}{\sqrt{2}} & 0 & \frac{1}{\sqrt{2}} \\
 \frac{1}{\sqrt{2}} & 0 & -\frac{1}{\sqrt{2}} & 0 \\
 0 & \frac{1}{\sqrt{2}} & 0 & \frac{1}{\sqrt{2}} \\
 -\frac{1}{\sqrt{2}} & 0 & -\frac{1}{\sqrt{2}} & 0 \\
\end{array}
\right)
\end{equation}
Also $\tilde{C}$ squares to $-\mathbf{1}_{4\times 4}$, so that we see that also $J^{Spin(1,3)}_{ab}$ are symplectic, yet in a different basis from that
of the generators $J^{Sp4}_{AB}$. We can easily find the conversion matrix. Indeed we find that:
\begin{equation}\label{Qumma}
  \mathcal{Q} \, = \, \left(
\begin{array}{cccc}
 -\frac{\sqrt{1+\sqrt{2}}}{2^{3/4}} & 0 & \frac{1}{2^{3/4}
   \sqrt{1+\sqrt{2}}} & 0 \\
 0 & \frac{\sqrt{\sqrt{2}-1}}{2^{3/4}} & 0 & \frac{1}{2^{3/4}
   \sqrt{\sqrt{2}-1}} \\
 0 & -\frac{\sqrt{1+\sqrt{2}}}{2^{3/4}} & 0 & \frac{1}{2^{3/4}
   \sqrt{1+\sqrt{2}}} \\
 \frac{\sqrt{\sqrt{2}-1}}{2^{3/4}} & 0 & \frac{1}{2^{3/4}
   \sqrt{\sqrt{2}-1}} & 0 \\
\end{array}
\right)
\end{equation}
satisfy the conditions:
\begin{equation}\label{trasmutazia}
  \mathcal{Q}\cdot \mathcal{Q}^T \, = \, \mathbf{1}_{4\times 4} \quad ; \quad \mathcal{Q}\cdot \tilde{C} \cdot \mathcal{Q}^T \, = \, C
\end{equation}
It means that $\mathcal{Q} \in \mathrm{SO(4)}$ is an orthogonal matrix which transforms the symplectic invariant matrix $\tilde{C}$ into
the symplectic invariant matrix $C$. Next we can consider inside the Lie Algebra generated by the $J^{Spin(1,3)}_{ab}$ generators the
$3$-dimensional solvable subalgebra whose $3$ generators take the following form:
\begin{eqnarray}\label{krikko}
  SpT_1 & = & \left(
\begin{array}{cccc}
 \frac{1}{2} & 0 & 0 & 0 \\
 0 & -\frac{1}{2} & 0 & 0 \\
 0 & 0 & \frac{1}{2} & 0 \\
 0 & 0 & 0 & -\frac{1}{2} \\
\end{array}
\right)  \quad ; \quad SpT_2 \, = \, \left(
\begin{array}{cccc}
 0 & \frac{1}{2} & 0 & 0 \\
 0 & 0 & 0 & 0 \\
 0 & 0 & 0 & \frac{1}{2} \\
 0 & 0 & 0 & 0 \\
\end{array}
\right) \nonumber\\
  SpT_3 & = & \left(
\begin{array}{cccc}
 0 & 0 & 0 & \frac{1}{2} \\
 0 & 0 & 0 & 0 \\
 0 & -\frac{1}{2} & 0 & 0 \\
 0 & 0 & 0 & 0 \\
\end{array}
\right)
\end{eqnarray}
Under the map between group (or Lie algebra) elements in the spinor representation $\mathcal{S}$ and the
corresponding group (or Lie algebra elements) $\mathcal{O}$ in the vector representation:
\begin{equation}\label{cogrutto}
  \mathcal{O}_a^{\phantom{a}b}[\mathcal{S}] \, = \, \ft 14 \, \text{Tr} \, \left( \mathcal{S}^{-1} \, \Sigma_a \, \mathcal{S}\, \Sigma_{c}\right)
  \, \eta^{cb}
\end{equation}
which is the analogue of eq. (7.46) of \cite{pgtstheory} and expresses the concrete form of the double covering, the three generators
in eq.(\ref{krikko}) exactly match the three generators of the solvable subalgebra of $\mathrm{SO(1,3)}$. It follows that:
\begin{eqnarray}
\label{spinLcosrep}
  \mathbb{L}_s\, &\equiv & \exp[w_1\,SpT_1 ] \cdot \exp[w_2\,SpT_2 ] \cdot \exp[w_3\,SpT_3 ] \nonumber\\
   &= & \left(
\begin{array}{cccc}
 e^{\frac{w_1}{2}} & \frac{1}{2} e^{\frac{w_1}{2}} w_2 & 0 &
   \frac{1}{2} e^{\frac{w_1}{2}} w_3 \\
 0 & e^{-\frac{w_1}{2}} & 0 & 0 \\
 0 & -\frac{1}{2} e^{\frac{w_1}{2}} w_3 & e^{\frac{w_1}{2}} &
   \frac{1}{2} e^{\frac{w_1}{2}} w_2 \\
 0 & 0 & 0 & e^{-\frac{w_1}{2}} \\
\end{array}
\right)
\end{eqnarray}
is the explicit form of the solvable coset representative in the spinor representation. Indeed using the map (\ref{cogrutto}) we get:
\begin{equation}\label{porcinaconcrauti}
\mathbb{L}_v \, = \, \mathcal{O}[\mathbb{L}_s]  \, = \,  \left(
\begin{array}{cccc}
 e^{w_1} & \frac{e^{w_1} w_2}{\sqrt{2}} & \frac{e^{w_1}
   w_3}{\sqrt{2}} & -\frac{1}{4} e^{w_1} \left(w_2^2+w_3^2\right)
   \\
 0 & 1 & 0 & -\frac{w_2}{\sqrt{2}} \\
 0 & 0 & 1 & -\frac{w_3}{\sqrt{2}} \\
 0 & 0 & 0 & e^{-w_1} \\
\end{array}
\right)
\end{equation}
which is the standard form of the upper triangular coset representative for $\mathrm{SO(1,3)/SO(3)}$ as displayed in eq. (7.22) of \cite{pgtstheory}
(it suffices to rename $\Upsilon_1 = w_1,\Upsilon_{2,1}=w_2,\Upsilon_{2,2}=w_3$).
\subsubsection{Reduction to the Tits Satake submanifold}
Starting from eq.(\ref{spinLcosrep}) it is very easy to perform the Tits Satake projection. It suffices to set $w_3 = 0$.
We obtain:
\begin{equation}\label{bisturi}
  \mathbb{L}_s^{TS} \, = \, \left(
\begin{array}{cccc}
 e^{\frac{w_1}{2}} & \frac{1}{2} e^{\frac{w_1}{2}} w_2 & 0 & 0 \\
 0 & e^{-\frac{w_1}{2}} & 0 & 0 \\
 0 & 0 & e^{\frac{w_1}{2}} & \frac{1}{2} e^{\frac{w_1}{2}} w_2 \\
 0 & 0 & 0 & e^{-\frac{w_1}{2}} \\
\end{array}
\right)
\end{equation}
Equation (\ref{bisturi}) displays a  feature that is very important for the relation between  harmonics
of the total space and  harmonics of the Tits-Satake submanifold which is the base manifold of the vector bundle. In the spinor representation,
as we see, the $4\times 4$ coset representative is block diagonal and repeats twice the same $2\times 2$ matrix. Such block diagonal property is
maintained also in the product. As we know the fundamental harmonics are the independent matrix elements of the symmetric matrix $\mathcal{M}^{AB}$
defined below (it works the same in the spinor and in the vector representation):
\begin{equation}\label{grignolio}
 \mathcal{M}_{s,v} \, \equiv \, \mathbb{L}_{s,v} \, \mathbb{L}_{s,v}^T
\end{equation}
Because of eq.(\ref{bisturi}) we see that under Tits Satake projection the following happens:
\begin{equation}\label{krumiricasale}
  \mathcal{M}_{s} \, \stackrel{\pi_{TS}}{\longrightarrow} \, \mathcal{M}^{TS}_{s} \, = \, \left(\begin{array}{c|c}
                                                                                                \mathfrak{M} & 0 \\
                                                                                                \hline
                                                                                                0 & \mathfrak{M}
                                                                                              \end{array}\right)
\end{equation}
where $\mathfrak{M}$ is the $2\times 2$ fundamental harmonic of the base manifold discussed in section \ref{armonia2dim}, eq.(\ref{ceffus}).
\subsubsection{Comparison with the solvable coset representative of \texorpdfstring{$\mathrm{Sp(4,\mathbb{R})/U(2)}$}{Sp(4,R)/U(2)}}
Finally, we make a comparison of the solvable group coset representative for the non-compact rank $r=2$  space
$\mathrm{Sp(4,\mathbb{R})/U(2)}$, as given in eq.(7.57) of \cite{pgtstheory} and the spinor coset representative of
$\mathrm{Spin(1,3)/SU(2)}$ displayed here in eq.(\ref{spinLcosrep}). For reader's convenience we recall here eq.(7.57) of  \cite{pgtstheory},
renaming $w_i =v_i$ ($i=1,\dots,6$):
\begin{equation}\label{romanzo}
\mathbb{L}^{\mathrm{Sp(4,\mathbb{R})}}(\mathbf{v})
 = \, \left(
\begin{array}{cccc}
 e^{\frac{1}{2} \left(-v_1-v_2\right)} & \frac{1}{2}
   e^{\frac{1}{2} \left(v_2-v_1\right)} v_6 & \frac{1}{4}
   e^{\frac{1}{2} \left(v_1+v_2\right)} \left(v_5 v_6-2 \sqrt{2}
   v_4\right) & \frac{1}{4} e^{\frac{1}{2} \left(v_1-v_2\right)}
   \left(2 v_5+\sqrt{2} v_3 v_6\right) \\
 0 & e^{\frac{1}{2} \left(v_2-v_1\right)} & \frac{1}{2}
   e^{\frac{1}{2} \left(v_1+v_2\right)} v_5 & \frac{e^{\frac{1}{2}
   \left(v_1-v_2\right)} v_3}{\sqrt{2}} \\
 0 & 0 & e^{\frac{1}{2} \left(v_1+v_2\right)} & 0 \\
 0 & 0 & -\frac{1}{2} e^{\frac{1}{2} \left(v_1+v_2\right)} v_6 &
   e^{\frac{1}{2} \left(v_1-v_2\right)} \\
\end{array}
\right)
\end{equation}
In order to compare one has to go to the same basis, so we calculate:
\begin{equation}\label{polpettone}
  \mathbb{L}^{\mathrm{Spin(1,3)}}(\mathbf{w}) \, \equiv \, \mathcal{Q}\, \mathbb{L}_s \, \mathcal{Q}^T \, = \, \left(
\begin{array}{cccc}
 e^{\frac{w_1}{2}} & -\frac{1}{2} e^{\frac{w_1}{2}} w_3 &
   \frac{1}{2} e^{\frac{w_1}{2}} w_2 & 0 \\
 0 & e^{-\frac{w_1}{2}} & 0 & 0 \\
 0 & 0 & e^{-\frac{w_1}{2}} & 0 \\
 0 & \frac{1}{2} e^{\frac{w_1}{2}} w_2 & \frac{1}{2}
   e^{\frac{w_1}{2}} w_3 & e^{\frac{w_1}{2}} \\
\end{array}
\right)
\end{equation}
We would like to see whether we can choose the $6$ parameters $v_i$ in equation (\ref{romanzo}) in such a way that
$\mathbb{L}^{\mathrm{Sp(4,\mathbb{R})}}(\mathbf{v})$ becomes $\mathbb{L}^{\mathrm{Spin(1,3)}}(\mathbf{w})$, as displayed in equation
(\ref{polpettone}). The reader can easily verify that this impossible. Both the $6$-parameter solvable group $\mathcal{S}_{\mathrm{Sp(4)}}$
defined by eq.(\ref{romanzo})  and the $3$-parameter solvable group $\mathcal{S}_{\mathrm{Spin(1,3)}}$ defined by eq.(\ref{polpettone})
are subgroups of $\mathrm{Sp(4,\mathbb{R})}$, yet $\mathcal{S}_{\mathrm{Spin(1,3)}}$ is not a subgroup of $\mathcal{S}_{\mathrm{Sp(4)}}$.
The two solvable subgroups intersect only on a common $2$-dimensional subgroup isomorphic to the solvable group of $\mathrm{SL(2,\mathbb{R})}\subset
\mathrm{Sp(4,\mathbb{R})}$. This is the reason why we have changed gamma matrix basis. There is not a smooth embedding of the solvable group of
a rank $r=1$ symmetric space $\mathrm{U/H}$ in the solvable group of a rank $r=2$ symmetric space $\mathrm{\hat{U}/\hat{H}}$ even if the former
simple group is a subgroup of the latter,
$\mathrm{U }\subset \hat{\mathrm{U}}$.
\subsection{Generalizing the construction to \texorpdfstring{${\rm Spin}(1,1+q)$}{Spin(1,1+q)}}
Consider the group ${\rm Spin}(1,1+q)$ and construct the Clifford algebra as follows:
\begin{equation}
\Gamma^0=i\,\sigma^2\otimes {\bf 1}\,,\,\,\Gamma^1=\sigma^1\otimes {\bf 1}\,,\,\,\Gamma^{1+i}=\sigma^3\otimes {\gamma}^i\,,\,\,i=1,\dots q\,,
\end{equation}
where ${\gamma}^i$ are a real representation of the ${\rm Spin}(q)$ Clifford algebra satisfying the property $(\gamma^i)^T=\gamma^i$ and are $d\times d$ matrices, where $d$ is the smallest dimension of the spinorial representation for which the above conditions on the $\gamma^i$ matrices can be satisfied.
The $\Gamma^a$ satisfy:
\begin{equation}
\{\Gamma^a,\,\Gamma^b\}=2\eta^{ab}\,{\bf 1}\,\,,\,\,\,\,\eta^{ab}={\rm diag}(-1,1,\dots,1)\,.
\end{equation}
Define the Cartan generator and the shift generators as follows:
\begin{equation}
H=\frac{1}{2}\Gamma^0\Gamma^1=\sigma_3\otimes {\bf 1}\,,\,\,E_i=\frac{1}{2}(\Gamma^0+\Gamma^1)\Gamma^{i+1}=-\sigma^+\otimes \gamma^i\,,
\end{equation}
where $\sigma_+\equiv (\sigma^1+i\,\sigma^2)/2$.
They satisfy:
$$[H,\,E_i]=E_i\,\,,\,\,\,[E_i,\,E_i^T]=2\,H\,.$$
Now define the coset representative:
\begin{equation}
\mathbb{L}=\exp(\xi H)\cdot \exp(\zeta^i\,E_i)=\left(\begin{matrix}e^{\frac{\xi}{2}}\,{\bf 1} & {\bf 0}\cr {\bf 0} & e^{-\frac{\xi}{2}}\,{\bf 1}\end{matrix}\right)\cdot \left(\begin{matrix}{\bf 1} & -\zeta^i\,\gamma_i\cr {\bf 0} & {\bf 1}\end{matrix}\right)\,,
\end{equation}
where, with reference to the previous sections, we can identify $\xi=w_1,\,\zeta^i=w_{i+1}$.
The matrix $\mathcal{M}$ reads
\begin{equation}
\mathcal{M}=\mathbb{L}\mathbb{L}^T=\left(\begin{matrix}e^\xi\,(1+|\zeta|^2)\,{\bf 1} & -\zeta^i\,\gamma_i\cr -\zeta^i\,\gamma_i & e^{-\xi}\,{\bf 1} \end{matrix}\right)\,.\label{genform}
\end{equation}
The above form follows from the property:
$$(\zeta^i\,\gamma_i)\cdot (\zeta^i\,\gamma_i)^T=|\zeta|^2\,{\bf 1}\,.$$
We see that the single scalar fields $\xi=w_1$ and $\zeta^i=w_{i+1}$ can be obtained through projections of the matrix $\mathcal{M}$:
\begin{equation}
e^{-\xi}=e^{-w_1}=\frac{1}{2d}\,{\rm Tr}\left[(({\bf 1}_2-\sigma^3)\otimes {\bf 1})\cdot \mathcal{M}\right]\,\,,\,\,\,\zeta_i=w_{i+1}=-\frac{1}{d}\,{\rm Tr}\left[(\sigma_+\otimes \gamma^i)\cdot \mathcal{M}\right]\,.
\end{equation}
\paragraph{$q=2$ case.}
The Clifford algebra is generated by:
\begin{equation}
\Gamma^0=i\,\sigma^2\otimes {\bf 1}_2\,,\,\,\Gamma^1=\sigma^1\otimes {\bf 1}_2\,,\,\,\Gamma^{2}=\sigma^3\otimes \sigma^1\,\,,\,\,\Gamma^{2}=\sigma^3\otimes \sigma^3\,,
\end{equation}
where we have chosen $\gamma^i=\{\sigma^1,\,\sigma^3\}$ and $d=2$.
The matrix $\mathcal{M}$ reads:
\begin{equation}
\mathcal{M}=\left(
\begin{array}{cccc}
 \left(\zeta _1^2+\zeta _2^2+1\right) e^{\xi } & 0 & -\zeta _2 & -\zeta _1 \\
 0 & \left(\zeta _1^2+\zeta _2^2+1\right) e^{\xi } & -\zeta _1 & \zeta _2 \\
 -\zeta _2 & -\zeta _1 & e^{-\xi } & 0 \\
 -\zeta _1 & \zeta _2 & 0 & e^{-\xi } \\
\end{array}
\right)\,.
\end{equation}
\paragraph{$q=3$ case.}
The Clifford algebra is generated by:
\begin{equation}
\Gamma^0=i\,\sigma^2\otimes {\bf 1}_2\otimes {\bf 1}_2\,,\,\,\Gamma^1=\sigma^1\otimes {\bf 1}_2\otimes {\bf 1}_2\,,\,\,\Gamma^{2}=\sigma^3\otimes \sigma^1\otimes {\bf 1}_2\,,\,\,\Gamma^{2}=\sigma^3\otimes \sigma^3\otimes {\bf 1}_2\,,\,\,\Gamma^{3}=\sigma^3\otimes \sigma^2\otimes \sigma^2\,,
\end{equation}
where we have chosen $\gamma^i=\{\sigma^1\otimes {\bf 1}_2,\,\sigma^3\otimes {\bf 1}_2,\sigma^2\otimes \sigma^2\}$  and $d=4$.
The matrix $\mathcal{M}$ has the form \eqref{genform} with
\begin{equation}
-\zeta^i\,\gamma_i=\left(
\begin{array}{cccc}
 -\zeta _2 & 0 & -\zeta _1 & \zeta _3 \\
 0 & -\zeta _2 & -\zeta _3 & -\zeta _1 \\
 -\zeta _1 & -\zeta _3 & \zeta _2 & 0 \\
 \zeta _3 & -\zeta _1 & 0 & \zeta _2 \\
\end{array}
\right)\,.
\end{equation}
We end noting that we do not need $\gamma^i$ to be real, but only that $\gamma^i=(\gamma^i)^\dagger$. This reduces the value of $d$ for various $q$ to $d=2^{\left[\frac{q}{2}\right]}$. If $\gamma^i$ are complex, the matrix $\mathcal{M}$, now hermitian, has to be defined as follows:
\begin{equation}
\mathcal{M}=\mathbb{L}\mathbb{L}^\dagger=\left(\begin{matrix}e^\xi\,(1+|\zeta|^2)\,{\bf 1} & -\zeta^i\,\gamma_i\cr -\zeta^i\,\gamma_i & e^{-\xi}\,{\bf 1} \end{matrix}\right)\,.
\end{equation}
With this prescription, for instance, in the $q=3$ case we could have chosen $\gamma^i=\sigma^i$.

\subsection{Harmonic Expansion on the 2-dimensional base manifold}
\label{armonia2dim}
The metric on the Hyperbolic Plane is written in the following way in terms of the solvable coordinates (\(w_1\),\(w_2\)):
\begin{equation}\label{metrapallosa}
  \text{ds}^2 ={dw}_1{}^2+\frac{1}{4} \left(w_2 {dw}_1+{dw}_2\right){}^2
\end{equation}
Instead the Cartesian coordinates in the Poincar{\' e} disk model of \(\mathbb{H}^2\) are related to the solvable coordinates as follows:
\begin{equation}\label{dapallasolvay}
  \{x,y\}=\left\{\frac{-4+e^{2 w_1} \left(4+w_2^2\right)}{4+8 e^{w_1}+e^{2 w_1} \left(4+w_2^2\right)},-\frac{4 e^{w_1} w_2}{4+8 e^{w_1}+e^{2 w_1}
\left(4+w_2^2\right)}\right\}
\end{equation}
In the metric (\ref{metrapallosa}), the Laplace Beltrami operator on \(\mathbb{H}^2\) takes the following form;
\begin{equation}\label{laplacbas}
  \pmb{\triangle}_s \mathit{f} =2 \left[-w_2 \,\frac{\partial \mathit{f}}{\partial{w_2}}+
  \left(1+\frac{w_2^2}{4}\right) \, \frac{\partial^2 \mathit{f}}{  \partial w_2^2}\, -\,\frac{\partial \mathit{f}}{\partial{w_1}}\,
  +\, w_2 \, \frac{\partial^2 \mathit{f}}{  \partial w_1 \partial w_2}\, + \, \frac{\partial^2 \mathit{f}}{  \partial w_1^2}\right]
\end{equation}
Recalling section 10 of the basic PGTS theory paper \cite{pgtstheory} and in particular eq.s (10.36-10.40) we know that the eigenvalues of
the invariant quadratic differential operator (\ref{laplacbas}) correspond to the values of the Casimir operator on the irreducible representations
of the group $\mathrm{SL(2,\mathbb{R})}\sim \mathrm{SO(1,2)}$ (where $\sim$ denotes the local isomorphism) and that the eigenfunctions pertaining
to each eigenvalue can be constructed in purely group theoretical terms by means of the tensor product of the fundamental harmonic
$\mathcal{M}_{AB}$ obtained from the solvable coset representative $\mathbb{L}$.
\par
In the present case we have:
\begin{equation}\label{quran}
 \mathbb{L} \, = \,  \left(
\begin{array}{cc}
 e^{\frac{w_1}{2}} & \frac{1}{2} e^{\frac{w_1}{2}}
   w_2 \\
 0 & e^{-\frac{w_1}{2}} \\
\end{array}
\right)
\end{equation}
and recalling eq.(10.41) of \cite{pgtstheory}, that introduces the convenient Young Tableau notation for harmomonics we can write:
\begin{equation}\label{ceffus}
  \young(AB)\, \equiv \, \mathbb{L}^A_i \mathbb{L}^B_j \, \delta^{ij} \, = \, \left(\mathbb{L} \, \mathbb{L}^T\right)^{AB} \, \equiv \,
  \mathfrak{M}^{AB} \, = \,  \left(
\begin{array}{cc}
 \frac{1}{4} e^{w_1} \left(w_2^2+4\right) &
   \frac{w_2}{2} \\
 \frac{w_2}{2} & e^{-w_1} \\
\end{array}
\right)
\end{equation}
In our normalizations the eigenvalue of the fundamental harmonic $\young(AB)$, which has three components, namely $\young(11)$,
$\young(12)$, $\young(22)$, is $4$:
\begin{equation}\label{laperio}
  \pmb{\triangle}_s \, \young(AB)\, = \, 4 \, \young(AB)
\end{equation}
We refer once again to \cite{pgtstheory}, section 10.6,  and to the relation (for maximally split algebras of type $\slal (N,\mathbb{R})$)
between Dynkin labels $n_i$ and Young labels $\lambda_i$ which is reported in eq.(10.46) of \cite{pgtstheory}, namely:
\begin{flalign}\label{chiccobello}
  n_1 &= \lambda_1 -\lambda_2 \nonumber\\
  n_2 &= \lambda_2 -\lambda_3 \nonumber\\
  \dots & = \dots \nonumber\\
  n_r &=\lambda_{r-1} -\lambda_r
\end{flalign}
where $\pmb{\lambda} $  (the Young labels) denote the partitions of an integer number $1 \leq n\in \mathbb{N}$ \textit{i.e.}
the $r$-tuples $\pmb{\lambda} $ of integer numbers (with $r\leq n$)
such that:
\begin{equation}\label{corifeo}
\pmb{\lambda} \, = \, \{\lambda_1, \lambda_2, \dots,\lambda_r\} \quad ; \quad \lambda_1 \geq \lambda_2 \geq\dots \geq \lambda_{r-1}
\geq \lambda_r \quad ; \quad \sum_{i=1}^r \, \lambda_i \, = \, n
\end{equation}
In the case of the  $\slal (2,\mathbb{R})$) algebra, $r=1$ and there is only one type of Young tableaux contributing
to the Laplacian spectrum, namely the bar with $2 J$ boxes:
\begin{equation}\label{barra}
  \underbrace{\young(\null\null\cdot\cdot \null\null)}_{\text{$2J$ boxes}}
\end{equation}
and we have:
\begin{equation}\label{coriandolo}
  \pmb{\triangle}_s \, \young(\null\null\cdot\cdot \null\null) \, = \, 2 (J + 1) \, J\, \young(\null\null\cdot\cdot \null\null)
   \quad ; \quad J\in \mathbb{N}
\end{equation}
which is what is predicted by eq.(10.40) of \cite{pgtstheory}, namely by the Weyl-Freudenthal formula in the case of rank $r=1$ that is a Lie algebra
with just one root.
\paragraph{\sc The $\mathrm{SL(2,\mathbb{R})}$ harmonic expansion.} It follows that any smooth function:
\begin{equation}\label{funziacosetta}
  \mathit{f} \, : \, \frac{\mathrm{SL(2,\mathbb{R})}}{\mathrm{SO(2)}} \, \longrightarrow \, \mathbb{R}
\end{equation}
on the Poincar\'e manifold can be expanded in a multiple power series of three fundamental objects as it follows:
\begin{equation}\label{harexpa2}
  \mathit{f}(p) \,=\, \sum_{i=0}^{\infty}\sum_{j=0}^{\infty}\sum_{k=0}^{\infty} \, c_{ijk} \, \mathit{m}_{11}(p)^i \,\mathit{m}_{12}(p)^j
  \mathit{m}_{22}(p)^k   \quad ; \quad p\in \frac{\mathrm{SL(2,\mathbb{R})}}{\mathrm{SO(2)}}
\end{equation}
In the above eq.(\ref{harexpa2}) the coefficients $c_{ijk}$ are constant real numbers and $p$ denotes a point of the manifold without the specific
mention of any coordinate system. The objects $\mathit{m}_{AB}(p)$ are the matrix elements of the fundamental harmonic $\mathcal{M}^{AB}$ mentioned
in eq.(\ref{ceffus}) and have an intrinsic meaning, independently from the used coordinate system, so as the coefficients $c_{ijk}$ do. In whatever
coordinate system one utilizes to label the points $p$, the harmonic expansion of a function $\mathit{f}(p)$ remains the same. In other words the
coefficients $c_{ijk}$ are an intrinsic coordinate-free definition of the smooth function $\mathit{f}(p)$.
The three objects $\mathit{m}_{AB}(p)$ are eigenstates of the Laplacian  $\pmb{\triangle}_s$ with eigenvalue $4$, yet a generic monomial
\begin{equation}\label{cromatopuzza}
  \mu^{ijk} \, \equiv \, \mathit{m}_{11}(p)^i \,\mathit{m}_{12}(p)^j
  \mathit{m}_{22}(p)^k
\end{equation}
is not an eigenstate of $\pmb{\triangle}_s$.
However for any fixed degree:
\begin{equation}\label{degrado}
  \mathit{d} \, \equiv \, i+j+k
\end{equation}
we can reorganize the vector space of the monomials of such degree into subspaces that are eigenspaces of the Laplacian and constitute
irreducible representations of the $\slal(2,\mathbb{R})$ algebra of dimension $2J+1$ with Casimir operator (and then Laplacian) eigenvalue
 $2J(J+1)$. At degree $\mathit{d}$
the maximum value of $J$ is $J=\mathit{d}$ and the corresponding $2J+1$ eigenfunctions are those mentioned in eq.(\ref{coriandolo}).
The representations with a value $J < d$ appear because the cofactors of the eigenfunctions of lesser degree that complete the considered degree
are polyonomials in the matrix entries exactly equal to the determinant of $\mathcal{M}^{AB}$. The latter is equal to $1$ and hence the actual
degree of the harmonic is $J$. In other words when we sum freely on the monomials as we do in eq.(\ref{harexpa2}) we overcount the actual harmonics
since the same harmonics are repeated over and over at higher degree. The correct harmonic expansion without overcounting is the following one:
\begin{equation}\label{harmexp2giusta}
  \mathit{f}(p) \,=\, \sum_{J=0}^{\infty}\sum_{\ell=-J}^{J} c_{J\ell} \,
  \pmb{\mathfrak{harm}}^{J}_\ell(p) \,
\end{equation}
where
\begin{equation}\label{armoniavivaldi}
  \pmb{\mathfrak{harm}}^{J}_\ell(p) \,\equiv \,  \mathrm{Polynomial}^{J}_\ell\left( \mathit{m}_{11}(p),\mathit{m}_{12}(p),\mathit{m}_{12}(p)\right)
\end{equation}
are those polynomials of degree $J$ in the three matrix entries that diagonalize the Casimir operator and hence the Laplacian and pertain to its
eigenvalue $2 J(J+1)$.
\par
The relevant point in the above discussion is that the free sum of eq.(\ref{harexpa2}) captures all the terms in the expansion of
eq.(\ref{armoniavivaldi}), missing no-one, rather capturing each several times. In a neural network algorithm this is not a problem: it simply means
that the unique characteristic coefficients $c_{J\ell}$ are the sum of several $c_{ijk}$, yet the function defined by the expansion
eq.(\ref{armoniavivaldi}) and by the expansion (\ref{harexpa2}) are the same upon such an identification.
\subsubsection{Euler coordinates versus solvable coordinates}
The harmonic functions $\pmb{\mathfrak{harm}}^{J}_\ell(p)$ constitute an orthogonal system of functions in the functional space
$C^{\infty}\left(\frac{\mathrm{SL(2,\mathbb{R})}}{\mathrm{SO(2)}} \right)$. Orthogonal with respect to which scalar product? Since the manifold is not
compact we cannot find a \textit{Haar measure} for integrals of harmonics on the base-space, yet there is a convenient alternative. It suffices
to use synergically with the solvable one also the \textbf{Euler parameterization} illustrated in depth in section 6.3 of \cite{pgtstheory}.
In our case which is the smallest possible one, by means of the rigorous procedure discussed in the quoted reference and already implemented in
various neural network algorithms in the twin papers \cite{TSnaviga,naviga}, the Euler parameterization trades the solvable coordinates $\{w_1,w_2\}$,
that span the entire $\mathbb{R}^2$, for a pair $\mu_1,\theta_1$ where the first is the unique Cartan coordinate spanning the real line $\mathbb{R}$,
while $\theta_1$ is an angle spanning the circle $\mathrm{SO(2)} \sim \mathbb{S}^1$. The essential thing is that the matrix $\mathcal{M}^{AB}$, as
given in equation (\ref{ceffus}), being by definition symmetric, can be diagonalized with an $\mathrm{SO(2)}$ rotation of a suitable angle.
So doing we get
\begin{eqnarray}\label{eulerone}
  \mathfrak{M}^{AB} & = & \left(
\begin{array}{cc}
 e^{\mu _1} \sin ^2 \theta _1  +e^{-\mu _1} \cos ^2 \theta _1
   & e^{\mu _1} \sin  \theta _1  \cos  \theta _1  -e^{-\mu _1}
   \sin  \theta _1   \cos  \theta _1   \\
 e^{\mu _1} \sin  \theta _1   \cos  \theta _1  -e^{-\mu _1}
   \sin  \theta _1   \cos  \theta _1   & e^{-\mu _1} \sin
   ^2 \theta _1  +e^{\mu _1} \cos ^2 \theta _1   \\
\end{array}
\right) \nonumber \\ & = & \left(
\begin{array}{cc}
 \frac{1}{4} e^{w_1} \left(w_2^2+4\right) &
   \frac{w_2}{2} \\
 \frac{w_2}{2} & e^{-w_1} \\
\end{array}
\right)
\end{eqnarray}
which, on one side allows to work out the transformation from the solvable to the Euler coordinates, on the other it already provides the expression of
fundamental harmonic building blocks $\mathit{m}_{AB}$ in terms of the Euler coordinates.
Using the latter, the geometry of the manifold can be described by means of the following zweibein:
\begin{eqnarray}\label{zweibein}
\mathbf{e}^1 & = &-\sqrt{2} \,\text{d$\mu $}_1 \nonumber\\
\mathbf{e}^2 & = & -2 \,\sqrt{2} \, \sinh (\text{$\mu_1 $}) \,\text{d$\theta $}_1 \nonumber\\
ds^2 & = & \mathbf{e}^1 \times \mathbf{e}^1 +  \mathbf{e}^2 \times \mathbf{e}^2 \, = \,
2\, \text{d$\mu $}_1^2+8 \, \sinh ^2\left(\mu _1\right)\,\text{d$\theta $}_1^2
\end{eqnarray}
In these coordinates we can calculate the volume form given by
\begin{equation}\label{voglioforma}
  Vol \, = \, \mathbf{e}^1 \, \wedge \, \mathbf{e}^2 \, = \, - \, 4 \, \sinh\left(\mu _1\right) \, \text{d$\mu $}_1 \, \wedge \text{d$\theta $}_1
\end{equation}
Given two functions $\Phi_A(\mu_1,\theta_1),\Phi_B(\mu_1,\theta_1)$ on $\frac{\mathrm{SL(2,\mathbb{R})}}{\mathrm{SO(2)}}$ written in the Euler
parameterization, their integral scalar product should be:
\begin{equation}\label{cremisi}
  \langle \Phi_A \, , \, \Phi_B \rangle  \, \equiv \, \int\int \left(- \, 4 \, \sinh\left(\mu _1\right) \,
  \Phi_A(\mu_1,\theta_1)\,\Phi_B(\mu_1,\theta_1)\right)
  d\mu_1 \,  d\theta_1
\end{equation}
but the integral on $\mu_1$ is not convergent. This problem is universally solved with a Wick rotation trick which works very well
in the present case and extends equally well to all dimensions and to all larger groups. Following the notations and conceptions
of \cite{pgtstheory}, the angular integral is generically over the Grassmaniann manifold $\mathcal{F}_{TS}$ defined in eq.(6.29) of
\cite{pgtstheory}, whose general scope is illustrated in section 6.3.1 of the same paper.
The overall geometry of the hyperbolic symmetric space, once written in Euler parameterization, already defines the correct
\textit{Haar measure} over the compact manifold $\mathcal{F}_{TS}$ and with respect to that integral, which is convergent, the
harmonics are already orthogonal to each other so that the integral over the non-compact Cartan directions $\mu_i$ is irrelevant.
It remains the problem of performing the integral in $\mu_i$ when we come to investigate the norm of one harmonic, namely
$\pmb{\mathfrak{harm}}^{J}_\ell(p)$ convolved with itself. It is here that the Wick rotation trick comes to rescue. We take
the definition of the scalar product in eq.(\ref{cremisi}) and we modify it as follows:
\begin{equation}\label{carriolavichiana}
\langle \Phi_A \, , \, \Phi_B \rangle  \, \equiv \, - \, 4 \, \int_{0}^\pi d\psi_1\int_{0}^{2\pi} d\theta_1 \sin\left(\psi _1\right) \,
  \Phi_A(-\mathit{i}\psi_1,\theta_1)\,\Phi_B(\mathit{i}\psi_1,\theta_1)
\end{equation}
 Now the integral is finite and allows to define the norm of each harmonic so that, rescaling each one by the inverse square root of its norm,
 we can erect the entire set of harmonics to a complete orthonormal basis of functions on the manifold. The coefficients of the harmonic expansion of
 any given function $\mathit{f}$ can be calculated by performing its harmonic integral (\ref{carriolavichiana}) with each element of the
 functional basis.
 \subsubsection{No regular harmonic  invariant against the action of the Fuchsian subgroup \texorpdfstring{$\Gamma$}{G}}
\label{cirimello}
As we argue later on in section \ref{pitture}, it makes sense to map the finite square grid of points representing a
typical image to the fundamental domain $\mathcal{F}_in\Gamma$ of a Fuchsian group, namely to a Riemann surface.
Visually, this is shown later on in fig.\ref{bolzanetoquadro}. Given this fact, it makes sense to consider the harmonic
expansion of functions that are requested to be invariant under the Fuchsian subgroup $\Gamma$.
This is the perfect analogue of what one systematically does in Euclidean geometry when the problem is that of considering functions over a compact portion of a hyperplane delimited by faces that make it into a box. In this case, one introduces a basis of periodic functions (Fourier series), namely, invariant under a discrete subgroup of the Euclidean translation group. In hyperbolic geometry, one might have parabolic translations; however, the most frequent case is that where the group that delimits the box is made of hyperbolic elements, namely, it is Fuchsian.
\par
For this reason, we would like to consider the space of the regular harmonics, each corresponding to a finite-dimensional
representation of the isometry group $\mathrm{ISO}_{H^2} \,=\, \mathrm{SL(2,\mathbb{R})}$, that are invariant with respect to the Fuchsian
group $\Gamma$. We show in this subsection that no regular harmonics exist having such a property. This fact is related to the general property of non-compact groups whose finite-dimensional irreducible representations are in one-to-one correspondence
with the finite-dimensional irreducible representations of the corresponding compact group\footnote{The corresponding compact group is one whose Lie algebra is the unique compact real section of the complexification of the Lie algebra of $U$.
In the present case the compact group corresponding to $\mathrm{SL(2,\mathbb{R})}$ is $\mathrm{SU(2)}$}. The unitary irreducible representations of compact groups are infinite-dimensional and are provided by suitable functional spaces. The very fact that regular harmonics invariant under the action of the Fuchsian group do not exist is what makes it necessary
to change the approach and consider other bases of $\Gamma$-invariant functions that are also eigenfunctions of the Laplacian and whose support is the fundamental domain $\mathcal{F}_\Gamma$. We show below how the non-existence of $\Gamma$-invariant regular harmonics emerges, focusing on the case of $\Gamma_{16}$. A completely analogous proof for $\Gamma_{8}$ and any other Fuchsian group might be obtained, but it is skipped as pleonastic.
\paragraph{\sc Transformation of the three fundamental harmonics under the Fuchsian group $\Gamma_{16}$.}
The discrete group $\Delta^+_{8,3,2} \subset \mathrm{SL(2,\mathbb{R})}$, defined by the presentation in eq.(\ref{equadelta1}),
whose relations are explicitly satisfied by the generators shown in eq.s (\ref{equadeltagenTS}), is a subgroup of the isometry group of
the Hyperbolic Plane. Since the three fundamental harmonics $\mathit{m}_{11},\mathit{m}_{12},\mathit{m}_{22}$
constitute a basis for the irreducible three dimensional representation of $\mathrm{SL(2,\mathbb{R})}\sim \mathrm{SO(1,2)}$,
it follows that they must be linearly mapped one into the other by the action  of   $\Delta^+_{8,3,2}$
and hence by the whole group. It is our interest to find such a matrix representation of the tessellation group.
To this effect, we first normalise the three fundamental harmonics to have an orthonormal basis of functions.
Using the Wick rotation trick, we found that the orthonormal basis is given by:
\begin{equation}\label{unibas}
  \boldsymbol{\mathcal{U}}(p) = \left\{\frac{1}{4} \sqrt{\frac{3}{2 \pi }} \mathit{m}_{22}(p)\, ,\, \frac{1}{4}
   \sqrt{\frac{3}{\pi }} \mathit{m}_{12}(p)\, ,\, \frac{1}{4} \sqrt{\frac{3}{2 \pi }}
   \mathit{m}_{11}(p)\right\}
\end{equation}
which in the solvable coordinates means the following:
\begin{equation}\label{unibas2}
  \boldsymbol{\mathcal{U}}(p) = \left\{\frac{1}{4} \sqrt{\frac{3}{2 \pi }} e^{-w_1}\, ,\, \frac{1}{8} \sqrt{\frac{3}{\pi
   }} w_2\, ,\, \frac{1}{4} \sqrt{\frac{3}{2 \pi }} e^{w_1}
   \left(\frac{w_2^2}{4}+1\right)\right\}
\end{equation}
With obvious substitutions, one also obtains the expression of the same functions in the Euler coordinate basis.
With the harmonic integrals defined in equation (\ref{carriolavichiana}) we are able calculate for every element $\gamma \in\Delta^+_{8,3,2}$
and in particular for the generators (\ref{equadeltagenTS}) the corresponding $3\times 3$ matrix $D_{ij}(\gamma)$ such that the following is true:
\begin{equation}\label{Darstellung}
  \boldsymbol{\mathcal{U}}_i(\gamma\cdot p) \, = \, D_{ij}(\gamma)\,\boldsymbol{\mathcal{U}}_j(p)
\end{equation}
Explicitly, we have:
\begin{alignat}{3}\label{Darstgen}
  D(\mathfrak{T}) & = \left(
\begin{array}{ccc}
 \frac{1}{4} \left(2+\sqrt{2}\right) & \frac{1}{2} & \frac{1}{4}
   \left(2-\sqrt{2}\right) \\
 -\frac{1}{2} & \frac{1}{\sqrt{2}} & \frac{1}{2} \\
 \frac{1}{4} \left(2-\sqrt{2}\right) & -\frac{1}{2} & \frac{1}{4}
   \left(2+\sqrt{2}\right) \\
\end{array}
\right)\nonumber\\
D(\mathfrak{S}) & = \left(
\begin{array}{ccc}
 \frac{1}{4}+\frac{1}{\sqrt{2}}-\frac{1}{\sqrt[4]{2}} & \frac{1}{4}
   \left(2+\sqrt{2}-2 \sqrt{4+3 \sqrt{2}}\right) & \frac{3}{4}+\frac{1}{\sqrt{2}}
   \\
 -\frac{1}{2}+\frac{1}{2^{3/4}}-\frac{1}{2 \sqrt{2}}+\frac{1}{\sqrt[4]{2}} &
   -\frac{1}{2}-\sqrt{2} & \frac{1}{4} \left(2+2 \sqrt[4]{2}+\sqrt{2}+2\
   2^{3/4}\right) \\
 \frac{1}{4} \sqrt{17+12 \sqrt{2}} & \frac{1}{4} \left(-2-\sqrt{2}-2 \sqrt{4+3
   \sqrt{2}}\right) & \frac{1}{4}+\frac{1}{\sqrt{2}}+\frac{1}{\sqrt[4]{2}} \\
\end{array}
\right) \nonumber\\
D(\mathfrak{R}) & = \left(
\begin{array}{ccc}
 \frac{1}{4} \left(1+\sqrt{2}\right) & \frac{1}{2}
   \left(-\frac{1}{\sqrt{2}}-\sqrt{4+3 \sqrt{2}}\right) & \frac{1}{4} \left(3+3
   \sqrt{2}+2\ 2^{3/4}\right) \\
 \frac{1}{2^{3/4}}-\frac{1}{2 \sqrt{2}}+\frac{1}{\sqrt[4]{2}} &
   -\frac{3}{2}-\frac{1}{\sqrt{2}} & \frac{1}{2^{3/4}}+\frac{1}{2
   \sqrt{2}}+\frac{1}{\sqrt[4]{2}} \\
 \frac{1}{4} \left(3+3 \sqrt{2}-2\ 2^{3/4}\right) & \frac{1}{4} \left(\sqrt{2}-2
   \sqrt{4+3 \sqrt{2}}\right) & \frac{1}{4} \left(1+\sqrt{2}\right) \\
\end{array}
\right)
\end{alignat}
and one can easily verify  that the matrices in eq(\ref{Darstgen}) satisfy in $d=3$ the same relations (\ref{equadelta1}) as their progenitors in
$d=2$.
\par
Next, either by matrix product or directly with the same harmonic integral method, we can calculate the representation on harmonics of the group element
$ \mathfrak{U} \, = \, \mathfrak{T}^2\cdot \mathfrak{S}^2$. We obtain:
\begin{equation}\label{DarstUgoth}
D(\mathfrak{U}) \, = \, \left(
\begin{array}{ccc}
 \frac{1}{4} \left(3+3 \sqrt{2}+2 \sqrt{4+3 \sqrt{2}}\right) & \frac{1}{4}
   \left(-4-\sqrt{2}-2 \sqrt{4+3 \sqrt{2}}\right) & \frac{1}{4}
   \left(1+\sqrt[4]{2}\right)^2 \\
 \frac{\sqrt[4]{2}-2}{2\ 2^{3/4}} & \frac{1}{2}+\frac{1}{\sqrt{2}} &
   -\frac{2+\sqrt[4]{2}}{2\ 2^{3/4}} \\
 \frac{1}{4} \left(\sqrt[4]{2}-1\right)^2 & \frac{1}{4} \left(4+\sqrt{2}-2
   \sqrt{4+3 \sqrt{2}}\right) & \frac{1}{4} \left(3+3 \sqrt{2}-2 \sqrt{4+3
   \sqrt{2}}\right) \\
\end{array}
\right)
\end{equation}
whose third power is the following:
\begin{equation}\label{DarstUcube}
  D(\mathfrak{U}^3) \, = \, \left(
\begin{array}{ccc}
 10+7 \sqrt{2}+2 \sqrt{48+34 \sqrt{2}} & -12-8 \sqrt{2}-\sqrt{2 \left(140+99
   \sqrt{2}\right)} & 7+5 \sqrt{2}+2 \sqrt{24+17 \sqrt{2}} \\
 -\sqrt[4]{2} \left(2+\sqrt{2}\right) & 3+2 \sqrt{2} & -\sqrt[4]{2}
   \left(2+\sqrt{2}\right) \\
 7+5 \sqrt{2}-2 \sqrt{24+17 \sqrt{2}} & 12+8 \sqrt{2}-\sqrt{2 \left(140+99
   \sqrt{2}\right)} & 10+7 \sqrt{2}-2 \sqrt{48+34 \sqrt{2}} \\
\end{array}
\right)
\end{equation}
As we know from section \ref{Gamma16normale} and eq.s (\ref{mattoneQU}--\ref{mholoni}) the normal subgroup $\Gamma_{16}$ is generated by $16$
generators obtained as the conjugation of the operator $\mathfrak{QU} \, = \,\mathfrak{U}^3$.
\par
A regular harmonic invariant with respect to any generator of the Fuchsian group does exist in the fundamental vector representation $J=1$ and in each of its tensor products. It is the eigensubspace corresponding to the eigenvalue $1$ of the matrix (\ref{DarstUcube}) or of
its conjugate with any element of the $\mathbb{Z}_8$ cyclic group generated by (\ref{Darstgen}). The eigenvalues $\pm 1$ are present for each generator in each irreducible representation obtained as the tensor product of the fundamental one, as we have explicitly checked, yet the corresponding eigenspaces have always dimension $1$. Since the eigenspaces corresponding to the eigenvalues $\pm 1$ are different for each of the
 16 generators (they are the $\mathrm{Z}_8$ images of the eigenspaces $\pm 1$ of \ref{DarstUcube}), it follows that there is never a regular harmonic invariant against all of the 16 generators at the same time. This concludes the proof of what we stated.

\section{The length spectrum and the heat kernel}
\label{straccioni}
In this section, we consider from a general viewpoint properties of the distance function for all the manifolds in the Tits Satake universality class of the Hyperbolic 2-plane $\mathbb{H}^2$, namely the case $r=1$  of the manifold families (\ref{titstot}).
\par
The aptness of the solvable coordinates parametrization becomes manifest in the simplicity of the Riemannian metric.
In the case $r=1$, for a manifold $\mathrm{SO(1,2+q)/SO(2+q)}$ the metric takes the following form
\begin{equation}\label{metrulla}
  ds^2 \, = \, \underbrace{dw_1^2 + \frac{1}{4} (w_2 dw_1 +dw_2)^2 }_{\text{metric of the TS submanifold}}+
   \underbrace{\sum_{i=1}^q\,\frac{1}{4} (w_{2+i} dw_1 +dw_{2+i})^2}_{\text{subPaint invariant addition}}
\end{equation}
The Tits Satake projection (see \cite{pgtstheory} for all the advocated concepts) simply corresponds to setting
$w_{2+i} \, = \, 0$, ($i=1,\dots, q$).
\par
Not surprisingly, the Laplacian operator, that is the unique invariant quadratic Laplace-Beltrami operator on the manifold, takes an equally very simple form in such coordinates. For the sake of  writing shortness we denote $\Delta^{(k)}_{LB}$ the Laplacian
for the case of the manifold $k$ in the considered Tits-Satake universality class. In particular, for the lowest case $k=0$, namely
for the maximally split $\mathrm{SO(1,2)/SO(2)}\sim \mathrm{SL(2,\mathbb{R})/SO(2)}$ Tits Satake submanifold
of the whole universality class,  we have:
\begin{equation}\label{Delta0LB}
  \Delta^{(0)}_{LB} \,\mathit{f}(\mathbf{w}) \, =\left( \partial^2_{w_1} +\left(4+w_2^2\right) \partial^2_{w_2} \, -
  \, 2\, \partial_{w_1} \partial_{w_1} \, + \, 2 \, w_2 \, \partial_{w_2} \, - \, \partial_{w_1} \right) \, \mathit{f}(\mathbf{w})
\end{equation}
for any function $\mathit{f}(\mathbf{w})$ of the solvable coordinates (Apart from an overall factor  the differential operator
in eq.(\ref{Delta0LB}) is the same as the differential operator in eq.(\ref{laplacbas})).
An extremely relevant point is that the determinant of the metric $\text{Det}g_{ij}(\mathbf{w})=\text{const}$ is
a constant for all manifolds of the universality class while working in solvable coordinates. Such a property
is true not only for the $r=1$ class but for all classes (see \cite{pgtstheory}).
It follows from this that the Riemannian integration measure on all manifolds of the class is simply the
integration measure on $\mathbb{R}^{2+k}$:
\begin{equation}\label{Riemeasure}
  \boldsymbol{d\mu}_{Rie}(\mathbf{w}) \equiv \prod_{i=1}^{2+k} dw_{i}
\end{equation}
This is in perfect harmony with the focal point of the whole PGTS theory, extensively discussed in \cite{pgtstheory},
namely with the fact the solvable coordinates, \textit{i.e} the parameters of the solvable
Lie group manifold $\mathcal{S}_{\mathrm{U/H}}$, metrically equivalent to the considered Riemannian non-compact symmetric
space $\mathcal{M}^{r,s}$ constitute a global chart  so that $\mathcal{M}^{r,s}$ is actually diffeomorphic to $\mathbb{R}^n$.
As we have pointed out in \cite{TSnaviga}, all non-compact symmetric spaces $\mathrm{U/H}$ are \textbf{Cartan Hadamard manifolds},
namely are \textbf{simply connected, geodesically complete and have everywhere non-negative sectional curvature}. The Cartan-Hadamard theorem
guarantees that all such manifolds are diffeomorphic to $\mathbb{R}^n$, having denoted by $n$ their dimensions.
In the present case, namely for hyperbolic spaces $\mathbb{H}^{2+k}$, we have that they are diffeomorphic to $\mathbb{R}^{2+k}$.
\par
For \textit{regular functions} defined on the whole $\mathbb{H}^{2+k}$ manifold the Laplacian spectrum:
\begin{equation}\label{lapspec}
  \Delta^{(0)}_{LB} \mathit{f}_{\Lambda} (\mathbf{w}) \, = \, \Lambda \, \mathit{f}_{\Lambda} (\mathbf{w})
\end{equation}
 is provided by the harmonics $\boldsymbol{\mathfrak{harm}}^{\Lambda}_{m}(\mathbf{w})$ defined and discussed in section \ref{spinatore},
 following the general principles outlined in \cite{pgtstheory}. In particular the values of the eigenvalues $\Lambda$ are provided by
 the values of the algebraic Casimir operator of the Lie Algebra $\so(1,2+k)$ corresponding to those irreducible representations that
 are admitted in the spectrum, namely that contain the singlet representation when reduced to the compact subalgebra $\so(2+k)$.
\par
The harmonics $\boldsymbol{\mathfrak{harm}}^{\Lambda}_{m}(\mathbf{w})$ are not square integrable functions, and just because of that, to define their orthogonality properties, we had to resort to the Wick rotation trick of eq.(\ref{carriolavichiana}).
On the other hand, any function on $\mathbb{H}^{2+k}$, also square integrable ones, namely $\Phi(\mathbf{w})$ such that:
\begin{equation}\label{normsqinte}
  \int_{\mathbb{H}^{2+k}} \mid \Phi(\mathbf{w}) \mid^2 \boldsymbol{d\mu}_{Rie}(\mathbf{w})  \, < \, \infty
\end{equation}
can be developed in harmonic series:
\begin{equation}\label{harmosvilup}
  \Phi(\mathbf{w})\, = \, \sum_{\Lambda=\Lambda_0}^{\infty}\sum_{m=1}^{\text{dim}D[\Lambda]} \,c_{\Lambda}^m
  \boldsymbol{\mathfrak{harm}}^{\Lambda}_{m}(\mathbf{w})
\end{equation}
where $D[\Lambda]$ denotes the irreducible representation of $\so(1,2+k)$ corresponding to the eigenvalue $\Lambda$,
$\Lambda_0$ identifies the smallest one in the admissible spectrum, and the index $m$ enumerates an orthogonal vector basis
in each $D[\Lambda]$.
In particular one can consider compact subregions of $\mathbb{H}^{2+k}$  identified as the fundamental domain of a generalized Fuchsian
subgroup $\Gamma$:
\begin{equation}\label{gromiko}
  \mathcal{F}_\Gamma \, \subset \, \mathbb{H}^{2+k} \quad ; \quad \mathcal{F}_\Gamma\ \, \equiv \, \mathbb{H}^{2+k} /
  \Gamma
\end{equation}
By generalized Fuchsian subgroup  we mean a \textbf{properly discontinuous subgroup} $\Gamma \subset \mathrm{\Spin(1,2+k)}$
such that all its elements are hyperbolic, namely, each of them should admit at least two fixed points on
$\mathbb{H}^{2+k}$. In the case of such compact regions as defined in eq.(\ref{gromiko}), one is interested in the space of
square integrable functions over them:
\begin{eqnarray}\label{canis}
  \Phi \quad  & \text{:}  &\mathcal{F}_\Gamma \, \longrightarrow \, \mathbb{R} \nonumber\\
 & \null  &\int_{\mathcal{F}_\Gamma} \mid \Phi(\mathbf{w}) \mid^2 \ \pi_{\mathcal{F}_\Gamma}^\star(\boldsymbol{d\mu}_{Rie}(\mathbf{w})) \, <\infty
\end{eqnarray}
where $\pi_{\mathcal{F}_\Gamma}^\star$ denotes the pull-back of the restriction map to the fundamental domain :
\begin{equation}\label{cartolinanera}
  \forall \mathbf{w} \in \mathbb{H}^{2+k} \quad : \quad \pi_{\mathcal{F}_\Gamma}^\star(\boldsymbol{d\mu}_{Rie}(\mathbf{w}))
  \, = \, \left\{ \begin{array}{cc}
  \boldsymbol{d\mu}_{Rie}(\mathbf{w}) & \text{if $\mathbf{w}\in \mathcal{F}_\Gamma$}\\
  0 & \text{if $\mathbf{w}\notin \mathcal{F}_\Gamma$}
  \end{array} \right.
\end{equation}
In addition one is interested in deriving functions on $\mathcal{F}_\Gamma$ that are eigenstates of the Laplacian:
\begin{equation}\label{eigvals}
  \Delta^{(k)}_{LB} \, \Phi^{\lambda}(\mathbf{w}) \, = \, \lambda \, \Phi^{\lambda}(\mathbf{w})
\end{equation}
Then, since any \textbf{regular function} on $\mathbb{H}^{2+k}$ is in particular a function on the fundamental domain
$\mathcal{F}_\Gamma \subset \mathbb{H}^{2+k}$ and since the spectrum of the Laplacian over $\mathbb{H}^{2+k}$ is already fixed
by Lie algebra theory in terms of the Casimir eigenvalues of $\so(1,2+k)$, one is tempted to conclude that the possible eigenvalues
and eigenfunctions of the Laplacian on $\mathcal{F}_\Gamma$ are already known; yet this is not quite correct. In the present case, the critical detail is the qualifier \textit{ regular}.
Certainly the harmonics, being polynomials in the solvable coordinates and their exponentials, are everywhere regular
over $\mathbb{H}^{2+k}$ and certainly one looks for functions (\ref{canis}) that are not only square integrable but also regular
in  $\mathcal{F}_\Gamma$, so that the harmonics are members of the $L^2( \mathcal{F}_\Gamma)$ functional space, yet the latter
contains also other functions, besides the harmonics and their series. Why? For the simple reason that there are functions
regular in $\mathcal{F}_\Gamma$ that can develop singularities out of it and, as such, cannot be easily described in terms
of the series development (\ref{harmosvilup}).
\par
Correspondingly, also the spectrum of the Laplacian, once restricted to the compact region $\mathcal{F}_\Gamma$, comes
under rediscussion. Another approach needs to be developed to solve the Laplace equation (\ref{eigvals}) more generally.
Such a different approach that we describe in the next subsection has a double interest in the context of PGTS  mathematical tools for
neural networks. On one side, it is related to the eigenvalue/eigenfunction problem for the compact region $\mathcal{F}_\Gamma$, on the other, it is related to the problem of the Heat Kernel and hence with the probability distribution functions over the hyperbolic manifolds.
\subsection{The alternative to harmonics: integral transforms and the hypergeometric function}
\label{scaccosavio}
The method of integral transform that we are going to describe is closely intertwined with the issue of the length spectrum of
Fuchsian group conjugacy classes in $\mathbb{H}^2$ tessellations and hence of primary closed geodesics on the corresponding Riemann surfaces, yet it has a more general application target also for other manifolds of the same Tits Satake universality class, so that, for a while, we keep
the parameter $k$ arbitrary, before fixing it to $k=0$, in relation with the tessellation of the base manifold of our Tits Satake vector bundles.
\par
\begin{figure}
\begin{center}
\vskip 1cm
\includegraphics[width=80mm]{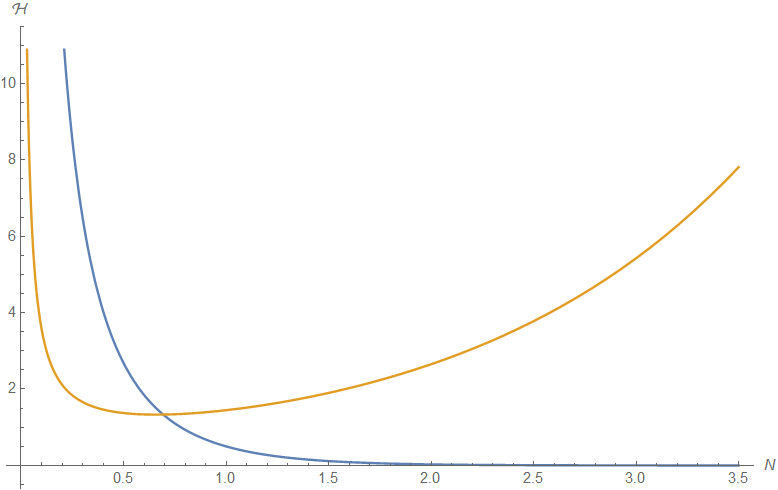}
\vskip 1cm
\caption{\label{crescisiono} Behavior of the two solutions of the Laplace equation once rephrased in Hypergeometric terms. The yellow line is the first solution, while the blue line is the second solution that decays exponentially at infinity.}
\end{center}
\end{figure}

We start from the notion of norm of the solvable Lie group elements that, as explained in \cite{pgtstheory}, is also the distance of the corresponding point in $\mathbb{H}^{(2+k)}$ from the origin, namely from the identity element of $\mathcal{S}_{\mathrm{U/H}}$.
Quoting from equation (5.12) of \cite{pgtstheory}, we have:
\begin{eqnarray}\label{norm}
 \mathrm{ N(\mathbf{w})} & = & \frac{1}{2} \, \text{arccosh}\left[\mathfrak{P}(\mathbf{w})\right]\nonumber\\
 \mathfrak{P}(\mathbf{w}) &= & \frac{1}{32} \left[e^{2 w_1} \left(w_2^2+
 \underbrace{\sum_{i}^k \, w_{2+i}^2}_{\text{subPaint invariant}}+4\right)^2+16 e^{-2 w_1}+8
   \left(w_2^2+\underbrace{\sum_{i}^k\,w_{2+i}^2}_{\text{subPaint invariant}}\right)\right]
\end{eqnarray}
Let us next consider the following variable:
\begin{equation}\label{zoniko}
 z =  z\left[\mathbf{w}\right]\, \equiv \, \frac{1}{\cosh^2\left(\mathrm{ N(\mathbf{w})}\right)}
\end{equation}
If we apply the Laplacian operator $\Delta^{(k)}_{LB}$ to a function $H\left(z\left[\mathbf{w}\right]\right)$ we obtain:
\begin{equation}\label{camillus}
  \Delta^{(k)}_{LB}H\left(z\left[\mathbf{w}\right]\right) \,=\, -4 (z-1) z^2 H''(z)-2 z (k-1\,+\,3 z) H'(z)
\end{equation}
where, for brevity, on the right-hand side of the equation, we omitted the dependence of $z$ from its argument $\mathbf{w}$.
Hence the function $H\left(z\left[\mathbf{w}\right]\right)$ is an eigenfunction of  Laplace Equation:
\begin{equation}\label{cirimella}
  \Delta^{(k)}_{LB}H\left(z\left[\mathbf{w}\right]\right)\, = \, \lambda^2 \, H\left(z\left[\mathbf{w}\right]\right)
\end{equation}
if, as a function of the argument $z$, the function  $H(z)$ satisfies the $2nd$ order differential equation:
\begin{equation}\label{cartulla}
 -4 (z-1) z^2 H''(z)-2 z (k-1\,+\,3 z) H'(z)-\lambda ^2 H(z) \, = \, 0
\end{equation}
Equation (\ref{cartulla}) is related to the hypergeometric differential equation, and its general integral is given by a linear combination
of the following two solutions:
\begin{eqnarray}\label{geninteg}
  {H}_1(z) &=& z^{\frac{1}{4} \left(-\sqrt{k^2+4 \lambda ^2+2 k+1}+k+1\right)} \,
   _2F_1\left(\frac{k}{4}-\frac{1}{4} \sqrt{k^2+2 k+4 \lambda
   ^2+1}+\frac{1}{4},\frac{k}{4}-\frac{1}{4} \sqrt{k^2+2 k+4 \lambda
   ^2+1}+\frac{3}{4};\right.\nonumber\\
   &&\left.1-\frac{1}{2} \sqrt{k^2+2 k+4 \lambda ^2+1};z\right)\nonumber \\
  {H}_2(z) &=& z^{\frac{1}{4} \left(\sqrt{k^2+4 \lambda ^2+2 k+1}+k+1\right)} \,
   _2F_1\left(\frac{k}{4}+\frac{1}{4} \sqrt{k^2+2 k+4 \lambda
   ^2+1}+\frac{1}{4},\frac{k}{4}+\frac{1}{4} \sqrt{k^2+2 k+4 \lambda
   ^2+1}+\frac{3}{4} ;\right.\nonumber\\
   && \left.\frac{1}{2} \sqrt{k^2+2 k+4 \lambda ^2+1}+1;z\right)
\end{eqnarray}
Converting $z$ to the original argument $\mathrm{N(\mathbf{w})}$ via equation (\ref{zoniko}) we obtain two general solutions of Laplace Equation,
that depend on the point in the manifold only through the  norm $\mathrm{N(\mathbf{w})}$ of the associated solvable Lie group element:
  \begin{eqnarray}\label{moscacieca}
 {H}_1\left(\lambda^2,k,\mathrm{N}\right) &=& \text{sech}^2(\mathrm{N})^{\frac{1}{4} \left(-\sqrt{k^2+4 \lambda ^2+2 k+1}+k+1\right)} \,
   _2F_1\left(\frac{k}{4}-\frac{1}{4} \sqrt{k^2+2 k+4 \lambda
   ^2+1}+\frac{1}{4}, \right.\nonumber\\
   &&\left. \frac{k}{4}-\frac{1}{4} \sqrt{k^2+2 k+4 \lambda
   ^2+1}+\frac{3}{4};
   1-\frac{1}{2} \sqrt{k^2+2 k+4 \lambda ^2+1};\text{sech}^2(\mathrm{N})\right)\nonumber \\
{H}_2\left(\lambda^2,k,\mathrm{N}\right) &=& \text{sech}^2(\mathrm{N})^{\frac{1}{4} \left(\sqrt{k^2+4 \lambda ^2+2 k+1}+k+1\right)} \,
   _2F_1\left(\frac{k}{4}+\frac{1}{4} \sqrt{k^2+2 k+4 \lambda
   ^2+1}+\frac{1}{4},\right.\nonumber\\
   &&\left.\frac{k}{4}+\frac{1}{4} \sqrt{k^2+2 k+4 \lambda
   ^2+1}+\frac{3}{4};\frac{1}{2} \sqrt{k^2+2 k+4 \lambda ^2+1}+1;\text{sech}^2(\mathrm{N})\right)
  \end{eqnarray}
  The boundary condition that requires a decreasing (actually exponentially decreasing) behavior as $\mathrm{N}\to \infty$,
  uniquely selects the second solution as we can see from fig.\ref{crescisiono}.
Hence we set:
\begin{equation}\label{carolus}
  \mathcal{K}(\lambda^2,k,N) \,\equiv\, {H}_2\left(\lambda^2,k,\mathrm{N}\right)
\end{equation}
The function $\mathcal{K}(\Lambda ,k,N) $ is the Green function for the operator $\mathcal{O} = \Delta^{(k)}_{LB} -\Lambda$.
Let us show how. Let $\gamma \in \mathrm{SO(1,2+k)}$ be any element of the isometry group, and let $\mathit{f}$ be a
smooth real function on $\mathbb{H}^{(2+k)}$, namely:
\begin{equation}\label{canocchia}
  \forall z \in \mathbb{H}^{(2+k)} \quad : \quad \mathit{f}(z) \in \mathbb{R}
\end{equation}
We define the action $T_{\gamma}$ on the space of functions by setting:
\begin{equation}\label{cricca}
  T_{\gamma}[\mathit{f}](z) \equiv \mathit{f}(\gamma(z))
\end{equation}
Let moreover denote by:
\begin{equation}\label{kricetus}
  \mathit{g} \, = \,\Delta^{(k)}_{LB}\left[\mathit{f}\right]
\end{equation}
  the new function $\mathit{g} $ one obtains by applying the Laplacian differential operator to the function $\mathit{f}$.
  The Laplacian is an invariant operator since one has:
  \begin{equation}\label{balbecco}
    \Delta^{(k)}_{LB}\left[T_\gamma\left[\mathit{f}\right]\right] \, = \, T_\gamma\left[\Delta^{(k)}_{LB}\left[\mathit{f}\right]\right]\quad ;
    \quad \forall \gamma \in  \mathrm{SO(1,2+k)} \quad \forall \mathit{f} \in \mathbb{C}^{\infty}(\mathbb{H}^{(2+k)})
  \end{equation}
  Consider now the function
  \begin{equation}\label{minosse}
    \mathit{g}^\Lambda(\mathbf{v}) \equiv \int_{\mathbb{H}^{(2+k)}} \mathcal{K}\left(\Lambda,k,\mathrm{N}\left(\mathbf{u}^{-1}\cdot \mathbf{v}\right)
    \right)\, \mathit{f}(\mathbf{u})\,d\mu(\mathbf{u})
    \end{equation}
    If we act on $\mathit{g}^\Lambda(\mathbf{v})$  with the Laplacian operator what we actually have to calculate is:
    \begin{equation}\label{kinofilo}
      \Delta^{(k)}_{LB|\mathbf{v}}\left[\mathcal{K}\left(\Lambda,k,\mathrm{N}\left(\mathbf{u}^{-1}\cdot \mathbf{v}\right) \right)\right]
      \, = \, \text{?}
    \end{equation}
    where we have specified with respect to which coordinates we have to calculate the Laplacian, namely $\mathbf{v}$.
    The key point is that the multiplication of an element of the solvable group, in this case $\mathbf{v}$ by another element of the same,
    in this case $\mathbf{u}^{-1}$ is an isometry (just one of the many transformations of $\mathrm{SO(1,2+k)}$, so we have:
    \begin{eqnarray}\label{chicco}
      &&\Delta^{(k)}_{LB}\left[T_{u^{-1}}\left[\mathcal{K}\left(\Lambda,k,\mathrm{N}\left(\mathbf{v}\right) \right)\right]\right]\, =
      \, \left[T_{u^{-1}}\left[\Delta^{(k)}_{LB}\mathcal{K}\left(\Lambda,k,\mathrm{N}\left(\mathbf{v}\right) \right)\right]\right]
      \, = \, \Lambda \, \mathcal{K}\left(\Lambda,k,\mathrm{N}\left(\mathbf{u}^{-1} \cdot\mathbf{v}\right) \right)
    \end{eqnarray}
 In conclusion we find:
\begin{equation}\label{pieratto}
  \left(\Delta^{(k)}_{LB} \, - \, \Lambda\right) \, \mathit{g}^\Lambda(\mathbf{v}) \, = \, 0
\end{equation}
\subsection{The heat kernel}
Let us know observe the following.  Suppose we are interested in constructing the Heat Kernel, that, as an integral transform:
\begin{equation}
  p\left(t,\rho(\mathbf{v})\right) \, = \, \int_{\mathbb{H}^{2+k}} \,\mathcal{H}^{k}\left(t, \rho\left({\mathbf{v}\cdot
  \mathbf{u}^{-1}}\right)\right)\,  p_0\left(\rho\left(\mathbf{u}\right)\right) \, d\mu\left(\mathbf{u}\right)
\end{equation}
provides the probability distribution $p(t,\rho)$ of occupation, at time $t$, of points $\mathbf{v}\in\mathbb{H}^2$ of norm $\rho = \mathrm{N}(\mathbf{v})$,
starting from a distribution $p_0(\rho)$, at time $t=0$. By definition the probability distribution $p(t,\rho)$ satisfies the parabolic
differential equation (see appendix \ref{markovka} for details):
\begin{equation}\label{goriziana}
\frac{\partial}{\partial t}\, p\left(t,\rho\right)  \, = \, \Delta^{(k)}_{LB} \, p\left(t,\rho\right)
\end{equation}
with boundary condition:
\begin{equation}\label{stecca}
  p\left(0,\rho\right) \, = \, p_0(\rho)
\end{equation}
In terms of the above developed Green function of the Laplacian it is natural to pose:
\begin{equation}\label{karamellus}
  \mathcal{H}^{k}\left(t, \rho\right) \, = \, \int \mathcal{K}\left(\Lambda,k,\rho\right)\, \exp\left[- t\, \Lambda\right] \, d\mu[\Lambda]
\end{equation}
where $d\mu[\Lambda]$ denotes a suitable integration measure on the Laplace eigenvalue $\Lambda$. Indeed the integrand of the integral
in eq.(\ref{karamellus}) satisfies, for each value of $\Lambda$, the parabolic equation (\ref{stecca}). The question is what is the suitable
integration measure $d\mu[\Lambda]$. We solve this problem by comparison with explicit results for the heat kernel on hyperbolic manifolds
$\mathbb{H}^{2+k}$ obtained in the literature \cite{buserbook,bielefeldo}.
\par
Considering the case $k=1$ of equations (\ref{moscacieca}-\ref{carolus}) we find that the corresponding Green function
\begin{eqnarray}\label{lineareB}
  \mathcal{K}\left(\Lambda,k,\rho\right)& = &  \text{sech}^2(\rho )^{\frac{1}{2} \left(\sqrt{\Lambda +1}+1\right)} \,
   _2F_1\left(\frac{1}{2} \left(\sqrt{\Lambda +1}+1\right),\frac{1}{2}
   \left(\sqrt{\Lambda +1}+2\right);\sqrt{\Lambda +1}+1;\text{sech}^2(\rho )\right) \nonumber\\
   & = & \frac{2^{\sqrt{\Lambda +1}} \left(1-{\tanh (\rho )}\right)^{\frac{1}{2}
   \left(\sqrt{\Lambda +1}+1\right)} \left({\tanh (\rho
   )}+1\right)^{\frac{1}{2}-\frac{\sqrt{\Lambda +1}}{2}}}{{\tanh (\rho )}}
\end{eqnarray}
admits the rewriting in terms of elementary transcendental functions displayed in the second line of the above equation.  Such simplification,
in this particular case, of the hypergeometric function  is what allows to calculate in closed form the integral of eq.(\ref{karamellus}).
On the other hand it was shown
in \cite{bielefeldo}  that the heat kernel for the same hyperbolic space is the following function:
\begin{equation}\label{garibaldi}
\mathcal{H}^{1}\left(t, \rho\right) \, = \, \sqrt{\frac{\pi }{2}} \,\frac{ \rho \, \text{csch}(\rho ) }{t^{3/2}} \, e^{-\frac{\rho ^2}{4
   t}-t}
\end{equation}
Setting
\begin{equation}\label{kamushko}
  q \, = \,\sqrt{\Lambda +1} \, \Rightarrow \, \Lambda = q^2 -1
\end{equation}
from eq.(\ref{lineareB}) we obtain:
\begin{equation}\label{tapisroulant}
 \mathcal{K}\left(q^2 -1,1,\rho\right) \, = \, 2^q \times \underbrace{\sqrt{2} \coth (\rho ) (1-\tanh (\rho ))^{\frac{q+1}{2}} (\tanh (\rho
   )+1)^{\frac{1}{2}-\frac{q}{2}}}_{\text{renorm. Green function}}
\end{equation}
where we have separated the factor $2^q$, which depends from $q$ but not from $\rho$, from its $\rho$-depending cofactor, that might equally well
be used as Green function at fixed $q$ and hence at fixed $\Lambda$. Hence let us give a separate name to such cofactor:
\begin{eqnarray}\label{carnenonvale}
  \mathcal{G}(q,\rho) & \equiv & \sqrt{2} \coth (\rho ) (1-\tanh (\rho ))^{\frac{q+1}{2}} (\tanh (\rho
   )+1)^{\frac{1}{2}-\frac{q}{2}} \nonumber\\
   & = & \frac{2 \sqrt{2}  \coth (\rho )}{e^{2 \rho }+1} \, \exp\left[\rho -q \rho \right]
\end{eqnarray}
and consider the reformulation of the integral (\ref{karamellus}) as follows:
\begin{equation}\label{riformaprotestante}
  \mathcal{H}^{k}\left(t, \rho\right) \, = \, \int \mathcal{G}(q,\rho) \, \exp\left[-(q^2-1) \,t \right] d\mu[q]
\end{equation}
If we set:
\begin{equation}\label{measure}
  d\mu[q] \, = \, q\, dq \, = \,\frac{1}{2} d\Lambda \, \equiv \, d\mu[\Lambda]
\end{equation}
we obtain:
\begin{equation}\label{integratallo}
  \int_{-\infty}^{\infty} \mathcal{G}(q,\rho) \, \exp\left[-(q^2-1) \,t \right] q\, dq \, = \, \sqrt{-1} \times
  \mathcal{H}^{1}\left(t, \rho\right)
\end{equation}
where the imaginary constant $\sqrt{-1}$ is simply the effect of calculating the integral with the residue method and can be factored out.
\par
The importance of the above result is that a general solution of the diffusion equation (\ref{goriziana}) can be represented with the integral
(\ref{karamellus}) where the integration measure on the laplacian eigenvalue $\Lambda$ is that given in eq.(\ref{measure}).
\par
The above discussion suggests that a viable substitute for the Heat Kernel in applications to machine learning Algorithms can  be the following
\textbf{spectral heat kernel}:
\begin{equation}\label{chocalatinus}
  {\mathcal{H}}_{spec}^{k}\left(t, \rho\right) \, = \, \ft 12 \sum_{\Lambda\in \text{spec}[k]} \mathcal{K}\left(\Lambda,k,\rho\right)\,
  \exp\left[- t\, \Lambda\right] \,
\end{equation}
where by $\text{spec}[k]$ we have denoted the infinite set of Casimir eigenvalues for the isometry Lie algebra $\so(1,2+k)$ representations
that contribute to the harmonic spectrum as discussed above and in \cite{pgtstheory} (see section 10 of this paper).
For instance in the case $k=0$ of the $\mathbb{H}^2$ Tits Satake submanifold of the whole class we have:
 \begin{equation}\label{speckern0}
    {\mathcal{H}}_{spec}^{0}\left(t, \rho\right)\, = \, \ft 12 \sum_{J=1}^{\infty} \, \mathcal{K}\left(J(J+1),0,\rho\right)\,
  \exp\left[- J(J+1)\, \Lambda\right]
 \end{equation}
 The above expression that can be easily generalized to all values of $k$ (it suffices to fix the spectrum of Casimir eigenvalues
 on the corresponding manifold) can be truncated, when useful, to a maximal value of $\Lambda$) providing a useful tool for numerical analysis.
 Furthermore it provides a link between the harmonic expansion of functions and their diffusion with the heat kernel, whose potentialities have to be
 explored.
 \subsection{The length spectrum}
 \label{pigrone}
 The so named length spectrum (utilized especially in the case $k=0$ of the Tits Satake submanifold, namely in the case of $\mathbb{H}^{2}$ )
 arises from the following concept of \textbf{length of an isometry}.
 \par
 Consider the distance formula:
 \begin{equation}\label{citrullica}
   d^2\left(\mathbf{u},\mathbf{v}\right) \, = \, \mathrm{N}^2\left( \mathbf{u}^{-1}\cdot \mathbf{v}\right)
 \end{equation}
 \begin{definizione}
 \label{defilungo}
   Let $g \in \mathrm{G_{iso}}$ be any group element of the isometry group and consider the following:
   \begin{equation}\label{cudrino}
     \ell(g) \, = \, \underbrace{\text{inf}}_{\mathbf{p} \in \mathbb{H}^{(2+k)}} \, d\left(\mathbf{p},g[\mathbf{p}]\right)
   \end{equation}
   The number $\ell(g)$ is named the \textbf{the length of the isometry} $g$
 \end{definizione}
 As one sees, the above definition \ref{defilungo} applies equally well, \textit{mutatis mutandis}, to all members of the $r=1$ universality class.
 In the case $k=0$, where $\mathrm{G_{iso}}=\mathrm{PSL(2,\mathbb{R})}\simeq \mathrm{SO(1,2)}$, we show below that for every $g\in \mathrm{G_{iso}}$
 we have:
 \begin{equation}\label{currentopoli}
   \ell(g) \, = \, 2 \, \text{arccosh}\left(\ft 12 \, \text{Tr} (\mathrm{m}_g) \right)
 \end{equation}
 where $m_g \in\mathrm{PSL(2,\mathbb{R})}$ denotes the $2\times 2$ matrix representation of the abstract group element.
 \par
 Let us proceed as far as we can at a generic $k$ level in the analysis of the definition \ref{defilungo}. The first observation is that the length
 of isometry elements is a property not of individual group elements, rather of conjugacy classes of elements. Indeed since the
 distance function is invariant under isometries, namely we have
 \begin{equation}
   \forall \gamma \in \mathrm{G_{iso}} \quad ; \quad d\left(\gamma[\mathbf{p}_1]\, , \, \gamma[\mathbf{p}_2]\right) \, = \,
   d\left(\mathbf{p}_1\, , \, \mathbf{p}_2\right)
 \end{equation}
 it follows that:
 \begin{eqnarray}
\forall \gamma,g \in \mathrm{G_{iso}} \quad ; \quad
\underbrace{\text{inf}}_{\mathbf{p} \in \mathbb{H}^{(2+k)}}d\left(p\, , \, \gamma^{-1}.g.\gamma[p]\right) & =
& \underbrace{\text{inf}}_{\mathbf{p} \in \mathbb{H}^{(2+k)}} d\left(\gamma[\mathbf{p}]\, , \, g[\gamma[\mathbf{p}]] \right) \, = \,
\underbrace{\text{inf}}_{\mathbf{p}^\prime \in \mathbb{H}^{(2+k)}} d\left(\mathbf{p}^\prime\, , \, g[\mathbf{p}^\prime] \right)
 \end{eqnarray}
 since the manifold $\mathbb{H}^{(2+k)}$ is invariant under the action of any isometry and since any isometry is surjective.
 Therefore one has:
 \begin{equation}\label{arsenico}
\forall \gamma,g \in \mathrm{G_{iso}} \quad ; \quad \ell\left(\gamma^{-1}\cdot g \cdot \gamma\right) \, = \, \ell\left(g\right)
 \end{equation}
Relying on the above property and on the solvable subgroup decomposition, we easily see that there are three types of conjugacy classes
in $\mathrm{SO(1,2+k)}$:
\begin{description}
  \item[\textbf{Parabolic classes})] These are the conjugacy classes of solvable group elements obtained as the exponential of \textbf{nilpotent
  elements} of the solvable Lie algebra, corresponding to finite group elements that have only $1$.s on the diagonal.
  In the $k=0$ case, \textit{i.e.} for $\mathrm{PSL(2,\mathbb{R})}$, the standard representative of the unique parabolic class is of the form:
  \begin{equation}\label{parbollito}
  g_{\mathrm{par}} \, = \, \left(\begin{array}{cc}
          1 & b \\
          0 & 1
        \end{array}
   \right) \quad ; \quad b\in \mathbb{R}
  \end{equation}
  \item[\textbf{Hyperbolic classes})] These are the conjugacy classes of solvable group elements obtained as the exponentials of
  \textbf{Cartan elements} of the solvable Lie algebra, and correspond to finite group elements that are diagonal, but with non-trivial diagonal entries.  In the $k=0$ case, \textit{i.e.} for $\mathrm{PSL(2,\mathbb{R})}$, the standard representatives of each hyperbolic class is
  of the form:
   \begin{equation}\label{iperbolino}
   g_{\mathrm{hyp}} \, = \, \left(\begin{array}{cc}
          a & 0 \\
          0 & 1/a
        \end{array}
   \right) \quad ; \quad a \in[0 ,\infty) \quad  \& \quad  a \neq 1 \quad  \&  \quad a \neq 0
   \end{equation}
  \item[\textbf{Elliptic classes})] These are the conjugacy classes of solvable group elements belonging to the maximal compact subgroup
  $\mathrm{H}=\mathrm{SO(2+k)}$. In the $k=0$ case, \textit{i.e.} for $\mathrm{PSL(2,\mathbb{R})}$, the standard representatives of each
  hyperbolic class is of the form:
   \begin{equation}
   g_{\mathrm{ellip}} \, = \, \left(\begin{array}{cc}
          \cos[\theta] & \sin[\theta] \\
          -\sin[\theta] & \cos[\theta]
        \end{array}
   \right) \quad ; \quad \theta \in [0,2\pi]
   \end{equation}
\end{description}
It suffices therefore to calculate the isometry length, as defined in definition \ref{defilungo}, for each of their three types,
namely parabolic, hyperbolic or elliptic. In the $k=0$ case, relying on the linear fractional transformation action of the group isometries and on
the explicit form of the norm, as given in eq.s (\ref{norm}) one finds that for both parabolic and elliptic isometries their length vanishes:
\begin{equation}\label{svampito}
  \ell(g_{par}) \, = \, \ell(g_{ellip}) \, = \, 0
\end{equation}
while it is non vanishing only for the hyperbolic elements:
\begin{equation}\label{castrum}
  \ell(g_{\mathrm{hyp}}) \, > \, 0
\end{equation}
In particular using eq.(\ref{norm}) and the standard representative (\ref{iperbolino}) we obtain:
\begin{equation}\label{cruscotto}
  \ell(g_{\mathrm{hyp}}) \, = \, \ft 12 \, \text{arccosh}\left[\frac{1+a^8}{2\, a^4} \right] \, = \, 2 \text{arccosh}\left[\ft 12 \,
  \left(a + \frac{1}{a}\right) \right]
\end{equation}
The last equation holds true for $a>1$ which is general (interchanging $a$ with $1/a$).
The three results (\ref{svampito}, \ref{castrum},\ref{cruscotto}) are consistent with equation (\ref{currentopoli}).
\par
Although we have not proved it with the necessary detailed calculations, we \textbf{conjecture} that eqs (\ref{svampito}) and (\ref{castrum}) hold
true also for the cases $k>0$. If the previous conjecture can be confirmed, it provides a quite relevant conceptual token about the discrete subgroups discussed in \cite{pgtstheory}. Similarly to what happens for the hyperbolic plane, we might consider \textit{generalized Fuchsian
groups} defined as discrete discontinuous groups made only of hyperbolic elements that might be organized in a length spectrum and explore
the construction of fundamental domains for the latter as fundamental tiles for the tessellation of hyperbolic spaces of higher dimensions.
\section{An exercise: mapping images to the Bolza surface}
\label{pitture}
In connection with the final comments of the previous subsection and with what we
anticipated in Section \ref{intergioco}, we present now a very rough outline of what might be the use of all the previously exposed lore in an algorithm for treating and representing \textit{images}, considering this a first exploratory step to treat data with a similar fiber bundle section structure. This type of data has already been the target of Cartan networks presented in the twin papers \cite{TSnaviga,naviga}; there, we relied on the naive machine learning approach, in which an image is treated as a datum vector mapped to a single point of the target manifold. The goal pursued was only class separation, no attempt
being made at the retrieval of geometric properties of surfaces. Considering the
latter as a more ambitious and highly desirable goal to be attained, we outline the steps of a possible algorithm focused on representing images as fiber bundle sections over a compact subset of the hyperbolic plane, emphasizing the difficulties and the missing required ingredients. Mapping images to the fiber bundle structure of the Bolza structure is a first, very important exercise in the construction of equivariant networks over the hyperbolic plane, but the reason we choose pictures as data is purely dimensional.
\par
\subsection{Map of the Euclidean Plane \texorpdfstring{$\mathbb{R}^2$}
{R2} to the Hyperbolic Plane
\texorpdfstring{$\mathbb{H}^2$}{H2}} 
\begin{figure}
\begin{center}
\vskip 1cm
\includegraphics[width=80mm]{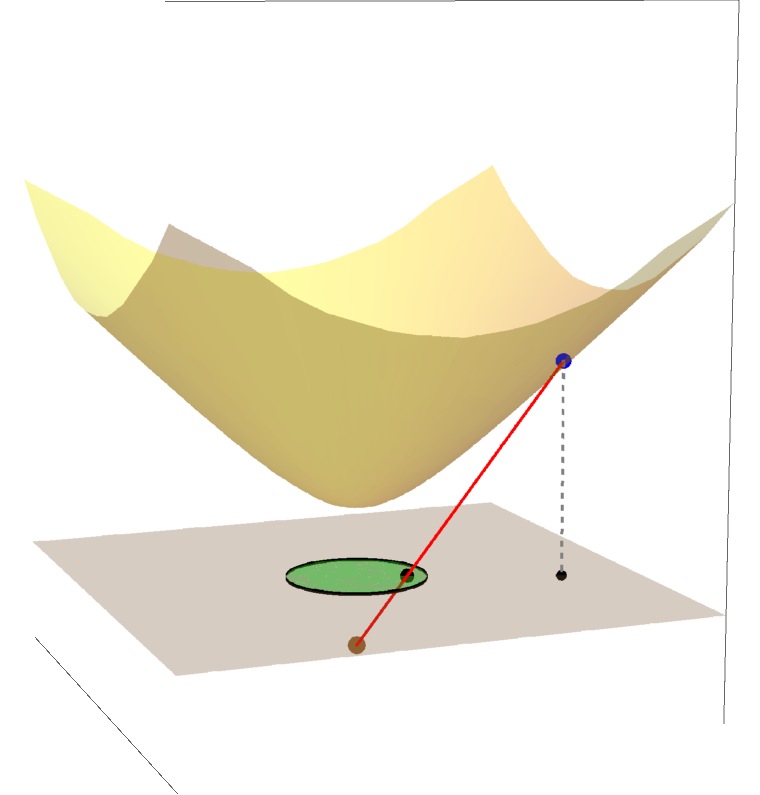}
\vskip 1cm
\caption{\label{rinfresco} Geometrical model of the $\mathbb{H}^2$ plane as a hyperboloid. Note the stereographic projection of any point
of the hyperboloid, marked in blue, to a point in the unit disk at $X_3=0$. The last is defined as the intersection (marked in black)
with the latter of the straight line segment (marked red) that joins the original hyperboloid point  with the point $\{0,0,-1\}$ (brown).
Note also the ordinary,
vertical projection (dashed line), from the hyperboloid to the plane. Hence a generic point $p$ of the hyperboloid has two projections,
the vertical one and
stereographic one.
 }
\end{center}
\end{figure}

The first step in the above-defined perspective construction is the map of the Euclidean plane $\mathbb{R}^2$ to the Hyperbolic Plane $\mathbb{H}^2$.
\begin{equation}\label{Uphold}
  \amalg_{EH} \, : \, \mathbb{R}^2 \, \longrightarrow \, \mathbb{H}^2
\end{equation}
To this effect, it is convenient to start from the geometrical model of the $\mathbb{H}^2$ space as
a hyperboloid in $\mathbb{R}^3$ defined as a connected component of  the following quadratic locus in $\mathbb{R}^3$
 \begin{equation}\label{hyperboloidus}
   X_1^2 \, + \, X^2_2 \, - \, X^2_3 \, =\, -1 \, ,
 \end{equation}
Indeed there are two disconnected  components in the solution to eq.(\ref{hyperboloidus}) and we choose only one of them, namely
the upper one as in fig.\ref{rinfresco}.
\par
Analytically the map is described as follows. If we denote $\{x,y\}\in \text{Disk}$ the cartesian coordinates in the unit disk,
the explicit form of the stereographic projection from the hyperboloid to the unit disk is the following:
\begin{equation}\label{downmapst}
\Pi_{st}\quad : \quad \text{Hyp} \, \longrightarrow \, \text{Disk}\quad ; \quad \{x,y\} \, = \,
\Pi_{st}\left[\left\{X_1,X_2,X_3\right\}\right] \,  = \, \left\{\frac{X_1}{X_3+1},\frac{X_2}{X_3+1}\right\}
\end{equation}
On the other hand the simple vertical projection from the hyperboloid to the flat Euclidean plane $\mathbb{R}^2$ whose coordinates we denote
$\{X,Y\}\in \mathbb{R}^2$ is simply given by:
\begin{equation}\label{downmapvert}
\Pi_{v}\quad : \quad \text{Hyp} \, \longrightarrow \, \text{$\mathbb{R}^2$} \quad ;
\quad \{X,Y\} \, = \,\Pi_{v}\left[\left\{X_1,X_2,X_3\right\}\right] \,  = \, \left\{X_1,X_2\right\}
\end{equation}
Taking into account the quadratic constraint \ref{hyperboloidus} which defines the hyperboloid, the above vertical projection map is invertible
and we have:
\begin{equation}\label{upmapvert}
\Pi_{v}^{-1}\quad : \quad \text{$\mathbb{R}^2$} \, \longrightarrow \, \text{Hyp} \quad ;
\quad \{X_1,X_2,X_3\} \, = \,\Pi^{-1}_{v}\left[\left\{X,Y\right\}\right] \,  = \, \left\{X,Y,\sqrt{X^2+Y^2+1}\right\}\in \text{Hyp}
\end{equation}
where we have taken in account the branch choice in order to map back to the upper connected component.
\par
\begin{figure}
\begin{center}
\vskip 1cm
\includegraphics[width=80mm]{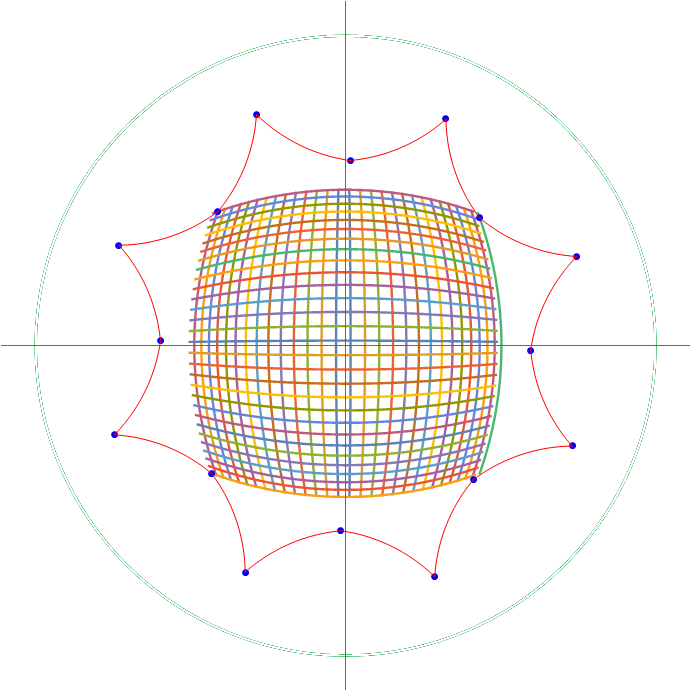}
\includegraphics[width=80mm]{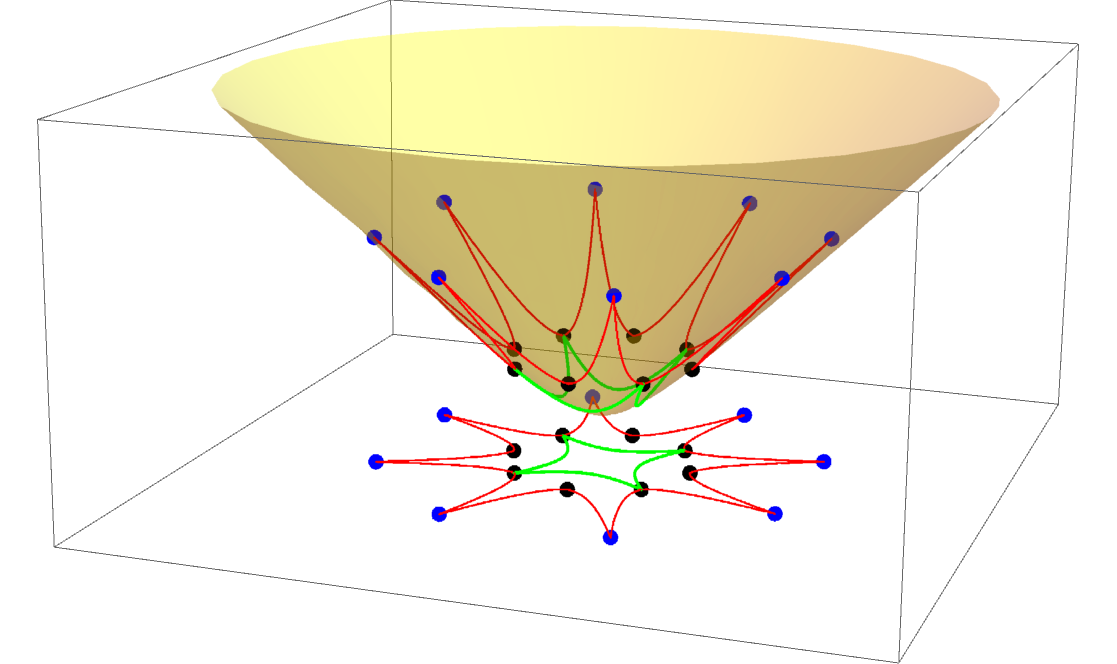}
\vskip 1cm
\caption{\label{bolzanetoquadro} In the figure on the left we show how the square grid representing a typical photo image appears, once mapped to the
disk model of the hyperbolic plane, in such a way as to be inserted into the fundamental domain of a Fuchsian group, in this case that of the
genus $2$ Bolza surface. Conversely, on the right, we show how the fundamental domain boundary is traced on the hyperboloid and mapped down
by the vertical projection onto $\mathbb{R}^2$ }
\end{center}
\end{figure}

 The composition of  two maps, \textit{i.e.} the stereographic projection with the uplift map $\Pi_{v}^{-1}$,
provides the desired map of eq.(\ref{Uphold})
\begin{eqnarray}\label{carnegie}
   \amalg_{EH} & = &\Pi_{st}\circ\Pi_{v}^{-1}  \, : \, \mathbb{R}^2 \, \longrightarrow \, \mathbb{H}^2 \nonumber\\
    \amalg_{EH}\left[\{X,Y\}\right] &=& \Theta(X) \, \left\{\frac{X^2}{\sqrt{X^2 \left(X^2+Y^2+1\right)}+X},\frac{X Y}{\sqrt{X^2
   \left(X^2+Y^2+1\right)}+X}\right\} \nonumber\\
   && + \Theta(-X) \,\left\{-\frac{\sqrt{X^2 \left(X^2+Y^2+1\right)}+X}{X^2+Y^2},\frac{X Y}{X-\sqrt{X^2
   \left(X^2+Y^2+1\right)}}\right\}
\end{eqnarray}
The inverse map from the Disk to $\mathbb{R}^2$ is also well defined and we have:
\begin{eqnarray}\label{cartolinapostale}
   \amalg_{EH}^{-1} & = &\Pi_{v}\circ\Pi_{st}^{-1}  \, : \, \mathbb{H}^2 \, \longrightarrow \, \mathbb{R}^2 \nonumber\\
    \amalg_{EH}^{-1}\left[\{x,y\}\right] &=& \left\{-\frac{2 x}{x^2+y^2-1},-\frac{2 y}{x^2+y^2-1}\right\}
\end{eqnarray}
Let us now consider a typical grid portion of a planar square lattice that might represent an image in small resolution. Such a finite grid is composed of $28$ equally spaced vertical straight lines and $28$  equally spaced horizontal straight lines.
Consider next the Bolza fundamental domain as described in fig.\ref{crullino} and the interior geodesic square that has its four vertices on the fundamental domain boundary $\partial\mathcal{F}_\Gamma$, precisely located on four of the eight Q-points (midpoints of the eight edges).
We can arrange the grid in such a way that its four vertices map to the four above-mentioned Q-points of the Bolza  fundamental
domain boundary $\partial\mathcal{F}_\Gamma$.
We set:
\begin{alignat}{4}\label{verticolini}
  \mathbf{v}_1 & = &  Q_2 & = & \left\{-\frac{1}{2} \sqrt{6-\frac{7}{\sqrt{2}}+\sqrt{28 \sqrt{2}-\frac{77}{2}}}
   \rho _s,-\frac{\rho _s}{2 \sqrt{\frac{2}{-4+7 \sqrt{2}-\sqrt{14 \left(8
   \sqrt{2}-11\right)}}}}\right\} \nonumber\\
  \mathbf{v}_2 & = &  Q_4 & = & \left\{\frac{1}{4} \sqrt{-8+14 \sqrt{2}-2 \sqrt{14 \left(8 \sqrt{2}-11\right)}}
   \rho _s,-\frac{1}{2} \sqrt{6-\frac{7}{\sqrt{2}}+\sqrt{28
   \sqrt{2}-\frac{77}{2}}} \rho _s\right\}  \nonumber\\
  \mathbf{v}_3 & = &  Q_6 & = & \left\{\frac{1}{2} \sqrt{6-\frac{7}{\sqrt{2}}+\sqrt{28 \sqrt{2}-\frac{77}{2}}}
   \rho _s,\frac{1}{4} \sqrt{-8+14 \sqrt{2}-2 \sqrt{14 \left(8
   \sqrt{2}-11\right)}} \rho _s\right\} \nonumber\\
  \mathbf{v}_4 & = &  Q_8 & = & \left\{-\frac{\rho _s}{2 \sqrt{\frac{2}{-4+7 \sqrt{2}-\sqrt{14 \left(8
   \sqrt{2}-11\right)}}}},\frac{1}{2} \sqrt{6-\frac{7}{\sqrt{2}}+\sqrt{28
   \sqrt{2}-\frac{77}{2}}} \rho _s\right\}
\end{alignat}
where $\rho^\star$ is defined in equation (\ref{rostarro}) and we define the preimages of such $4$ vertices in the $\mathbb{R}^2$ plane:
\begin{equation}\label{vertoni}
  \mathbf{V}_i \, \equiv \, \amalg_{EH}^{-1}\left[\mathbf{v}_i\right] \quad ; \quad (i=1,\dots,4)
\end{equation}
Naming $V^{a=1,2}_{i=1,\dots, 4}$ the components of $\mathbf{V}_i$ we can subdivide the square in $\mathbb{R}^2$ that has the four vertices
(\ref{vertoni}) into a grid traced by 28 horizontal and 28 vertical lines that we define as follows:
\begin{alignat}{4}\label{grid}
  {\mathrm{Hor}}_k(t) & = & \left\{t \left(V_2^1-V_1^1\right)+V_1^1,\quad\frac{1}{28} k
   \left(V_4^2-V_1^2\right)+V_1^2\right\}\quad;\quad  &\quad  k=1,\dots, 28 \quad ;\quad &\quad  t\in [0,1] \nonumber \\
  {\mathrm{Vert}}_k(t) & = & \left\{\frac{1}{28} k \left(V_4^1-V_3^1\right)+V_3^1,\quad t
   \left(V_4^2-V_1^2\right)+V_1^2\right\}\quad;\quad  &\quad  k=1,\dots, 28 \quad ;\quad &\quad  t\in [0,1]
\end{alignat}
We obtain the image of the square grid (\ref{grid}) inside the Bolza fundamental domain by using the map $\amalg_{EH}$ as defined
in eq.(\ref{carnegie}):
\begin{alignat}{4}
\mathit{h}_k(t) & = & \amalg_{EH}\left[{\mathrm{Hor}}_k(t) \right] \quad;\quad  &\quad  k=1,\dots, 28 \quad ;\quad &\quad  t\in [0,1] \nonumber \\
\mathit{v}_k(t) & = & \amalg_{EH}\left[{\mathrm{Vert}}_k(t) \right] \quad;\quad  &\quad  k=1,\dots, 28 \quad ;\quad &\quad  t\in [0,1] \nonumber \\
\end{alignat}
The result is shown in fig.\ref{bolzanetoquadro}.
The key point is that the images $\mathit{h}_k(t),\mathit{v}_k(t)$ of the straight lines
${\mathrm{Hor}}_k(t) , {\mathrm{Vert}}_k(t)$ are not geodesics of the hyperbolic plane and
vice-versa the images in $\mathbb{R}^2$ of hyperbolic plane geodesics are not straight,
namely are not geodesics of the target manifold. What matters is that in $2$ dimensions
all metrics are conformally flat since, by choosing coordinates appropriately, we can
always put any metric in the form $ds^2(x) = \Omega(x)ds^2_{flat}(x)$  where
$\Omega(x)$ is a positive function. Therefore, the uploading to $\mathbb{H}^2$ of a photo-grid is a conformal map that does not preserve lengths but preserves shapes and
angles.
\subsection{Peculiarities of the Bolza surfaces}
What allows us to map "images" to Riemann surfaces of genus $g\geq 2$ is the
uniformization theorem. All surfaces with $g\geq 2$ are hyperbolic, and it is only with
such a map that we can insert a fiber bundle into the hyperbolic machine, able to act on
the data without point-wise activation functions.

The intrinsic non-abelian character of the Fuchsian group is, on the one hand, the major difficulty of the problem, but can also be an advantage, once this difficulty is overcome. The quotient of the tessellation group with
respect to the Fuchsian subgroup produces a finite non-abelian quotient group
$\mathrm{GD_{index}}$ whose action on the original data space is nontrivial, raising hopes of unveiling hidden correspondences or even approximate symmetries. The choice of Bolza surface is also interesting for several reasons:
\begin{enumerate}
  \item It has the minimal genus $g=2$. \item In genus $g=2$ it is a maximal surface, since it has the maximal order of the automorphism group, namely $48$.
  \item It is maximal also from the point of view of the systoles (primitive closed geodesics with extremal length). Bolza surface maximizes
  the systole lengths \cite{extrasistola}.
  \item The Fuchsian group is arithmetic, according to the definition recalled in \cite{Attar_2022}, and, for the same reason, the length spectrum
  is based on a simple lattice $m+n\sqrt{2}$, $m,n \in \mathbb{Z}$
\end{enumerate}

\subsection{A schematic review of Selberg pair method for eigenvalues and eigenfunctions of the Laplacian}
\label{collinare}
 As we emphasized in sect. \ref{cirimello}, implementing the action of the Fuchsian group generators on the regular harmonics, one does not
 find any regular harmonic that is invariant against the action of $\Gamma$; this appears to be a rather general result. The essential
 reason for this failure is that in the fundamental irreducible representation of the full group $\mathrm{SO(1,2)}$, the 3-vector one,
 there is no Fuchsian invariant subspace;  this implies that the same remains true in all non-trivial tensor products of the latter.
 Therefore, one has to resort to different methods to find eigenfunctions and eigenvalues of the Laplacian in the fundamental domain $\mathcal{F}_\Gamma$. In the existing literature of the last two decades, a set of such methods centered around the notions of the  \textit{geodesic length spectrum} and the Selberg trace theorem \cite{vendicollina} has been developed and
 utilized \cite{buserbook,Chutassella,cuocogiuseppe,extrasistola,macloffo,Attar_2022,howarth2023}. We will provide a short summary in this section. These methods are insufficient for our purposes, as they do not provide a comprehensive set of tools. Indeed, we need an explicit basis of harmonics, square integrable in the fundamental domain, to calculate the coefficients of the harmonic expansion of a generic square integrable function defined over the domain, which is our goal:
 \begin{eqnarray}\label{cratarcone}
   \forall f(p) \in L^2(\mathcal{F}_\Gamma) \quad ; \quad f(p) & = & \sum_{\lambda\in\text{spectrum}}\sum_{m=1}^{\text{deg}[\lambda]} \, c_{\lambda}^m
   \, \Phi_m^\Lambda (p)\nonumber\\
   \vartriangle_{LB} \, \Phi_m^\Lambda (p) & = & \lambda \, \Phi_m^\Lambda (p) \nonumber\\
   n[\lambda] & = & \text{dim} \, \text{of an Irrep of the quotient group $\mathrm{GD}_{\mathrm{index}}$}
 \end{eqnarray}
 The Irreps of a finite group like $\mathrm{GD}_{\mathrm{index}}$ are in a finite number, equal to the number of conjugacy classes of the latter,
 that are therefore repeated over and over through the spectrum of $\lambda$ eigenvalues, which is instead infinite.
 The determination of the entire spectrum and, what is even more important, the explicit form of the functions $\Phi_m^\Lambda (p)$
 has not been determined for any of the Riemann surfaces of interest, neither in the older nor in the more recent literature we have been able to consult. This failure is probably intrinsic to the methods based on the Selberg trace conception that we are going to summarize in the following lines. We present them to outline the difficulties inherent in the problem and emphasize the need for different methods.
\subsubsection{Square integrable functions on \texorpdfstring{$\mathcal{F}_\Gamma$}{Fg}}
Following \cite{macloffo},\cite{cuocogiuseppe} we consider the action of the isometry group $\mathrm{Iso_{\mathbb{H}^2}} \, \sim\, \mathrm{SO(1,2)}$
on functions $f\in \mathrm{C}^{\infty}\left(\mathbb{H}^2 \right)$:
\begin{equation}\label{trumpet}
  \forall g\in\mathrm{Iso_{\mathbb{H}^2}} \quad ; \quad  T_g[f](p) \, = \, f(g^{-1}.p)
\end{equation}
and we define Fuchsian invariant functions as those that satisfy the condition:
\begin{equation}\label{fuchsian}
   T_\gamma[f] \, = \, f \quad \quad \forall \gamma \in \Gamma
\end{equation}
where $\Gamma \subset \mathrm{Iso_{\mathbb{H}^2}} $ is the relevant Fuchsian subgroup. The space of square integrable functions on the fundamental
domain $\mathcal{F}_\Gamma$ is made of those $\Gamma$-invariant functions on $\mathbb{H}^2$ whose squared norm  has a finite integral on
$\mathcal{F}_\Gamma$:
\begin{equation}\label{carabina}
  f \in L^2(\mathcal{F}_\Gamma) \quad \Leftrightarrow \quad  f\in \mathrm{C}^{\infty}\left(\mathbb{H}^2 \right) \quad \text{and}
  \quad T_\gamma[f] \, = \, f \quad \forall \gamma \in \Gamma \quad \text{and}
  \quad \int_{\mathcal{F}_\Gamma} \, |f(p)|^2 \, d\mu(p) \, < \, \infty
\end{equation}
The token used to study the Laplacian spectrum on $\mathcal{F}_\Gamma$ consists of 4 steps:
\begin{description}
  \item[1)] The definition of an \textbf{$h$-weighted bi-invariant point-pair kernel} $k^{(h)}(p_1,p_2)$.
  Starting from the Green function (\ref{lineareB}), choosing $k=0$ and renaming $\Lambda \to \lambda$ we first set:
  \begin{equation}\label{bardineto}
    \mathcal{G}\left(\lambda,\mathbf{v},\mathbf{u}\right) \, \equiv \, \mathcal{K}\left(\lambda,0,\mathrm{N}\left[\mathbf{v}.\mathbf{u}^{-1}\right]
    \right)
  \end{equation}
  Next choosing some \textit{test function}:
   \begin{equation}\label{pelandrone}
     h \quad : \quad \mathbb{R}_+ \, \longrightarrow \, \mathbb{R}_+
   \end{equation}
   that is required to be regular in the whole interval and decay exponentially at infinity, we define:
   \begin{equation}\label{cherubino}
     k^{(h)}(\mathbf{v},\mathbf{u})\, \equiv \, \int^{\infty}_{0} \,\mathcal{G}\left(\lambda,\mathbf{v},\mathbf{u}\right) \,
     h(\lambda) \, d\lambda  \quad \quad p_1 = \mathbf{v} \, ; \, p_2 = \mathbf{u}
   \end{equation}
  \item[2)] The construction of a \textbf{Selberg's point-pair invariant kernel} \cite{vendicollina}, namely of a two-point function
      $k^{(h)}_\Gamma(p_1,p_2)$, derived from   $k^{(h)}(\mathbf{v},\mathbf{u})$ that is invariant  with respect to the Fuchsian group $\Gamma$
      and it has the following properties:
  \begin{alignat}{3}\label{collinakernel}
    k^{(h)}_\Gamma &\quad :\quad\quad& \mathbb{H}^2\times \mathbb{H}^2 & \longrightarrow \quad \mathbb{C} \nonumber\\
    \forall p_{1,2} \in \mathbb{H}^2 &\quad :\quad\quad & k^{(h)}_\Gamma(p_1,p_2) & = k^{(h)}_\Gamma(p_2,p_1) \nonumber\\
    \forall \gamma \in \Gamma &\quad :\quad\quad  &  k^{(h)}_\Gamma(\gamma.p_1,p_2)& = k^{(h)}_\Gamma(p_1,p_2)
  \end{alignat}
  \item[3)] The introduction of an integral transform of functions $f$.s defined over $\mathcal{F}_\Gamma$ explicitly defined as follows:
  \begin{equation}\label{pantarhei}
    L^{h}[f](\mathbf{v}) \, \equiv \, \int_{\mathcal{F}_\Gamma} \, k^{(h)}_\Gamma\left(\mathbf{v},\mathbf{u}\right) \, f(\mathbf{u}) \,
    d\mu[\mathbf{u}]
  \end{equation}
  Because of the properties of the Green function explained in sect. \ref{scaccosavio}, we can easily demonstrate that, if the function
  $f(p)=\Phi^\lambda$ is an eigenfunction of the Laplacian as in eq. (\ref{cratarcone}), then its transform (\ref{pantarhei})
  is an eigenfunction of the linear transform with eigenvalue $h(\lambda)$:
  \begin{equation}\label{scagazzini}
    L^{h}[\Phi^\lambda_m](\mathbf{v}) \, =\, h(\lambda) \, \Phi^\lambda_m(\mathbf{v})
  \end{equation}
  \item[4)] Attempts, by means of educated ans\"atze, of the test function $h(\lambda)$, of the candidate harmonic $\Phi^\lambda_m$ and of the eigenvalue
  $\lambda$, utilizing various upper and lower bounds available from various arguments, to solve the eigenvalue problem (\ref{pantarhei}) (see, for instance,
  \cite{Chutassella} and \cite{cuocogiuseppe}). General information on the spectrum is obtained by the use of the Selberg trace identities, yet this information does not lead to the explicit determination of any of the harmonics.
\end{description}
As one sees, the two critical points in the implementation of the above method are the following:
\begin{enumerate}
\item Derivation of the $\Gamma$ invariant kernel $k^{(h)}_\Gamma$ from the $h$-weighted kernel $k^{(h)}$ defined in eq.(\ref{cherubino}).
\item Solution of the integral transform eigenvalue problem (\ref{pantarhei}).
\end{enumerate}
Let us first comment on point one. If the Fuchsian group were a finite group, the solution would be immediate:
\begin{equation}\label{trombatore}
  k^{(h)}_\Gamma(\mathbf{v},\mathbf{u}) \, = \, \sum_{\gamma \in \Gamma} \, k^{(h)}(\gamma.\mathbf{v},\mathbf{u})
\end{equation}
The problem is that the group is not finite, nor can its elements be easily enumerated and put in a summable generic form.
The infinite sum implicit  in eq.(\ref{trombatore}) is partially tamed using, in a quite complicated way, three weapons:
\begin{enumerate}
  \item The sum in (\ref{trombatore}) can be reduced to the conjugacy classes of $\Gamma$-group elements.
  \item The conjugacy classes can be associated with the complicated length spectrum of primitive elements  $\gamma_p$, namely
  those that are not powers of any other element.
  \item Finally, the kernel $k^{(h)}_\Gamma(\mathbf{v},\mathbf{u})$ can be written as a superposition of kernels for fundamental domains of
  one generator Fuchsian subgroups named \textit{hyperbolic cylinders}.
\end{enumerate}
As one sees, the whole machinery is very complicated and, from the perspective of our goals, leads to very modest results since, at best, few low-lying eigenvalues and harmonics can be constructed. For this reason, we tried to think in a different direction, inspired by the construction of UIR representations of the isometry group.
\subsection{The UIR of the hyperbolic isometry group, according to Bargmann}
 In his  1947 seminal paper \cite{barmanno},  Valentine Bargmann constructed the Unitary Irreducible Representations (UIR.s) of the groups
  $\mathrm{SO(1,2)}$ and $\mathrm{SO(1,4)}$ setting also conceptual standard for the unitary irreducible representations of all
  non-compact Lie groups that are all infinite dimensional. Such a standard can be spelled out as follows:
 \begin{enumerate}
   \item The carrier space of the UIR.s of a non-compact Lie group $\mathrm{G}$ is a Hilbert space realized as a \textit{ $L_{q}^2(\mathcal{M})$
   functional space} of square integrable functions over a Riemannian manifold $\mathcal{M}$ that admits the considered Lie group $\mathrm{G}$
   as the isometry group.
   \item The carrier space  $L_{q}^2(\mathcal{M})$ is specified by an integration measure:
   \begin{equation}\label{cratinus}
     d\mu_q(p) \, = \, \mu_q(p) \, \text{det}(g_{ij}(p))d^n p
   \end{equation}
   which depends on $q$, namely on the eigenvalue of the first quadratic Casimir invariant $\mathfrak{Cas}$  of the Lie algebra $\mathbb{G}$:
   \begin{equation}\label{Casimir}
     \mathfrak{Cas} \, \equiv \, \kappa_{A B} \, T^A \, T^B
   \end{equation}
   having denoted by $\kappa_{A B}$ the Killing metric.
   The algebraic Casimir operator $\mathfrak{Cas}$ corresponds to a \textit{quadratic invariant differential operator $\vartriangle_{LB}$}
   on $\mathcal{M}$.
   \item The action of the group elements $g$ on a function $f(p)$ of the carrier functional space is of the following form:
   \begin{equation}\label{carpamarinata}
     \forall g \in \mathrm{G} \quad , \quad \forall f(p) \in L_{q}^2(\mathcal{M}) \quad ; \quad T(g)[f](p) \, = \, \nu_q(p,g) \, f(g^{-1}.p)
   \end{equation}
   where the cocycle $\nu_q(p,g)$, or multiplier, as it was named by Bargmann, satisfies appropriate cocycle conditions to ensure that
   the operator $T[g]$ in eq.(\ref{carpamarinata}) defines a true group representation and furthermore makes the scalar product:
   \begin{equation}\label{canaglia}
     \left( f\, , \, h \right)_q \, \equiv \, \int_{\mathcal{M}} \bar{f}(p) \, h(p) \, d\mu_q(p)
   \end{equation}
   invariant with respect to $\mathrm{G}$:
   \begin{equation}\label{cranata}
     \forall g \in \mathrm{G} \quad , \quad  \left( T(g)[f]\, , \, T(g)[h] \right)_q \, = \, \left( f\, , \, h \right)_q
   \end{equation}
 \end{enumerate}
 For the group of interest to us, namely for $\mathrm{SO(1,2)}$ Bargmann found the following results.
 \begin{description}
   \item[a)] In all UIR.s the spectrum of eigenvalues $m$ of the unique Cartan generator $H$ is either integer or half-integer.
   There are four classes of UIR.s: two, where the spectrum of eigenvalues $q$ of the Casimir $\mathfrak{Cas}$ is continuous and
   two where it is discrete.
   \item[b)] \textbf{Class $C^0_q$}. $0< q < \infty$ and $m=0,\pm 1, \pm 2, \dots$.
   \item[c)] \textbf{Class $C^{\ft 12}_q$}. $\ft 14 < q < \infty$ and $m= \pm \ft 12, \pm \ft 32,\dots $.
   \item[d)] \textbf{Class $D_J^{+}$}. $q=J(J+1)$ where either $J=1,2, \dots$ or $J= \ft 12, \ft 23, \dots$.  Correspondingly either
   $m=0, 1,  2, \dots$, or $m=  \ft 12,  \ft 32,\dots $.
   \item[e)] \textbf{Class $D_J^{+}$}. $q=J(J+1)$ where either $J=1,2, \dots$ or $J= \ft 12, \ft 23, \dots$.  Correspondingly either
   $m=0, -1,  -2, \dots$, or $m=  -\ft 12, - \ft 32,\dots $
 \end{description}
 As one sees the Laplacian eigenvalues associated with the previously constructed regular harmonics cover only one part of the eigenvalue spectrum
 of the Laplacian, associated with UIR.s built on the basis of appropriate square integrable functions. Indeed with regular harmonics
 we retrieve only the eigenvalue spectrum $J(J+1)$ with $J$ integer. At the same time, as we have shown in section \ref{cirimello},
 using regular harmonics we cannot  find any basis of functions that are invariant against the action of the Fuchsian subgroup (either
 $\Gamma_{16}$ for the Fermat Quartic or $\Gamma_8$ for the Bolza surface and more generally for any other conceivable Fuchsian group.
 These two fact match each other. The Fuchsian invariant functions have to be looked for, not in the basis of regular harmonics, rather elsewhere.
\subsection{Where to search for Fuchsian invariant functions}
A priori, there are essentially two strategies that might lead to the desired result, and we just outline both of them, even though we are inclined towards the second, which looks much more promising. In this paper, we will not develop the necessary constructions and calculations, which require
a dedicated and extended work, postponed to a future publication. Before proceeding, we emphasize another conceptual point.
The solution to the posed problem is very important for mathematics, but is it necessary for our ML learning goals? Let us stress that our principle concern is \textit{a leap further in "machine learning"} where data of the discussed \textit{"image typology"} have still to be treated as \textit{points in a manifold} and not yet, as they should,  as \textit{points in the space of bundle sections}. Focusing on this task and considering image \ref{bolzanetoquadro}, we see that what we need is the space of integrable continuous functions on the interior of
a fundamental domain $\mathcal{F}_\Gamma$. Are these necessarily $\Gamma$-invariant functions? The question is the same as asking, with reference to
the universal approximator theorem (see \cite{TSnaviga}), if the integrable functions on the hypercube $\mathrm{HyC}\,\equiv\,[0,1]^n$ have
to be periodic on its boundary.
Certainly the periodic functions provide a functional basis for $L^2(\mathrm{HyC})$ but they are not necessary. Using the periodic functions is
equivalent to investigating the space of integrable functions on a multi-torus. Alternatively $\mathrm{HyC}$ can be looked at as a compact domain
with boundary and one can use whatever else different basis of functions that are bounded on $\mathrm{HyC}$, for example the polynomials. With
the same logic there is no need to utilize $\Gamma$-invariant functions in our case and can we utilize the regular harmonics described in previous
sections that are bounded in the fundamental domain. Hence the problem of the Laplacian spectrum on the Riemann surface is interesting per s\'e from
the mathematical point ov view, yet unessential for the ML goals.
\par
This being clarified we present for the solution of the Laplacian problem on compact Riemann surfaces
the two investigation directions that we were able to single out.
\subsubsection{Functional spaces \`a la Bargmann}
The first of the two possible lines of thought is more conservative one with respect to the existing approach outlined in sect.\ref{collinare}
and its subsections. One still tries to find a space of invariant functions against the action of the Fuchsian group $\Gamma$ starting from
functions that form a representation of the full isometry group. The only difference is that instead of imposing invariance with respect to
the $\Gamma$-generators $\mathcal{KS}_i$ within finite dimensional representations (the regular harmonics) or on integral kernels, one tries
to impose it on functions belonging to a UIR representation. Just to outline a possible scheme, we consider a functional space
of holomorphic functions $f(z)$  of the coordinate:
\begin{equation}\label{complessino}
  z \, \equiv \, x\, + \, \mathit{i} \, y
\end{equation}
having denoted by ${x,y}$ the cartesian coordinate of a point $p$ in the unit disk.
\par
To define the action of the isometry group on the holomorphic functions $f(z)$ it is convenient to start from the representation of its Lie algebra
$\so(1,2) \sim \slal(2,\mathbb{R})$  by means of holomorphic first order differential operators. The relation of this complex coordinate with
the solvable coordinates has been treated extensively in the foundational paper \cite{pgtstheory} and we do not dwell on it.
\par
Given a positive real number $\mathbb{R} \ni \mathit{j} >0$, inspired by \cite{barmanno}, we can write the following first order differential operators:
\begin{alignat}{5}\label{quaemarenavigerum}
  L_- & = z^2 \, \frac{d}{dz} - 2\mathit{j} \, z  \quad ; \quad &
  L_+ & = - \frac{d}{dz} \quad ;\quad &
  L_0 & = -z \, \frac{d}{dz} \, + \, \mathit{j}
\end{alignat}
closing the following Lie algebra commutation relations:
\begin{equation}\label{falena}
  \left[ L_0 \, , \, L_\pm \right] \, = \, \pm \, L_\pm \quad ; \quad \left[ L_+ \, , \, L_- \right] \, = \, 2 \, L_0
\end{equation}
which precisely corresponds to  the $\so(1,2) \sim \slal(2,\mathbb{R})\sim \su(1,1)$ Lie algebra. Furthermore we calculate the Casimir:
\begin{equation}\label{jaghellone}
  \mathfrak{Cas} \, = \, \ft 12 \left( L_{+} \, L_- \, +\, L_- \, L_+ \, + \, 2\, L_0 \right) \, = \,  \mathit{j} \, (1\, + \, \mathit{j})
\end{equation}
The operators (\ref{quaemarenavigerum}) correspond, at the finite level to the following realization of the action of any $\mathrm{SU(1,1)}$
group element on the space of holomorphic functions:
\begin{equation}\label{tournedo}
  \forall g\, = \, \left(
                     \begin{array}{cc}
                       \bar{\alpha} & \bar{\beta}\\
                       \beta  & \alpha \\
                     \end{array}
                   \right) \in \mathrm{SU(1,1)} \quad ; \quad T_{g}[f](z) \, = \, \left(\beta\, z \, + \, \alpha\right)^{\mathit{j}} \,
                   f\left(\frac{\bar{\alpha} \, z \, + \, \bar{\beta} }{\beta\, z \, + \, \alpha} \right)
\end{equation}
which ensures that the Hilbert space scalar product:
\begin{equation}\label{gilbertoscala}
  \langle f\, , \, h \rangle \, = \, \int_{\mathbb{H}^2}  \, \bar{f}(\bar{z}) \, h(z) \, \left(1- \tilde{z} \, z\right)^{-2-\mathit{j}} \, dz \, d\bar{z}
\end{equation}
 is invariant with respect to all transformations of the group $\mathrm{SU(1,1)}$.
 \par
 Considering the finite transformation (\ref{tournedo}) and the explicit form (after Cayley transformation to $\mathrm{SU(1,1)}$) of the Fuchsian
 group generators the question is whether one succeeds in finding for specific $\mathit{j}.s$ holomorphic square integrable functions that are invariant
 with respect to all generators of $\Gamma$. In such terms the problem turns out to be rather formidable and of very uncertain solution.
\subsubsection{Use of Riemann surface algebraic geometry tools}
\label{nuovastrategia}
\begin{figure}
\begin{center}
\vskip 1cm
\includegraphics[width=90mm]{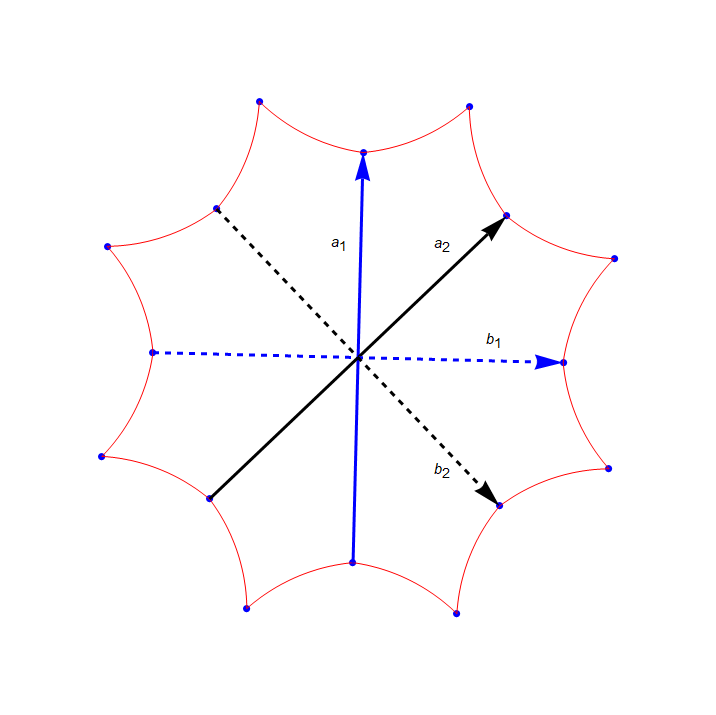}
\includegraphics[width=90mm]{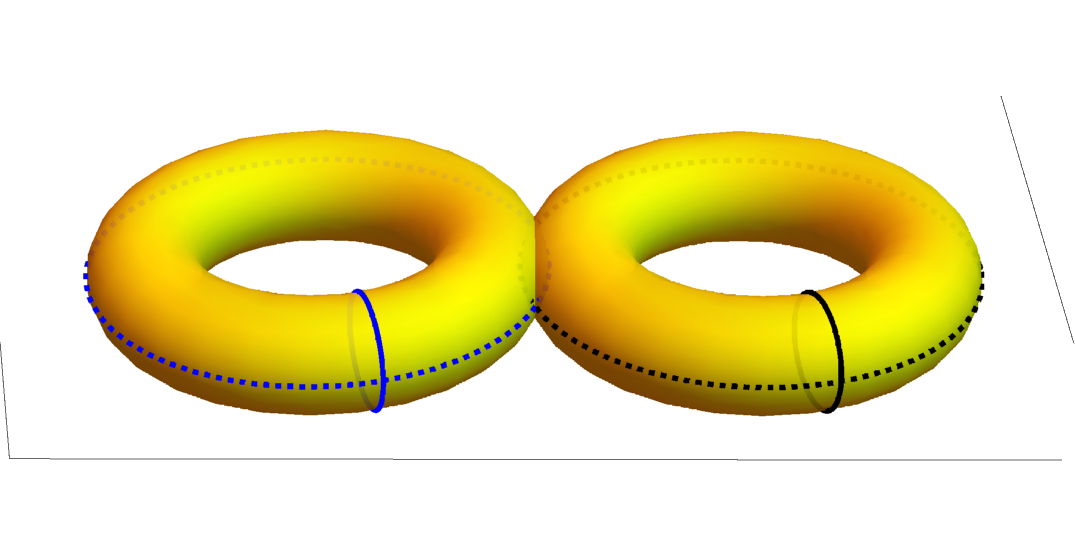}
\vskip 1cm
\caption{\label{omologiabolza} The standard homology cycles for the genus $2$ Bolza surface. The cycles $a_{1,2}$ are displayed in solid lines,
respectively blue and black, while the $b_{1,2}$ are displayed in dashed lines, also blue and black respectively. The cycles are shown both on
the surface of the bi-torus and in the Bolza octagonal fundamental domain. The four positive generators of the $\Gamma_8$ group are those that map
four consecutive edges of the 8 octagon edges in their respective opposite edge in this way realizing the folding of the fundamental domain
into the bi-torus. The arrowed straight lines in the above figure are closed geodesics since the starting and arrival point are identified. They
correspond to the four cycles $a_{1,2},b_{1,2}$.}
\end{center}
\end{figure}

The second line of thought relies on the matter of fact that tessellations of $\mathbb{H}^2$ and hence, via uniformization theorem,
fundamental domains, are essentially equivalent to compact orientable smooth Riemann surfaces of genus $g \geq 2$. Hence in terms of
Riemann surface coordinates the quotient with respect to the corresponding Fuchsian group $\Gamma$ is automatically done.
The appropriate instrument to be utilized in the context of our specific problem therefore seems to be the \textbf{Abel-Jacobi map} that injects
the genus $g$ Riemann surface $\Sigma_g$  in its Jacobian variety $J(\Sigma_g)$ the latter being defined as follows (see \cite{farkaskra}).
Let $\mathcal{C}_A \, = \, \{a_i,b_i\}$ ($i=1,\dots, g$) be a canonical basis of homology $1$-cycles with canonical intersection matrix:
\begin{equation}\label{gardellino}
 \mathcal{C}_A\bigcap\mathcal{C}_B \, \equiv \, \mathcal{I}_{A,B}\, = \, \left(
                             \begin{array}{c|c}
                               \mathbf{0}_{g\times g}  &\mathbf{1}_{g\times g} \\
                               \hline
                               -\mathbf{1}_{g\times g} & \mathbf{0}_{g\times g}  \\
                             \end{array}
                           \right)
 \end{equation}
 For the case of the Bolza surface the cycles are $4$ and they are displayed in fig.\ref{omologiabolza}.
Correspondingly, in cohomology one has $2g$ linear independent harmonic $1$-forms (closed and co-closed) that can be reorganized as the real
and imaginary parts of $g$ holomorphic one forms:
\begin{equation}\label{trallalla}
 \omega_{i}\, = \, \varpi_i(z)dz
\end{equation}
each representing a non-trivial cohomology class of the Dolbeault cohomology group $H^1_\partial(\Sigma_g)$. The basis (\ref{trallalla}) can be arranged
in such a way that:
\begin{equation}\label{perrone}
  \int_{a_j} \,\omega_{i} \, = \, \delta_{ij}
\end{equation}
Once the basis is fixed in the above way we have that the so named \textbf{period matrix}:
\begin{equation}\label{periodona}
  \Pi_{ij}(\Sigma_g) \, = \, \int_{b_j} \,\omega_{i}
\end{equation}
is a \textbf{constant  $g\times g$ complex matrix with positive definite imaginary part}
encoding fundamental geometric properties of the Riemann surface $\Sigma_g$. (see for
instance \cite{farkaskra} chapter III page 91 and following ones). In view of
eq.(\ref{perrone}) one introduces a lattice constructed as follows. Let $\boldsymbol{e}^{i}$
($i=1,\dots ,g$) be an orthonormal basis of $\mathbb{R}^g$ and consider the $g$
complex vectors:
\begin{equation}\label{krugnat}
  \boldsymbol{\pi}^i \, \equiv \, \{\Pi_{i1}, \, \Pi_{i2}, \, \dots, \, \Pi_{ig}\}
\end{equation}
the lattice $\Lambda(\Sigma_g)$, isomorphic to $\mathbb{Z}^g \times \mathbb{Z}^g$ is defined as follows:
\begin{equation}\label{craligione}
  \mathbb{C}^g \, \supset \, \Lambda(\Sigma_g) \, \equiv \, \left\{ \mathbf{w} \in \mathbb{C}^g \, | \, \mathbf{w} \, = \,\sum_{i=1}^g m_i \,
  \boldsymbol{e}^{i} \, + \, \sum_{k=1}^g n_k \, \boldsymbol{\pi}^k \quad ; \quad m_i,n_k \in \mathbb{Z }\right\}
\end{equation}
The Jacobian variety of the Riemann surface $\Sigma_g$ is defined as the quotient of $\mathbb{C}^g$ with respect to the lattice $\Lambda(\Sigma_g)$
\begin{equation}\label{kratillo}
  J(\Sigma_g) \, = \, \mathbb{C}^g / \Lambda(\Sigma_g)
\end{equation}
and by its very definition it is a complex $g$ torus. The \textbf{Abel-Jacobi map} is an injection map of the original Riemann surface $\Sigma_g$
into its own Jacobian variety:
\begin{equation}\label{cartolka}
  \mu_{AB} \, : \, \Sigma_g \, \longrightarrow \, J(\Sigma_g)
\end{equation}
Explicitly the map is realized as follows:
\begin{equation}\label{circabolla}
  \forall p\in \Sigma_g \quad : \quad  \mu_{AB}(p) \, = \, \left\{\int_{p_0}^{p} \omega_1 , \, \int_{p_0}^{p} \omega_2, \, \dots ,\,
   \int_{p_0}^{p} \omega_g\right\} \, \text{mode} \, \Lambda(\Sigma_g)
\end{equation}
where the integrals are performed along a path going from some reference point $p_0$ to
the point $p$ and, at first sight, it might seem that (\ref{circabolla}) is not a map
depending only on the arrival point $p$ but also on the chosen integration path. This is
not true because any change in path can be reabsorbed by a lattice addition. It follows
that the identification of Riemann surface points by means of the corresponding points in
the Jacobian variety is a sure way of performing the quotient with respect to the Fuchsian
group. Functions on the image of $\Sigma_g$ inside $J(\Sigma_g)$ are by definition,
functions defined on the fundamental Fuchsian domain. What we have done is trading the quotient with respect to the non-abelian Fuchsian group with the quotient with
respect to the lattice $\Lambda(\Sigma_g)$. The latter quotient has also to be
automatized in order to obtain coordinates that are completely Fuchsian invariant. This
can be done by means of the Siegel Theta Function.
\subsubsection{The Siegel Theta function}
Given the period matrix $\Pi$ one can construct the \textit{Siegel Theta function} that is a map:
\begin{equation}\label{pratisco}
  \Theta[\Pi] \, : \, \mathbb{C}^g \, \longrightarrow \, \mathbb{C}
\end{equation}
whose explicit expression is the following one:
\begin{equation}\label{ocsitarp}
  \Theta[\Pi \, | \mathbf{w}]\, = \, \sum_{\mathbf{n}
  \in \mathbb{Z}^g} \,\exp\left[2\,\pi \,\mathit{i} \left( \ft 12 \,\mathbf{n}\cdot \Pi \cdot \mathbf{n} \, + \, \mathbf{n}\cdot \mathbf{w}\right)
  \right]
\end{equation}
The function (\ref{ocsitarp}) is not exactly periodic on the lattice $\Lambda(\Sigma_g)$ defined in eq.(\ref{craligione}) but almost periodic indeed
under the addition to its argument $\mathbf{w}$ of a lattice element:
\begin{equation}\label{trasformazia}
  \Theta[\Pi \, | \mathbf{w} + \mathbf{m} + \mathbf{k}\cdot\Pi] \, = \, \exp\left[2\, \pi \,\mathit{i}\left(\mathbf{k}\cdot \Pi \cdot \mathbf{k}
  + \ft 12 \, \mathbf{w}\cdot \Pi \cdot \mathbf{k}\right) \right] \times \Theta[\Pi \, | \mathbf{w}] \quad \quad \text{if} \quad \quad
  \mathbf{m},\mathbf{k} \in
  \mathbb{Z}^g
\end{equation}
Hence, the Siegel Theta function is not exactly a function on the Jacobian variety $J(\Sigma_g)$ but rather a section of a $\mathrm{U(1)}$-bundle over it,
since the transformation (\ref{trasformazia}) is only due to a phase-factor. Hence, if we finally consider:
\begin{equation}\label{thetacappello}
  \widehat{\Theta}[\Pi \, |\, p] \, \equiv \, \Theta\left[\Pi \, | \mu_{AB}(p) \right]
\end{equation}
where $\mu_{AB}(p)$ is the Abel-Jacobi map (\ref{circabolla}) we have a \textbf{section of a $\mathrm{U(1)}$ bundle over $\Sigma_g$}.
A fundamental fact is that the holomorphic differentials and the period matrix transform in a precise way under the finite automorphism group
$\mathrm{GD_{index}}$ (see \cite{farkaskra} page 289 and following ones). Indeed, necessarily, there is a homomorphism:
\begin{equation}\label{cancellatone}
 h \, : \,  \mathrm{GD_{index}} \, \longrightarrow \,\mathrm{ Sp(2g\, , \, \mathbb{Z})}
\end{equation}
and the symplectic images of the automorphism group elements act crystallographically on the homology cycle basis, determining the
transformation of the period matrix and of Jacobian variety coordinates.
\par
It follows from this short summary that the eigenfunction of the Laplacian, with precise transformation properties under the automorphism group
$\mathrm{GD_{index}}$, have to be properly constructed in terms of the Abel-Jacobi map and of the Siegel theta functions.
\par
Such a job is left for a future publication, yet it is appropriate to stress once more, that its result is extremely important not only
for pure mathematics but also in order to implement
the \textit{leap forward in machine learning algorithms} able to codify a \textit{"discretized image"} as a section of a fiber bundle
and not simply as a point in a manifold. Mapping the image to a Riemann surface of genus $g=1$, namely to a torus, preserves the metric relations
of the grid, yet loses the non-linear power of hyperbolic manifolds and the much more sophisticated structure of the harmonic basis functions that
can enlighten hidden classifying structures in data analysis.
\section{The construction of codimension \texorpdfstring{$1$}{1} separators in non-compact symmetric spaces \texorpdfstring{$\mathcal{M}^{[r,q]} $}{Mrq}}
\label{separatini}
In this section we discuss the construction of a hypersurface $\mathfrak{S}=\mathcal{H}^{[r,q]}$ inside the hyperbolic symmetric spaces of the Tits Satake  Universality class (\ref{Mrq}) that satisfies all the properties mentioned in the definition 6.6 of \cite{TSnaviga} to function as a separator.  To facilitate our reader we repeat here the above-mentioned definition:
\begin{definizione}
\label{separatoni}
Let $\mathcal{M}=\mathrm{U/H}$ be any non-compact symmetric space. A separator for $\mathcal{M}$ is a \textit{codimension
one} submanifold $\mathfrak{S} \subset \mathcal{M}$, with the following properties:
\begin{description}
  \item[a)] $\mathfrak{S}$ is a homogeneous space.
  \item[b)] $\mathfrak{S}$ intersects the boundary at infinity of $\mathcal{M}$ : $\partial \mathcal{M}\cap \mathfrak{S} \neq \emptyset$
  \item[c)] $\mathfrak{S}$ induces a binary partition of $\mathcal{M}$ defined as follows.
  There exists two submanifolds $\mathcal{M}_\pm \subset \mathcal{M}$ such that
  \begin{enumerate}
    \item $\mathcal{M} \, = \, \mathcal{M}_+\bigcup\mathcal{M}_-$
    \item $\text{dim}\mathcal{M}_\pm  \, = \, \text{dim} \mathcal{M}$
    \item $\mathcal{M}_+\bigcap\mathcal{M}_- \, = \, \mathfrak{S} $
  \end{enumerate}
\end{description}
\end{definizione}
With reference to the above, we point out that, being a hypersurface, $\mathcal{H}^{[r,q]}\subset \mathcal{M}^{[r,q]} $ has by definition codimension 1; furthermore, as we show below it is a homogeneous submanifold by construction and since it includes all the Cartan coordinates it extends to infinity. The crucial property \textbf{c)} will be demonstrated below with care.
\par
This is the announced result that allows the extension to any value of the non-compact rank $r$ of the classification algorithms elaborated in \cite{TSnaviga,naviga} for the case $r=1$ of the hyperbolic planes. Indeed the surface $\mathcal{H}^{[r,q]}$ to be constructed below, is just the standard separator $\mathfrak{S}_{stand}$, \textit{i.e.} the analogue of the symmetric space $\mathcal{M}^{[1,q-1]}\subset \mathcal{M}^{[1,q]}$ used in \cite{TSnaviga,naviga}; similarly to the $r=1$ case all the members of the conjugacy class of $\mathcal{H}^{[r,q]}$ under the action of $\mathrm{SO(r,r+q)}$ can be used as
separators $\mathfrak{S}$ in the multi-class separation algorithm described in \cite{TSnaviga}. The structural difference between the $r=1$ case and all the $r\geq 2$ cases is that, for $r=1$, the standard separator is not only a homogeneous space but it is also symmetric, being totally geodesic\footnote{ If we define as \emph{index} of a symmetric space $\mathcal{M}$ the smallest possible codimension of a proper, totally geodesic submanifold of $\mathcal{M}$, by Theorem 11.1.6 (Iwahori) of \cite{Olmo} the index of an irreducible, Riemaniann symmetric space $\mathcal{M}$ is equal to 1 (i.e. $\mathcal{M}$ admits a totally geodesic hypersurface), only if $\mathcal{M}$ is isometric to a sphere $S^n$, to a real projective space $\mathbb{C}P^n$ or to a real hyperbolic space $\mathbb{H}P^n$, for $n\ge 2$. As we are considering non-compact symmetric manifolds of negative curvature, the above theorem implies that only the $r=1$ spaces $\mathcal{M}^{[1,q]}\sim \mathbb{H}P^{q+1} $ admit a totally symmetric hypersurface ($\mathcal{H}^{[1,q]}\equiv\mathcal{M}^{[1,q-1]}$).}. Apart from the loss of such an exceptional property, not necessary for the statistical classification algorithm, all the rest works in the same way as in \cite{TSnaviga} by utilizing the conjugacy class of the standard separator $\mathfrak{S}_{stand}\, \equiv\, \mathcal{H}^{[r,q]}$. Let us then proceed with the construction of the hypersurface $\mathcal{H}^{[r,q]}$.
\par
We begin by  introducing the following notation for the
generators of the solvable Lie algebra $Solv$ in $\mathfrak{so}(r,r+q)$:
\begin{align} & T^A \, \equiv \, \left\{ H_i=H_{\boldsymbol{\epsilon}_i},\,E_{\boldsymbol{\epsilon}_i\pm
\boldsymbol{\epsilon}_j},\,E_{\boldsymbol{\epsilon}_i^I}\right\}\,\,;\,\,\,\, i=1,\dots, r\,\,,\,\,\,\,I=1,\dots q\,.
\end{align}
where $\boldsymbol{\epsilon}_i$ is an orthonormal basis in $\mathbb{R}^r$ and $\boldsymbol{\epsilon}_i\pm \boldsymbol{\epsilon}_j$  are the long positive roots $\boldsymbol{\alpha}$, that are singlets under the Paint Group (see \cite{pgtstheory}\footnote{Let us stress that  \cite{pgtstheory}, which is partly a review, partly an original systematic mathematical paper, has been purposely written to
expose the Paint Group Tits Satake theory of non-compact symmetric spaces as a basis for the development a long term deep learning project of which the present paper just encodes the third step.  The motivations and principles of such long term project are mentioned in the introduction \ref{introibo}. As for the PGTS theory, it is a chapter of classical mathematics, in particular of Lie Group theory, yet it is not wildly known in the mathematical reference communities of algebraic geometers, differential geometers and algebrists, since it is one of the several original new branches of mathematical lore that were developed in the context of Supergravity/Supersymmetry/Brane theories. The reasons while
the Superworld was so much fertile and original in its mathematical fruits are analysed and discussed in \cite{pgtstheory}. These remarks are written for the benefit of the ML and Math readers of the present paper. Differently from what typically happens in ML literature, the type of mathematics invoked as an essential structural item by our ML constructions is not available in standard math textbooks, and it is much better known to theoretical physicists familiar with supersymmetry than to pure mathematicians.  }), while $\boldsymbol{\epsilon}_i^I$ are
the short positive roots, to be generically denoted by $\boldsymbol{\sigma}^I$, the index $I$, spanning the
fundamental representation of the paint group $\mathrm{G_{paint}} \, = \, {\rm SO}(q)$, labels their multiplicity.
These generators are normalised in such a way as to satisfy the following conditions:
\begin{align}
[H_i,\,E_{\boldsymbol{\alpha}}]&= \alpha(H_i)\,E_{\boldsymbol{\alpha}}\,,\,\,\,[H_i,\,E_{\boldsymbol{\sigma}^I}]
=\boldsymbol{\sigma}^I(H_i)\,E_{\boldsymbol{\sigma}^I}\,,\nonumber\\
[E_{\boldsymbol{\alpha}},\,E_{-\boldsymbol{\alpha}}]&=4\,\boldsymbol{\alpha}(H_i)\,H_i\,,\,\,\,[E_{\boldsymbol{\sigma}^I},\,E_{-\boldsymbol{\sigma}^I}]
=2\,\boldsymbol{\sigma}^I(H_i)\,H_i\,,\nonumber\\
[E_{\boldsymbol{\epsilon}_i^I},\,E_{\boldsymbol{\epsilon}_j^J}]&=
-\delta_{IJ}\,E_{\boldsymbol{\epsilon}_i+\boldsymbol{\epsilon}_j}\,,\,\,\,[E_{\boldsymbol{\epsilon}_i^I},
\,E_{-\boldsymbol{\epsilon}_j^J}]=-\delta_{IJ}\,E_{\boldsymbol{\epsilon}_i-\boldsymbol{\epsilon}_j}\,,
\end{align}
where we take $E_{-\boldsymbol{\alpha}}=E_{\boldsymbol{\alpha}}^T$ and  $E_{-\boldsymbol{\sigma}^I}=E_{\boldsymbol{\sigma}^I}^T$.
\par
Within $\mathcal{M}^{[1,q]}$ a symmetric hypersurface $\mathcal{H}^{[1,q]}$  of dimension $q$ can be constructed
and, as we mentioned above, it is the symmetric space $\mathcal{M}^{[1,q-1]}$:
\begin{equation}\label{juancamomillo}
  \mathcal{H}^{[1,q]}=\mathcal{M}^{[1,q-1]}\subset \mathcal{M}^{[1,q]}\,.
\end{equation}
This hypersurface is defined by a \emph{foliation} of $\mathcal{M}^{[1,q]}$ in which $\mathcal{H}^{[1,q]}$ is a leaf, defined by the constant value of a suitable coordinate. We shall introduce two sets of Gaussian coordinates: \emph{skew Gaussian} coordinates $\{\hat{w}_m,\,y\}$ and \emph{normal  Gaussian} coordinates $\{\tilde{w}_m,\,\tilde{y}\}$ \cite{Synge:1960ueh}, where $\hat{w}=(\hat{w}_m)$ and $\tilde{w}=(\tilde{w}_m)$, in the two coordinate systems, span the hypersurface which is defined by a constant value of $y$ or of $\tilde{y}$. This description of the manifold $\mathcal{M}$ as foliated in the hypersurfaces $\mathcal{H}$ and described by the two systems of Gaussian coordinates,  also applies to the more general case $r>1$.

The coordinate $y$ in the \emph{skew Gaussian} system can be chosen to be the parameter of the nilpotent generator $E_{\boldsymbol{\epsilon}_1^q}$
while the isometry algebra of the hypersurface is:
\begin{equation}
\mathfrak{so}(1,q)={\rm Span}\left(H_1,\,E_{\pm\boldsymbol{\epsilon}_i^I}\right)_{I=1,\dots, q-1}\,.
\end{equation}
The coset representative of $\mathcal{M}^{1,q}$ can then be written in the following form:
\begin{equation}
\mathbb{L}(\hat{w},\,y)=\mathbb{L}(\hat{w})\cdot e^{y\,E_{\boldsymbol{\epsilon}_1^q}}\,.
\end{equation}
In this coordinate system, the metric has the general form:
$$ds^2=(dy+{\bf w})^2+ ds_\mathcal{H}(y,\hat{w})\,,$$
where ${\bf w}={\bf w}(\hat{w},y)=y\,e^{-\varphi_1}\,d\varphi_1$ is a 1-form on the hypersurface and $\varphi_1$ is the parameter of $H_1$.
The hypersurface is located at $y=0$.
Alternatively, we introduce the normal Gaussian coordinates by defining the coset representative as follows:
\begin{equation}
\mathbb{L}(\tilde{w},\,\tilde{y})=\mathbb{L}(\tilde{w})\cdot e^{\tilde{y}\,K_y}\,,
\end{equation}
where $K_y\equiv -(E_{\boldsymbol{\epsilon}_1^q}+(E_{\boldsymbol{\epsilon}_1^q})^T)/2$.
One can prove that the metric has the general form:
$$ds^2=d\tilde{y}^2+ d\tilde{s}_\mathcal{H}(\tilde{y},\tilde{w})\,,$$
so that these coordinates are indeed normal.
The following relation holds between $y$ and $\tilde{y}$:
\begin{equation}y=\sinh(\tilde{y})\,,\label{yyt}\end{equation}
while $\hat{w}^m=\hat{w}^m(\tilde{w},\,\tilde{y})$.
We wish to emphasise that \emph{both Gaussian coordinate systems are global on} $\mathcal{M}^{[r,q]}$. \par
Let us now recall, for the reader's convenience, the definition 6.7 of \cite{TSnaviga}, of the distance of a point of a manifold from a hypersurface:
\begin{definizione}
\label{oriadistanza} Let $\mathcal{M}$ be a non-compact symmetric space and $\mathfrak{S}\subset \mathcal{M}$ a parameterized separator. We define the  \textit{unoriented distance} of any point $\mathbf{p}\in \mathcal{M}$ from the separator as:
\begin{equation}\label{curtalengo}
 \delta\left(\mathbf{p},\mathfrak{S} \right) \, \equiv \, \underbrace{\text{\rm inf}}_{\mathbf{u} \in \mathfrak{S}} \, \mathrm{d}\left(\mathbf{p},\mathbf{u}\right)
\end{equation}
where $\mathrm{d}( \, , \, )$ is the distance function in $\mathcal{M}$. Given a point $\mathbf{p}$ in $\mathcal{M}$ of normal  Gaussian coordinates $(\tilde{w}^m,\,\tilde{y})$, a generic point ${\bf u}$ on the hypersurface will have coordinates $(\tilde{w}^{\prime m},0)$. One can verify that $d({\bf p},{\bf u})$ is minimum when ${\bf u}={\bf u}_0\equiv (\tilde{w}^{m},0)$, so that:
$$\delta\left(\mathbf{p},\mathfrak{S} \right)=\mathrm{d}\left(\mathbf{p},\mathbf{u}_0\right)=
\frac{ |\tilde{y}|}{\sqrt{2}}\left({\rm Tr}\left(K_y^2\right)\right)^{\frac{1}{2}}=|\tilde{y}|\,.$$
The above distance is measured by the length of the geodesic arc connecting ${\bf p}$ with ${\bf u}_0$.
 Given such a geodesic curve, which in a non-compact symmetric space is unique, one can consider its tangent vector $\mathfrak{t}_0$  in $\mathbf{u}_0$ and evaluate the projection of the latter onto the normal bundle to the separator in the same point. Since the separator has codimension one we can always split the tangent bundle $T\mathcal{M} \, =\, T\mathfrak{S}  \oplus T\mathcal{N}$ and with respect to any standard frame of the normal bundle the geodesic tangent vector can have a positive or negative projection. This sign converts the unoriented distance $\delta\left(\mathbf{p},\mathfrak{S} \right) $ into the \textbf{oriented} one that we denote with a hat:
\begin{equation}\label{corretopino}
  \widehat{\delta}\left(\mathbf{p},\mathfrak{S} \right) \, = \, \underbrace{\pm}_{\text{if } \mathfrak{t}_0\cdot T\mathcal{N} = \pm}   \, \delta\left(\mathbf{p},\mathfrak{S} \right)
\end{equation}
\end{definizione}
Indeed, the point $\mathbf{u}_0=(\tilde{w}^m,\,0)$ defined above is just the one that minimises the distance in eq. (\ref{curtalengo}) as it can be verified using the general formulae of the geodesic distance discussed in \cite{pgtstheory} (see Sects. 4.2, 4.3, 4.4 of such paper).
Since the normal bundle is spanned by the vector field $\partial/\partial\tilde{y}$,
one can easily define the orientation of the normal bundle
as the sign of the coordinate $\tilde{y}$, and this leads to setting:
$$\widehat{\delta}\left(\mathbf{p},\mathfrak{S} \right) \, = \,\tilde{y}={\rm arcsinh}(y)\,.$$
Note, however, that $\tilde{y}$, as opposed to $y$, is not a solvable coordinate since it does not parametrise a generator of the solvable Lie algebra associated with the isometry group of $\mathcal{M}$.
\par
The above construction of the hypersurface $\mathcal{H}^{[r,q]}$, for $r=1$, straightforwardly extends to the case $r>1$. We define the hypersurface $\mathcal{H}^{[r,q]}$ for $r>1$ by induction. We start from $r=2$ and consider the following isometric representation of $\mathcal{M}^{[2,q]}$:
\begin{equation}
\mathcal{M}^{[2,q]}=\frac{{\rm SO}(2,2+q)}{{\rm SO}(2)\times {\rm SO}(2+q)}\sim \left[{\rm SO}(1,1)\times \frac{{\rm SO}(1,1+q)}{{\rm SO}(1+q)}\right]\ltimes \exp({\bf N}_{({\bf q+2})_{+1}})\,,
\end{equation}
where ${\bf N}_{({\bf q+2})_{+1}}$ is the Abelian subspace of $\mathfrak{so}(2,2+q)$ with positive grading with respect to the adjoint action of ${\rm SO}(1,1)$ and transforming in the fundamental representation of ${\rm SO}(1,1+q)$, with respect to the adjoint action of the latter. We define the hypersurface $\mathcal{H}^{[2,q]}$ as:
\begin{equation}\label{coccinglia}
\mathcal{H}^{[2,q]}\equiv \left[{\rm SO}(1,1)\times \mathcal{H}^{[1,q]}\right]\ltimes \exp({\bf N}_{({\bf q+2})_{+1}})\,
\end{equation}
It is obtained by acting on the symmetric space $ \mathcal{H}^{[1,q]}$ defined above, for fixed $y$, by means of the solvable group generated by the following algebra:
\begin{equation}\label{cardiodattero}
  \mathfrak{so}(1,1)\oplus {\bf N}_{({\bf q+2})_{+1}}\,
\end{equation}
which, together with $\mathfrak{so}(1,q)$, close the non-semisimple Lie algebra. In particular, we define the following solvable Lie algebra associated with the hypersurface:
\begin{equation}
Solv[\mathcal{H}^{[2,q]}]=\mathfrak{so}(1,1)\oplus Solv[\mathfrak{so}(1,q)]\oplus {\bf N}_{({\bf q+2})_{+1}}={\rm Span}(H_i,\,E_{\boldsymbol{\epsilon}_1+\boldsymbol{\epsilon}_2},\,
E_{\boldsymbol{\epsilon}_1-\boldsymbol{\epsilon}_2},\,E_{ \boldsymbol{\epsilon}_1^I},E_{ \boldsymbol{\epsilon}_2^a})\,,\,\,\,a=1,\dots, q-1\,,
\end{equation}
where:
\begin{equation}\label{grillo2}\mathfrak{so}(1,1)={\rm Span}(H_1)\,,\,\,Solv[\mathfrak{so}(1,q)]={\rm Span}(H_2,\,E_{ \boldsymbol{\epsilon}_2^a})\,,\, {\bf N}_{({\bf q+2})_{+1}}={\rm Span}(E_{\boldsymbol{\epsilon}_1+\boldsymbol{\epsilon}_2},\,
E_{\boldsymbol{\epsilon}_1-\boldsymbol{\epsilon}_2},\,E_{ \boldsymbol{\epsilon}_1^I})\,
\end{equation}
This solvable Lie algebra is obtained by subtracting from the solvable Lie algebra of $\mathfrak{so}(2,2+q)$ the only nilpotent generator $E_y\equiv E_{ \boldsymbol{\epsilon}_2^q}$, to be parametrized by $y$.
Now, by induction, we define the hypersurface for $r+1$ in terms of the one defined for $r$:
\begin{equation}
\mathcal{H}^{r+1,q}\equiv \left[{\rm SO}(1,1)\times \mathcal{H}^{[r,q]}\right]\ltimes \exp({\bf N}_{({\bf q+2r})_{+1}})\,.
\label{inclusionchain}
\end{equation}
The hypersurface $\mathcal{H}^{[r,q]}$ is spanned by the transitive action of a solvable group of isometries generated by
\begin{equation}\label{grillo3}
Solv[\mathcal{H}^{[r,q]}]={\rm Span}(H_i,\,
E_{\boldsymbol{\epsilon}_i\pm\boldsymbol{\epsilon}_j},\,E_{ \boldsymbol{\epsilon}_u^I},\,E_{ \boldsymbol{\epsilon}_r^a})\,,\,\,\,a=1,\dots, q-1,\,\,u=1,\dots, r-1\,,
\end{equation}
This solvable Lie algebra is obtained from that generating $Solv[\mathcal{M}^{[r,q]}]$, by subtracting the nilpotent generator
$E_y\equiv E_{ \boldsymbol{\epsilon}_r^q}$.
Since the first term in the inclusion chain
\begin{equation}
\mathcal{H}^{[1,q]}\,\hookrightarrow \,\mathcal{H}^{[2,q]}\,\hookrightarrow \,\dots\,\hookrightarrow \,\mathcal{H}^{[r,q]}\,,
\end{equation}
 implied by eq. (\ref{inclusionchain}), is $\mathcal{H}^{[1,q]}$ whose  solvable Lie algebra has  automorphism group ${\rm SO}(q-1)$, which is the \emph{subpaint} group of $\mathcal{M}^{[r,q]}$, this group will also be the automorphism group of $Solv[\mathcal{H}^{[r,q]}]$ and thus part of its isotropy group.
 \par
The above construction suggests a description of $\mathcal{M}^{[r,q]}$
as a foliation in the hypersurface $\mathcal{H}^{[r,q]}$, which is associated with the definition of Gaussian coordiantes on $\mathcal{M}^{[r,q]}$. Just as in the $r=1$ case, we define skew and normal Gaussian coordiantes, $(\hat{w},\, y)$ and $(\tilde{w},\, \tilde{y})$, which, as in the $r=1$ case, are global on $\mathcal{M}^{[r,q]}$, and correspond to the following definitions of the coset representative:
\begin{equation} \label{binocchio1}
\mathbb{L}_s(\hat{w},\,y)\equiv \mathbb{L}_{\mathcal{H}}(\hat{w})\cdot e^{y\,E_y}\,\,,\,\,\,\,\mathbb{L}(\tilde{w},\,\tilde{y})\equiv\mathbb{L}_{\mathcal{H}}(\tilde{w})\cdot e^{\tilde{y}\,K_y}\,\,\,,\,\,\,\,\mathbb{L}_{\mathcal{H}}(\hat{w})\,\in\, \exp\left(Solv[\mathcal{H}^{[r,q]}]\right)\,.
\end{equation}
where $K_y$ was defined earlier and the relation between $y$ and $\tilde{y}$ is still given by eq. \eqref{yyt}.
As in the $r=1$ case, the distance of a point $\mathbf{p}$ of coordinates $(\tilde{w}^m,\,\tilde{y})$ from the hyperplane is the geodesic distance of $\mathbf{p}$ from the point $\mathbf{u}_0$ of coordinates $(\tilde{w}^m,\,0)$.
Applying the general formula for the distance we still find:
\begin{eqnarray}\label{corrilepre2}
\delta (\mathbf{p},\mathcal{H})&\equiv & \mathrm{d}(\mathbf{p},\mathbf{u}_0)=|\tilde{y}|=|{\rm arcsinh}(y)|\,.
\end{eqnarray}
The \emph{oriented distance} has the same definition as for the $r=1$ case:
$$\widehat{\delta}\left(\mathbf{p},\mathfrak{S} \right) \, = \,\tilde{y}={\rm arcsinh}(y)\,.$$
Eq. (\ref{corrilepre2}) generalizes to all manifolds $\mathcal{M}^{[r,q]}$ the result (6.42-6.43) of \cite{TSnaviga} and it allows to utilize in all cases the analogue $\hat{\sigma}(x)\, \equiv \, \sigma(\sinh (x))$ of the logistic sigmoid as in eq.(6.44) of \cite{TSnaviga}. In other words, we use the coordinate $y$ as an argument of the standard sigmoid or the softmax exponentials.
\begin{figure}
    \centering
    \includegraphics[width=0.50\linewidth]{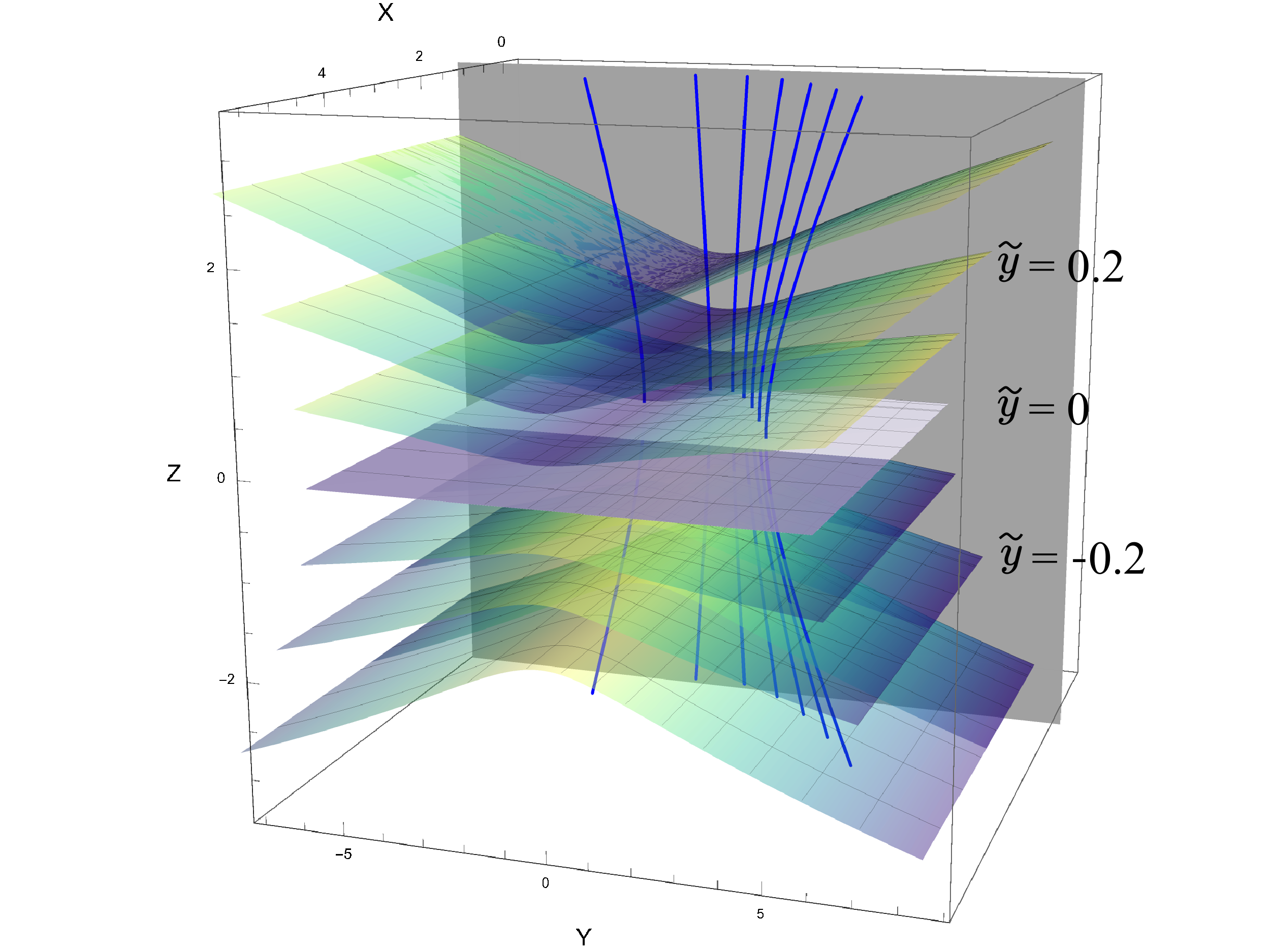} \quad \includegraphics[width=0.45\linewidth]{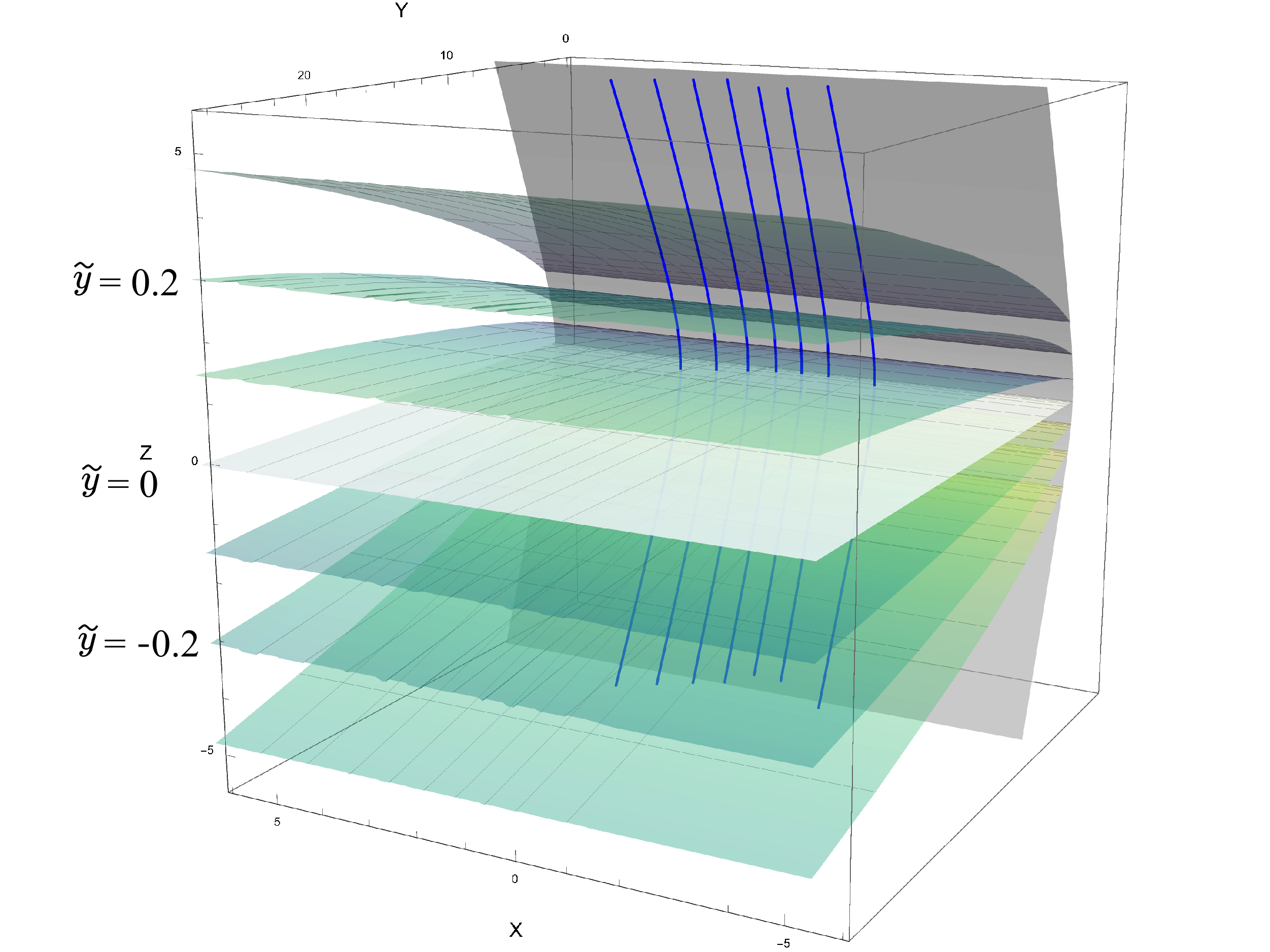}
    \caption{Parametric plot of different slides of $\mathcal{H}^{[2,1]} \subset \mathcal{M}^{[2,1]}$ in the normal Gaussian coordinates embedded in a non-isometric way in $\mathbb{R}^3$. The different surfaces are labelled by different values of $\tilde y$ and parameterised by the coordinates along the two Cartan generators. The blue lines correspond to geodesics whose tangent vector is $\partial/\partial \tilde y$.
    Left: The coordinates $X,Y,Z$ are identified with the components $M_{1 1}, M_{2 4}, M_{2 5}$ and the gray plane corresponds to the bounce surface located where the coordinate along the first Cartan vanishes.
    Right: The coordinates $X,Y,Z$ are identified with the components $M_{1 3}, M_{2 2}, M_{2 5}$, and the gray surface corresponds to the bounce surface due to the projection located where the coordinate along the second Cartan goes to 0. In both cases, $M_{ij}$ corresponds to the components of $\mathbb{L}\mathbb{L}^T$.}
    \label{fig:enter-label}
\end{figure}
\paragraph{The $\mathcal{M}_n={\rm SL}(n,\mathbb{R})/{\rm SO}(n)$ model.} It is now interesting to see that we can further generalize the formulation of the separators to the most general case where the ambient space is the complete non-compact maximally split symmetric space ${\rm SL}(n,\mathbb{R})/{\rm SO}(n)$. Indeed, we can use recursively the following isometric description of that manifold:
\begin{equation}
   \frac{ {\rm SL}(n,\mathbb{R})}{{\rm SO}(n)}=\left[{\rm SO}(1,1)\times \frac{{\rm SL}(n-1,\mathbb{R})}{{\rm SO}(n-1)}\right]\ltimes \exp\left({\bf N}_{({\bf n-1})_{+1}}\right)\,.
\end{equation}
in order to express $\mathcal{M}_n$ in terms of
\begin{equation}\label{erbacipollina1}
\mathcal{M}_2\sim\mathcal{M}^{[1,1]}=\frac{{\rm SO}(1,2)}{{\rm SO}(2)}\,,
\end{equation}
of which we have defined the hypersurface $\mathcal{H}^{1,1}$ following the above construction. We then define the hypersurface $\mathcal{H}_n\subset \mathcal{M}_n$ by induction through the recursive formula:
\begin{equation}\label{erbacipollina2}
\mathcal{H}_n=\left[{\rm SO}(1,1)\times \mathcal{H}_{n-1}\right]\ltimes \exp\left({\bf N}_{({\bf n-1})_{+1}}\right)\,.
\end{equation}
Let us denote by $\pm (\boldsymbol{\epsilon}_i-\boldsymbol{\epsilon}_j) $, $i<j=1,\dots, n$, the roots of $\mathfrak{sl}(n,\mathbb{R})$, so that the isometry solvable Lie algebra of $\mathcal{M}_n$ reads:
\begin{equation}\label{erbacipollina3}
Solv[\mathcal{M}_n]={\rm Span}\left(H_i,\,E_{\boldsymbol{\epsilon}_i-\boldsymbol{\epsilon}_j}\right)\,\,,\,\,\,(i<j=1,\dots, n)\,.
\end{equation}
The solvable isometry algebra of $\mathcal{H}_n$ is then constructed as follows:
\begin{equation}\label{erbacipollina4}
Solv[\mathcal{H}_n]=Solv[\mathcal{M}_n]\ominus {\rm Span}(E_{\boldsymbol{\epsilon}_{n-1}-\boldsymbol{\epsilon}_n})\,.
\end{equation}
where now we identify $E_y$ with $E_{\boldsymbol{\epsilon}_{n-1}-\boldsymbol{\epsilon}_n}$.
We still define the coset representative in the solvable Lie group of isometries isometric to $\mathcal{M}_n$, as:
\begin{equation}   \mathbb{L}_s(\hat{w},\,y)=\mathbb{L}_{\mathcal{H}}(\hat{w})\cdot e^{y\,E_y}\,\,,\,\,\,\,\mathbb{L}_{\mathcal{H}}(\hat{w})\,\in\, \exp\left(Solv[\mathcal{H}_n]\right)\,.
\end{equation}
This, just as in the case of $\mathcal{M}^{[r,q]}$, defines the skew Gaussian coordinates relative to the hypersurface $\mathcal{H}_n$.
The matrix representing $E_y$ in this case has all zero entries, except for a 1 in the position $(n-1,n)$.
We also define the normal  Gaussian coordinates $\tilde{w}^m,\,\tilde{y}$ by the coset representative:
\begin{equation}   \mathbb{L}(\tilde{w},\,\tilde{y})=\mathbb{L}_{\mathcal{H}}(\tilde{w})\cdot e^{\tilde{y}\,K_y}\,.
\end{equation}
\par
Computing the oriented distance, we find once again:
\begin{equation}\label{erbacipollina}
\widehat{\delta}\left(p,\mathcal{H}\right)\equiv \tilde{y}= {\rm arcsinh}(y)\,.
\end{equation}

\section{Conclusions}
In this paper, we have addressed a certain number of mathematical problems whose concrete and computational solution is propaedeutical to the further development of Cartan neural networks \cite{pgtstheory,naviga,TSnaviga}. In particular, we have obtained the following results:
\begin{enumerate}
  \item We have determined the algebraic construction algorithm of codimension 1 separators for all non-compact symmetric spaces, which is the necessary prerequisite in order to extend to all such $\mathrm{U/H}$ manifolds the geometric softmax algorithm for the classification task as described in \cite{naviga, TSnaviga}.
  \item We have introduced the general conception of Tits Satake vector bundles that can provide the basis for Cartan Convolutional
  neural networks.
  \item We have explicitly constructed in closed form the $\Delta^+_{8,3,2}$ tiling group of the Poincaré plane, and we have determined, also in closed form, the generators of two normal Fuchsian subgroups of finite index, respectively named $\Gamma_{16}$ and $\Gamma_8$, the first quotient group being a solvable group $\mathrm{GD_{96}}$ of order $96$, the second a solvable group $\mathrm{GD_{48}}$ of order $48$ which happens to be a normal subgroup of the former. A posteriori, we have identified the first
  quotient as the uniformization of a specific genus $g=3$ Riemann surface, namely the Fermat Quartic, the second as the uniformization of the genus $g=2$ Bolza surface.
  For the case of the Bolza surface, we have explicitly constructed the fundamental
  domain determining in terms of radicals the coordinates of all the vertices of its
  boundary. Such mathematical structures are not new in the literature, but what is new is the completely explicit and fully detailed form of their construction,  which is a prerequisite for their prospective use in any kind of machine learning algorithm.
  \item As far as regular harmonics are concerned, we have revisited their construction in terms of the spinor representation rather than in terms of the vector one.
  \item We have revisited with an in-depth analysis the methods to derive the Green Function of the Laplacian for all
  the members of the $r=1$ Tits Satake universality class, namely for all $\mathrm{SO(1,2+k)/SO(2+k)}$. We have obtained
  a general closed-form expression in terms of Hypergeometric Functions. From such an expression of the Green function, we have derived the construction of the Heat kernel as an integral transform over the Laplacian eigenvalue.
  \item We have determined the maps from the plane to the hyperbolic plane of compact regions.
  \item We have critically analyzed the existing strategies for the derivation of the
  Laplacian spectrum on uniformized Riemann surfaces, claiming that such methods are
  not able to provide the explicit form of the Laplacian eigenfunctions, that are not indispensable for
  machine learning algorithms but costitute a very interesting open problem for pure mathematics.
  Reviewing the alternatives we have proposed a different
  approach based on the Abel-Jacobi map of the Riemann surface to its own Jacobi variety
  and on the use of Siegel Theta functions.
\end{enumerate}
Summarizing, the present paper, which is mathematical in its format and conception,
should be considered as the natural continuation of papers \cite{pgtstheory, TSnaviga} that have laid down the theoretical basis for new geometrical neural networks dubbed Cartan neural networks, already tested with the explicit implementations of the case
$r=1$ presented in \cite{naviga}. The \textbf{long-term goal} pursued in the Cartan-PGTS  programme is the final \textbf{full removal} of \textbf{point-wise activation functions} that, on one hand, provided the spectacular success of classical neural networks,
as \textbf{universal approximators} of functions with a bounded compact support,
on the other hand are the \textbf{main intrinsic obstacle} to a \textbf{covariant structure} of the networks and to a \textbf{geometrical interpretability} of the learned parameters. The advances in the development of such a programme, whose first step is represented by the twin papers \cite{TSnaviga,naviga}, are to follow three directions:
\begin{description}
    \item[a)] Extension of the already developed algorithms for the $r=1$ to
    $r\geq 2$ and higher non-compact rank, according to the theory fully developed in \cite{pgtstheory}.
    \item[b)] Development of convolutional neural networks based on the reinterpretation of the members of Tits Satake universality classes as the total spaces of Tits Satake vector bundles.
    \item[c)] Reinstallation of the \textit{continuum view point} that represents, for instance, image data, as a discretization of a surface and hence of a section of a TS fiber bundle rather than as a single vector to be mapped to a single point in a manifold whose relation with the data is completely arbitrary.
    \item[d)] Investigation of the improvement in expressivity when considering layers with different non-compact-rank and Lie algebra type.
\end{description}
The mathematical results outlined in the present paper are all relative to the above first three points. The algebraic-geometric construction of the separator manifold in $r>1$ non-compact $\mathrm{U/H}$ spaces provides the only missing tool necessary for the development of point a). The
general structure of the Tits Satake vector bundles has been presented in the introductory section of the present paper, and we can now develop the crucial result for the separator. Finally, the long study of tessellations, harmonics, and Fuchsian fundamental domains has led to a clear-cut working direction. We have to complete the study of Laplacian eigenfunctions on the Fuchsian fundamental domain with the alternative strategies outlined in section \ref{nuovastrategia}. If this proposal is successful, we will
obtain at the same time a new result in pure mathematics, and a way to implement point c) with periodic functions rather than with
regular harmonics, which can already be done at the moment. The comparison of the two methods can be very instructive from the point of view of machine learning techniques.
\section{Acknowledgements}
M.T. and M.O. are deeply grateful to A. J. Di Scala and D.V. Alekseevsky for enlightening discussions. In particular we thank  A. J. Di Scala for pointing us to reference \cite{Olmo}.
\newpage
\appendix
\section{Structure of the quotient group \texorpdfstring{$\mathrm{GD_{96}}$}{GD96}}
\label{margarina96}
In this appendix, we analyse the structure of the finite quotient group $\mathrm{GD_{96}}$.
Abstractly, as it follows from the analysis of section \ref{gigioproietto}, the group can be defined using the following presentation:
\begin{equation}\label{presento96}
  \text{GD}_{96}\pmb{ }\equiv \pmb{ }\pmb{\langle }\pmb{\mathfrak{T}}\pmb{,}\pmb{\mathfrak{S}}\pmb{ }\pmb{,}\pmb{\mathfrak{R}\left|
\mathfrak{T}^8\right.}\pmb{=}\pmb{\mathfrak{S}
^3}\pmb{=}\pmb{\mathfrak{R}^2}\pmb{=}\pmb{\mathfrak{T}\mathfrak{S}\mathfrak{R}}\pmb{=}\pmb{(\mathfrak{T}\mathfrak{R}\mathfrak{S})}^3\pmb{ }
\pmb{=}\pmb{\text{Id}}\pmb{\rangle
}
\end{equation}
The most direct and surest way to study its structure is by means of a faithful linear representation. We were able to derive the form of
an irreducible 6-dimensional crystallographic representation, provided by the following integer valued generators:
\begin{equation}\label{STdegenerati}
 \mathfrak{T}=\left(
\begin{array}{cccccc}
 0 & 0 & 0 & 1 & 0 & 0 \\
 0 & 0 & 0 & 0 & 1 & 0 \\
 0 & 0 & 0 & 0 & 0 & 1 \\
 0 & 1 & 0 & 0 & 0 & 0 \\
 -1 & 0 & 0 & 0 & 0 & 0 \\
 0 & 0 & -1 & 0 & 0 & 0 \\
\end{array}
\right)\quad ; \quad \mathcal{S}=\left(
\begin{array}{cccccc}
 0 & 0 & 0 & 0 & 0 & 1 \\
 0 & 0 & -1 & 0 & 0 & 0 \\
 0 & 0 & 0 & 0 & 1 & 0 \\
 -1 & 0 & 0 & 0 & 0 & 0 \\
 0 & -1 & 0 & 0 & 0 & 0 \\
 0 & 0 & 0 & -1 & 0 & 0 \\
\end{array}
\right)
\end{equation}
that satisfy all the relations implied by the presentation (\ref{presento96}).
Indeed in agrement with eq.s (\ref{equadeltagenTS}-\ref{Udefinition}) we introduce the additional generators
\begin{equation}\label{romboldo}
  \mathcal{R} \equiv \mathfrak{T}\mathfrak{S}\quad;\quad\mathfrak{U} \equiv  \mathfrak{T}^2\mathfrak{S} ^2
\end{equation}
that take the following explicit form
\begin{equation}\label{RUdefini}
 \mathcal{R}=\left(
\begin{array}{cccccc}
 -1 & 0 & 0 & 0 & 0 & 0 \\
 0 & -1 & 0 & 0 & 0 & 0 \\
 0 & 0 & 0 & -1 & 0 & 0 \\
 0 & 0 & -1 & 0 & 0 & 0 \\
 0 & 0 & 0 & 0 & 0 & -1 \\
 0 & 0 & 0 & 0 & -1 & 0 \\
\end{array}
\right) \quad ; \quad \mathcal{U} = \left(
\begin{array}{cccccc}
 0 & 0 & 0 & 0 & -1 & 0 \\
 0 & 0 & 0 & 1 & 0 & 0 \\
 0 & 1 & 0 & 0 & 0 & 0 \\
 0 & 0 & 1 & 0 & 0 & 0 \\
 0 & 0 & 0 & 0 & 0 & 1 \\
 -1 & 0 & 0 & 0 & 0 & 0 \\
\end{array}
\right)
\end{equation}
and all the relations in (\ref{presento96}) evaluate to the identity.
\par
With our Mathematica Code for finite groups we have explicitly constructed the 96 group elements and applying
an appropriate routine we have derived the conjugation classes. They are as shown below:
\begin{eqnarray}\label{lepido}
\begin{array}{|c|c|c|c|c|c|c|c|c|c|c|}
\hline
  \searrow   & \pmb{\mathcal{C}_1} & \pmb{\mathcal{C}_2} & \pmb{\mathcal{C}_3} & \pmb{\mathcal{C}_4} & \pmb{\mathcal{C}_5}
& \pmb{\mathcal{C}_6} & \pmb{\mathcal{C}_7} & \pmb{\mathcal{C}_8} & \pmb{\mathcal{C}_9} & \pmb{\mathcal{C}_{10}} \\
\hline
 \text{class popul.}  & 1 & 3 & 3 & 3 & 6 & 12 & 12 & 12 & 12 & 32 \\
\hline
 \text{order}   & \pmb{1} & \pmb{2} & \pmb{4} & \pmb{4} & \pmb{4} & \pmb{2} & \pmb{8} &
\pmb{8} & \pmb{4} & \pmb{3} \\
\hline
  \text{represent.}  & \mathcal{E} & \mathcal{S}^2.\mathfrak{T}^4.\mathcal{S} & \mathcal{S}^2.\mathfrak{T}.\mathcal{S}^2.\mathfrak{T}
& \mathcal{S}.\mathfrak{T}^2.\mathcal{S}^2 & \mathcal{S}^2.\mathfrak{T}.\mathcal{S}^2.\mathfrak{T}^3 & \mathfrak{T}.\mathcal{S} &
\mathcal{S}.\mathfrak{T}.\mathcal{S}^2.\mathfrak{T}^4
& \mathfrak{T}^3.\mathcal{S}^2.\mathfrak{T}^2 & \mathcal{S}.\mathfrak{T}^5 & \mathcal{S}^2.\mathfrak{T}^2.\mathcal{S}^2 \\
\hline
\end{array}&&\nonumber\\
\end{eqnarray}
A deeper understanding of the group structure is obtained by introducing the following additional group elements
\begin{equation}\label{addigenni}
  \Psi \equiv  \mathfrak{T}^3\mathcal{S}\mathfrak{T}^3\mathcal{S}\quad ;\quad \Phi  \equiv  \mathfrak{T}^6
\end{equation}
that evaluate to the following matrices:
\begin{equation}\label{z4trini}
  \Psi =\left(
\begin{array}{cccccc}
 -1 & 0 & 0 & 0 & 0 & 0 \\
 0 & -1 & 0 & 0 & 0 & 0 \\
 0 & 0 & 0 & 0 & 0 & -1 \\
 0 & 0 & 0 & 0 & -1 & 0 \\
 0 & 0 & 0 & 1 & 0 & 0 \\
 0 & 0 & 1 & 0 & 0 & 0 \\
\end{array}
\right) \quad ; \quad  \Phi  = \left(
\begin{array}{cccccc}
 0 & -1 & 0 & 0 & 0 & 0 \\
 1 & 0 & 0 & 0 & 0 & 0 \\
 0 & 0 & -1 & 0 & 0 & 0 \\
 0 & 0 & 0 & 0 & -1 & 0 \\
 0 & 0 & 0 & 1 & 0 & 0 \\
 0 & 0 & 0 & 0 & 0 & -1 \\
\end{array}
\right);
\end{equation}
Both $\Psi$ and $\Phi$ are of order $4$ and commute with each other. Together they generate a subgroup:
\begin{equation}\label{cristino}
 \mathrm{GD_{96}} \, \supset \, \mathrm{GD_{16}} \, \sim \, \mathbb{Z}_4 \times \mathbb{Z}_4
\end{equation}
whose elements are listed below:
\begin{eqnarray}\label{elementG16}
  \mathfrak{g}_i & = & \left\{\mathcal{E},\Phi ,\Psi ,\Phi .\Phi ,\Phi .\Psi ,\Psi .\Psi ,\Phi .\Phi .\Phi ,\Phi .\Phi .\Psi ,
  \Psi .\Phi .\Psi ,\right.\nonumber\\
  &&\left.\Psi .\Psi
.\Psi ,\Phi .\Phi .\Phi .\Psi ,\Phi .\Psi .\Phi .\Psi ,\Psi .\Psi .\Phi .\Psi ,\Phi .\Phi .\Psi .\Phi .\Psi ,\Phi .\Phi .\Psi .\Psi .\Psi ,\Psi .\Psi
.\Phi .\Phi .\Phi .\Psi \right\}
\end{eqnarray}
If we add the generator $\mathcal{S}$ to the group elements of $\mathrm{GD_{16}}$ we obtain a group $\mathrm{GD_{48}}$ of order $48$
with respect to which, $\mathrm{GD_{16}}$ is a normal subgroup of index 3:
\begin{equation}\label{normalsubbo1}
  \mathrm{GD_{16}} \lhd \mathrm{GD_{48}}
\end{equation}
The quotient group $\mathrm{GD_{48}}/\mathrm{GD_{16}}$ is obviously $\mathbb{Z}_3$ and the 48 group elements can be written as follows:
\begin{equation}\label{G48elementi}
  \mathfrak{p}_i=\mathfrak{g}_i \quad ; \quad  \mathfrak{p}_{i+16 }= \mathcal{S} \mathfrak{g}_i
  \quad ; \quad\mathfrak{p}_{i+32 } = \mathcal{S}^2 \mathfrak{g}_i \quad ; \quad (i=1,\text{..},16)
\end{equation}
Finally we can verify that with respect to $\mathrm{GD_{96}}$, the group $\mathrm{GD_{48}}$ is a normal subgroup of index 2:
\begin{equation}\label{normalsubbo2bis}
\mathrm{GD_{96}} \rhd \mathrm{GD_{48}} \rhd \mathrm{GD_{16}} \sim \mathbb{Z}_4 \times \mathbb{Z}_4
\end{equation}
It follows from eq.(\ref{normalsubbo2}) that the quotient group defined by eq.s(\ref{presentaG96}-\ref{sequela2}) is a solvable group with a chain of
normal subgroups of prime index at each step. Consequently all the irreducible representations of the this group can be constructed
with the iterative procedure described in section 4.4  of the book \cite{fre2023book}.
\section{Markov Processes and the Heat Kernel on Riemannian Manifolds} \label{markovka}
Let $\mathbf{x},\mathbf{y}\in \mathcal{M}$ be two points of a Riemannian manifold $\left(\mathcal{M}, g\right) $ ,
let $t\in \mathbb{R}_+$ denote an abstract \textit{time variable} and let us name:
\begin{equation}\label{padroncina}
\mathfrak{p}\left(t,\mathbf{x},\mathbf{y}\right)
\end{equation} the probability density that in the interval of time $[0,t]$, the underlying dynamical system makes a transition
from the point $\mathbf{x}$ to the point $\mathbf{y}$. According with the general viewpoint elaborated in \cite{molcano},
such a probability density is governed by the following differential equation and boundary condition:
\begin{eqnarray}\label{servetta}
\partial_t \, \mathfrak{p}\left(t,\mathbf{x},\mathbf{y}\right) &=& \, \mathcal{A}_\mathbf{x} \,
\mathfrak{p}\left(t,\mathbf{x},\mathbf{y}\right)\nonumber\\ \mathfrak{p}\left(0,\mathbf{x},\mathbf{y}\right) & =&  \delta_\mathbf{x}(\mathbf{y}) \, = \, \text{appropriate Dirac delta function}
\end{eqnarray}
where the operator $\mathcal{A}_\mathbf{x}$ acting on the variable $\mathbf{x}$ is the following
\begin{equation}\label{paltaperuviana}
  \mathcal{A}_\mathbf{x} \, = \, \ft 12 \, \Delta_{LB} \, - \, \mathbf{U}
\end{equation}
having denoted by $\Delta_{LB}$ the Laplace-Beltrami operator for scalar functions on the Riemannian
manifold $\left(\mathcal{M}, g\right) $ and by $\mathbf{U} \in \Gamma\left[T\mathcal{M},\mathcal{M}\right]$
a section of the tangent bundle, namely a vector field.
This means that in any open chart $\mathcal{V}$ of an atlas covering the manifold $\mathcal{M}$ we have
\begin{eqnarray}
\label{diffondo}
\Delta_{LB} \,\phi(\mathbf{x}) &=& \frac{1}{\sqrt{\mathrm{det} (g)} }\partial_i \,
\left( \sqrt{\mathrm{det} (g)} \,g^{ij} \partial_j \phi(\mathbf{x})\right) \quad ; \quad \phi \in \mathbb{C}^{\infty}(\mathcal{V})
\nonumber\\ \mathbf{U} \phi(\mathbf{x}) &=& U^i(\mathbf{x}) \partial_i \phi(\mathbf{x})
\end{eqnarray}
The first order term of the differential operator $\mathcal{A}_\mathbf{x}$ has been named  $\mathbf{U}$
because it conspires, if brought on the left hand side of the equation, to describe the classical diffusion equation
by means of thermal convection with a hot fluid transporting the heat. Indeed if $U^i(\mathbf{x})$ is the velocity field of the fluid
the temperature field $T(t,\mathbf{x})$ does indeed satisfy the following diffusion equation:
\begin{equation}\label{convectdiffuse}
\partial_t \, T + \mathbf{U} \, T \, = \, \Delta_{LB} \, T
\end{equation}
and given the temperature profile at initial time $T(0,\mathbf{x})=T_0(\mathbf{x})$ the solution
of equation (\ref{convectdiffuse}) with such a boundary condition can be expressed in terms of the probability
density $\mathfrak{p}\left(t,\mathbf{x},\mathbf{y}\right)$ by means of the
integral: \begin{equation}\label{integralindo}
 T(t,\mathbf{x}) \, = \, \int_{\mathcal{M}} \, \mathfrak{p}\left(t,\mathbf{x},\mathbf{y} \right)\,
 T_0(\mathbf{y})\, \mathrm{d}\mu(\mathbf{y})
\end{equation}
which identifies the transition probability density of the Markov process with the classical Heat kernel:
\begin{equation}\label{ittokernello}
  \mathfrak{p}\left(t,\mathbf{x},\mathbf{y} \right) \, = \,  H\left(t,\mathbf{x},\mathbf{y} \right)
\end{equation}
at least in the case of static diffusion at $\mathbf{U}=0$.
\par
The most relevant result in the analysis of  Molkhanov, author of \cite{molcano}, developed on the basis of classical previous results by
Kolmogorov \cite{kolmogo}, Dynkin \cite{dynkusprob} and Varadhan \cite{varado}, is that in any open chart $\mathcal{V}\subset \mathcal{M}$ which
is sufficiently small such that any two points $\mathbf{x},\mathbf{y} \in \mathcal{V}$ can be joined by a unique shortest geodesics
$\gamma_{\mathbf{x},\mathbf{y}}$,  the behavior of the \textit{heat-probability kernel} is the following one:
\begin{eqnarray}\label{keyrelata1}
 \mathfrak{p}\left(t,\mathbf{x},\mathbf{y} \right) \, \stackrel{t \, \to \, 0}{\sim} \, \frac{B(\mathbf{x},\mathbf{y})}{t^{n/2}}
 \, \exp\left[-\frac{\mathrm{d}_{geo}^2(\mathbf{x},\mathbf{y})}{2\,t}\right] \nonumber\\
 \mathrm{d}_{geo}^2(\mathbf{x},\mathbf{y})\, \equiv \, \int_{\gamma_{\mathbf{x},\mathbf{y}}} \,
 \frac{d x^i(s)}{ds} \, \frac{d x^j(s)}{ds} g_{ij}\left( \mathbf{x}(s)\right) \, ds
\end{eqnarray}
where the time independent function $B(\mathbf{x},\mathbf{y})$ is a characteristic feature of the specific Riemannian manifold under consideration
while the function $\mathrm{d}_{geo}^2(\mathbf{x},\mathbf{y})$  is by definition the geodesics distance between the two manifold points
$\mathbf{x},\mathbf{y}$ in those open charts where it can be defined. \par As a consequence of eq. (\ref{keyrelata1}) we have the following limit:
\begin{equation}\label{keyrelata2}
  \lim_{t \to 0} \, - 2 \, t \, \mathfrak{p}\left(t,\mathbf{x},\mathbf{y} \right) \, = \, \mathrm{d}_{geo}^2(\mathbf{x},\mathbf{y})
\end{equation}
From the epistemological point of view, the great value of this analysis is due to Dynkin \cite{dynkusprob}, and it was pointed out by
Molkhanov at the very beginning of his paper. It is the following. Under quite general smoothness assumptions,
in any strong Markov process on a smooth manifold $\mathcal{M}$, the transition probability density (\ref{padroncina}) always
satisfies a parabolic differential equation of the second order, such as that in eq.s (\ref{servetta},\ref{paltaperuviana}).
The interpretation of the variable dependent coefficients of the second order differential operator as a metric on a Riemannian metric
on the manifold $\mathcal{M}$ comes a posteriori. In other words, the underlying Riemannian structure of a given variety can be probed through the analysis of diffusion Markov processes occurring on that variety. \par In the case of deep learning, we might reveal the underlying structure of a Riemannian manifold through random walks on the manifold itself.

\newpage
\newpage
\bibliography{allesbiblio}
\bibliographystyle{ieeetr}
\end{document}